\documentclass{ieeetj}

\usepackage{cite}
\usepackage{amsmath,amssymb,amsfonts}
\usepackage{graphicx,color}
\usepackage{textcomp}
\usepackage{xcolor}
\usepackage{hyperref}
\hypersetup{hidelinks=true}

\usepackage[utf8]{inputenc} 
\usepackage[T1]{fontenc}    
\usepackage{url}            
\usepackage{booktabs}       
\usepackage{amsfonts}       
\usepackage{nicefrac}       
\usepackage{microtype}      

\usepackage{bm}
\usepackage{caption}
\usepackage{subcaption}
\usepackage{multirow}
\usepackage{makecell}
\usepackage{fontawesome}
\usepackage{marvosym}
\usepackage{wrapfig}
\usepackage{algpseudocode}
\usepackage{cleveref}
\usepackage{rotating}

\def\BibTeX{{\rm B\kern-.05em{\sc i\kern-.025em b}\kern-.08em
    T\kern-.1667em\lower.7ex\hbox{E}\kern-.125emX}}
\AtBeginDocument{\definecolor{tmlcncolor}{cmyk}{0.93,0.59,0.15,0.02}\definecolor{NavyBlue}{RGB}{0,86,125}}

\def\OJlogo{\vspace{-4pt}$<$Society logo(s) and publication title will appear here.$>$}
\def\seclogo{\vspace{10pt}$<$Society logo(s) and publication title will appear here.$>$}

\def\authorrefmark#1{\ensuremath{^{\textbf{#1}}}}

\begin{document}
\receiveddate{XX Month, XXXX}
\reviseddate{XX Month, XXXX}
\accepteddate{XX Month, XXXX}
\publisheddate{XX Month, XXXX}
\currentdate{XX Month, XXXX}
\doiinfo{XXXX.2022.1234567}

\markboth{}{Anonymous author}

\title{Multi-Robot System for Cooperative Exploration in Unknown Environments: A Survey}

\author{
Chuqi Wang\authorrefmark{1*}, Chao Yu\authorrefmark{1*}, Xin Xu$^{1}$, Yinuo Chen\authorrefmark{1}, Yuman Gao\authorrefmark{2}, Xinyi Yang\authorrefmark{1}, Wenhao Tang\authorrefmark{3}, Shu'ang Yu\authorrefmark{14}, Feng Gao\authorrefmark{1}, Zhuozhu Jian\authorrefmark{3}, Xinlei Chen\authorrefmark{3}, Fei Gao\authorrefmark{2}, Boyu Zhou\authorrefmark{5}, Yu Wang\authorrefmark{1}, Fellow, IEEE}

\affil{Tsinghua University, Beijing, 100084, China.}
\affil{Zhejiang University, Zhejiang, 310009, China.}
\affil{Tsinghua Shenzhen International Graduate School,
Shenzhen, 518055, China.}
\affil{Shanghai Artificial Intelligence Laboratory,
Shanghai, 200030, China.}
\affil{Southern University of Science and Technology, 
Guangdong, 518055, China.}
\corresp{* Equal Contribution. \\ Corresponding authors: Chao Yu and Yu Wang(email: \{yuchao,yu-wang\}@tsinghua.edu.cn).}
\authornote{This research was supported by National Natural Science Foundation of China (No.62406159, 62325405, 2022YFC3300703), Postdoctoral Fellowship Program of CPSF under Grant Number (GZC20240830, 2024M761676), China Postdoctoral Science Special Foundation 2024T170496. The authors are grateful to Huan Yu for his insightful comments on the motion planning module.}

\begin{abstract}

With the real need of field exploration in large-scale and extreme outdoor environments, cooperative exploration tasks have garnered increasing attention. This paper presents a comprehensive review of multi-robot cooperative exploration systems. First, we review the evolution of robotic exploration and introduce a modular research framework tailored for multi-robot cooperative exploration. Based on this framework, we systematically categorize and summarize key system components. As a foundational module for multi-robot exploration, the localization and mapping module is primarily introduced by focusing on global and relative pose estimation, as well as multi-robot map merging techniques. The cooperative motion module is further divided into learning-based approaches and multi-stage planning, with the latter encompassing target generation, task allocation, and motion planning strategies. Given the communication constraints of real-world environments, we also analyze the communication module, emphasizing how robots exchange information within local communication ranges and under limited transmission capabilities. In addition, we introduce the actual application of multi-robot cooperative exploration systems in DARPA SubT Challenge. Finally, we discuss the challenges and future research directions for multi-robot cooperative exploration in light of real-world trends. This review aims to serve as a valuable reference for researchers and practitioners in the field.
\end{abstract}

\begin{IEEEkeywords}
Communication, Cooperative Planning, Localization and Mapping, Multi-robot Cooperative Exploration System 
\end{IEEEkeywords}


\maketitle

\section{INTRODUCTION}
\label{sec: intro}

\begin{figure*}
    \centering
    \begin{subfigure}[b]{0.3\linewidth}
        \includegraphics[width=\linewidth]{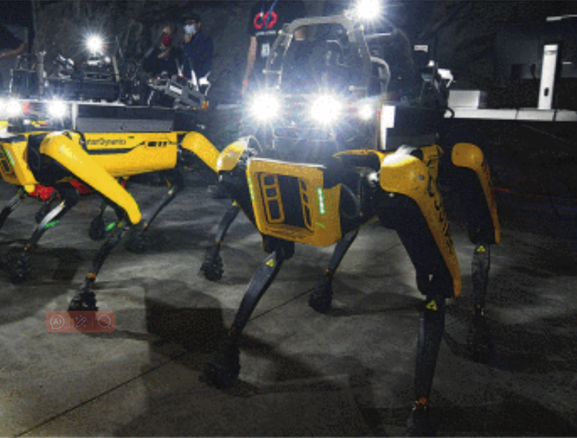}
        \caption{Multiple quadrupeds, each equipped with detection devices, are preparing for cave exploration.~\cite{Ackerman2022DarpaAutoRobo}}
        \label{fig:osprey}
    \end{subfigure}
    \hfill
    \begin{subfigure}[b]{0.3\linewidth}
        \includegraphics[width=\linewidth]{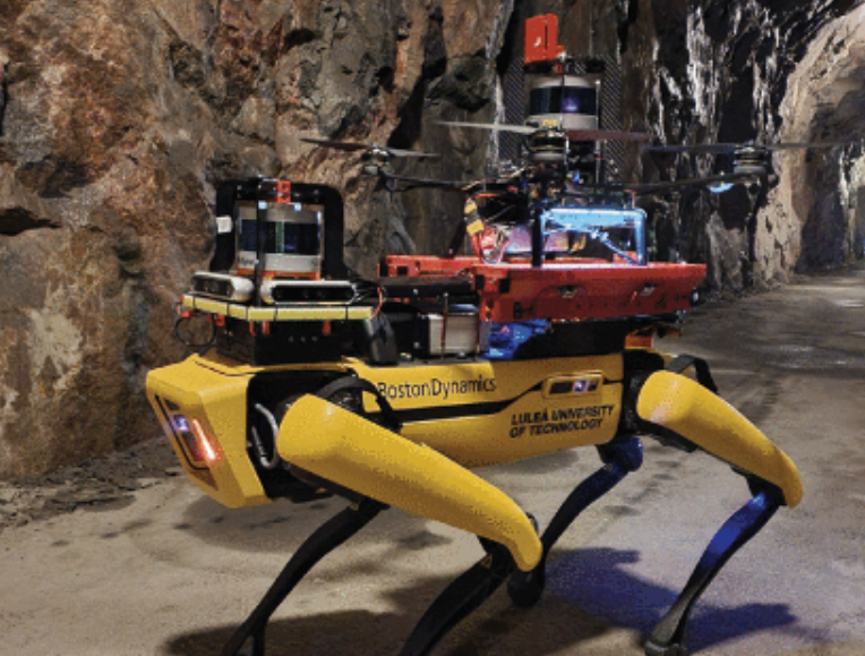}
        \caption{Legged-aerial explorer equipped with its full sensor suite alongside the quadrotor carrier platform.~\cite{Morrell2024CoSTAR}}
        \label{fig:CoSTAR}
    \end{subfigure}
    \hfill
    \begin{subfigure}[b]{0.3\linewidth}
        \includegraphics[width=\linewidth]{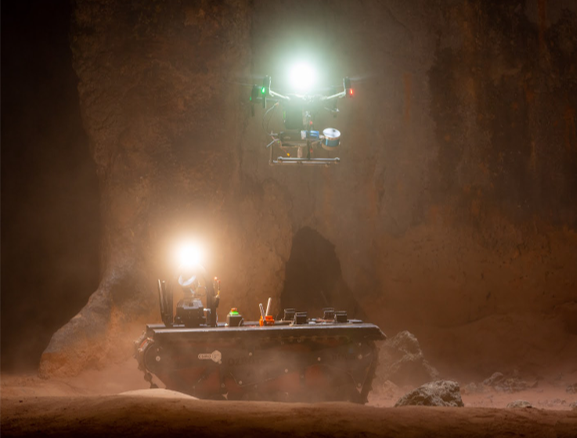}
        \caption{Unmanned Aerial Vehicle (UAV) and Unmanned Ground Vehicle (UGV) collaborative exploration in the cave.~\cite{CSIRO2022Heterogeneous}}
        \label{fig:heterogeneous}
    \end{subfigure}
    \hfill
    \begin{subfigure}[b]{0.3\linewidth}
        \includegraphics[width=\linewidth]{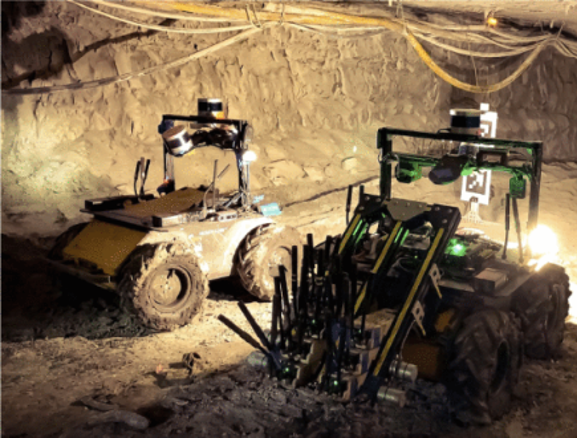}
        \caption{Two autonomous ground rovers explore the network of underground tunnels.~\cite{Ebadi2020LAMP}}
        \label{fig:heterogeneous}
    \end{subfigure}
    \hfill
    \begin{subfigure}[b]{0.3\linewidth}
        \includegraphics[width=\linewidth]{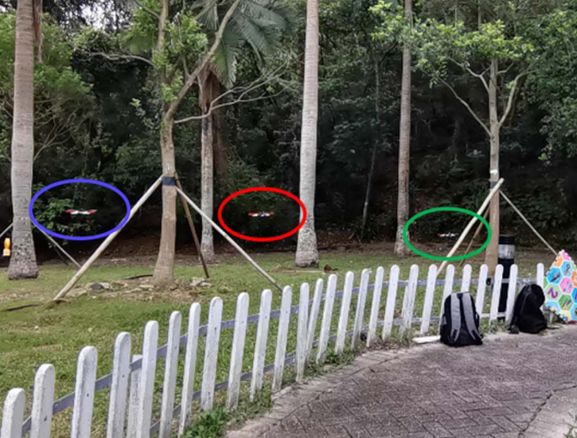}
        \caption{Three quadrotors are collaborating to explore the forest.~\cite{zhou2023racer}}
        \label{fig:heterogeneous}
    \end{subfigure}
    \hfill
    \begin{subfigure}[b]{0.3\linewidth}
        \includegraphics[width=\linewidth]{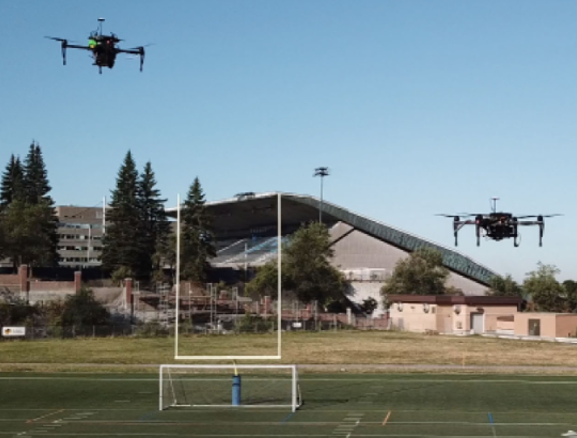}
        \caption{Two quadrotors are calibrating their positioning on the playground.~\cite{door-slam}}
        \label{fig:heterogeneous}
    \end{subfigure}
    \caption{Multiple field robots performing exploration tasks.}
    \label{fig:field_robot}
\end{figure*}

\IEEEPARstart{M}{ulti-robot} systems provide substantial benefits in cooperative tasks, especially in information-gathering missions, by improving reliability, robustness, and operational efficiency. Common applications include multi-robot cooperative search and rescue~\cite{queralta2020collaborative}, surveillance~\cite{roman2006framework}, inspection~\cite{xu2024Cost}, and mine surveying~\cite{miller2020mine}. 
Among these applications, exploration in initially unknown environments is often a key yet challenging prerequisite, as it provides maps for navigation and decision-making. 
For exploration tasks, multi-robot systems can greatly enhance efficiency by enabling parallel area coverage and flexible task allocation, particularly in large-scale environments. 
Furthermore, compared to single-robot systems, multi-robot teams offer increased system-level robustness by providing redundancy, ensuring the mission can continue even if individual robots fail. 
However, multi-robot systems face challenges, including more complex planning, precise localization issues, and scalability concerns related to the number of robots in the system, all of which require further investigation and solutions.

Initially, multi-robot cooperative exploration systems are limited to handling lab settings or very simple environments, such as small offices~\cite{yamauchi1997frontier}.
Driven by real-world needs, these systems have evolved to handle large-scale and extreme outdoor environments, such as caves~\cite{Ackerman2022DarpaAutoRobo,Morrell2024CoSTAR,CSIRO2022Heterogeneous}, underground tunnels~\cite{Ebadi2020LAMP}, forests~\cite{zhou2023racer}, and playgrounds~\cite{door-slam} (shown in Fig.~\ref{fig:field_robot}). The complexity of these environments originates from four core challenges: denied GNSS signals, large-scale scenarios, tough terrains, and intermittent communication. These challenges drive the development of heterogeneous robot teams to improve system robustness and exploration efficiency, which also becomes a challenge. 
Developing multi-robot cooperative exploration systems for extreme and large-scale field environments presents several challenges, which can be broadly categorized as follows:

\textbf{Accurate localization and map fusion}: 
For multi-robot exploration, robots within a team need a shared understanding of both the environment and their respective locations. However, in many practical unknown environments, such as caves and buildings, the Global Navigation Satellite System (GNSS) may be unavailable, and external base station signals may be absent. As a result, robots must rely on proprioception for localization and mapping.
For cooperative localization, robots need to localize both themselves and their positions relative to others, but this process will be affected by noise and accumulated errors, leading to localization drift and even failure. To correct localization, multi-robot systems often require extensive data exchange, which introduces challenges such as incorrect data association and perceptual aliasing.
For cooperative mapping, efficiently integrating map data from different robots while maintaining global consistency is a significant challenge of multi-source alignment. Additionally, cooperative mapping requires continuously fusing the map while localization results are dynamically corrected, further increasing the complexity of the process.

\textbf{Efficient cooperative mechanism}: 
In multi-robot cooperative exploration systems, the cooperative mechanism must address two key aspects: driving robots to acquire information about unknown environments and ensuring balanced workload distribution through optimized goal assignment methods. 
Ineffective goal assignment can lead to workload imbalance, where certain robots remain actively engaged while others have completed their tasks. This problem is especially pronounced in environments with complex topological structures, such as subterranean cave networks or benthic terrains, potentially reducing exploration efficiency and even causing mission failure.
Moreover, the mechanism should integrate trajectory generation algorithms capable of producing collision-free, dynamical feasible, and efficient trajectories to navigate to assigned targets.

\textbf{Constrained communication}:
Real-world cooperative exploration missions inevitably involve communication-restricted environments, making it a crucial factor to consider.
Two primary communication constraints should be addressed: Bandwidth constraints, often imposed by the payload limitations of onboard equipment, restrict data transmission rates and prevent lossless data exchange. To mitigate these limitations, it is essential to develop compact data representation methods that balance transmission efficiency with reconstruction accuracy. 
Moreover, range constraints, arising from both the intrinsic limitations of hardware transmission capabilities and environmental attenuation effects, pose the risk of partial system disconnections.
This challenge becomes particularly severe in extreme environments such as subterranean voids or deep-sea fields.
This necessitates the integration of resilient communication strategies with adaptive protocols.

\begin{figure*}[ht]
    \centering
    \includegraphics[width=0.8\linewidth]{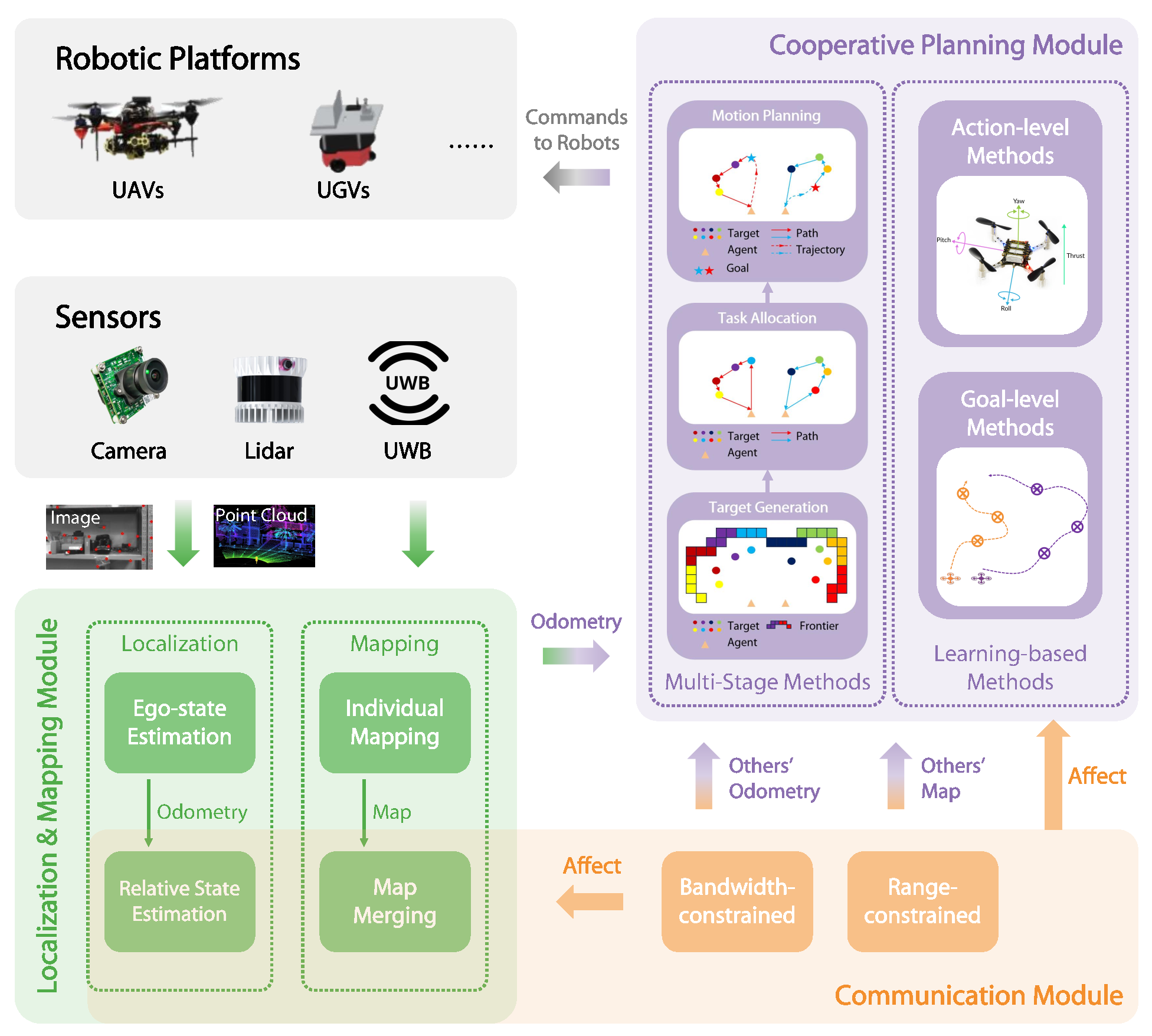}
    \caption{The coordination between the modules of the multi-robot cooperative exploration system.}
    \label{fig:overview}
\end{figure*}

All the challenges guide the researchers to develop a multi-robot cooperative exploration system with accurate localization and map fusion, efficient cooperative mechanism, representation methods with small data transmission volume and high precision, and efficient communication strategies. 
Based on the summary of the main challenges faced by multi-robot cooperative exploration systems when exploring extreme and large-scale outdoor environments, we divide the technical composition of multi-robot cooperative exploration systems into three corresponding modules: \textit{\textbf{localization and mapping}}, \textit{\textbf{cooperative planning}}, and \textit{\textbf{communication}}. Each module is designed to address the specific difficulties. Fig.~\ref{fig:overview} describes the coordination between the modules of the multi-robot cooperative exploration system.
First, the robots need to perform self-localization and determine their relative positions through localization techniques. Next, they construct an environmental map based on sensor data, which serves the subsequent path planning and navigation. On this basis, the robots perform cooperative planning by considering the current state, environmental information, and other factors, to explore the unknown environment efficiently. Throughout the process, the communication module facilitates information exchange, such as transmitting pose data and local maps. These modules are interrelated and mutually influence each other, forming the technical framework for multi-robot cooperative exploration.

\begin{sidewaysfigure*}[htp]
    \centering
    \includegraphics[width=\textheight]{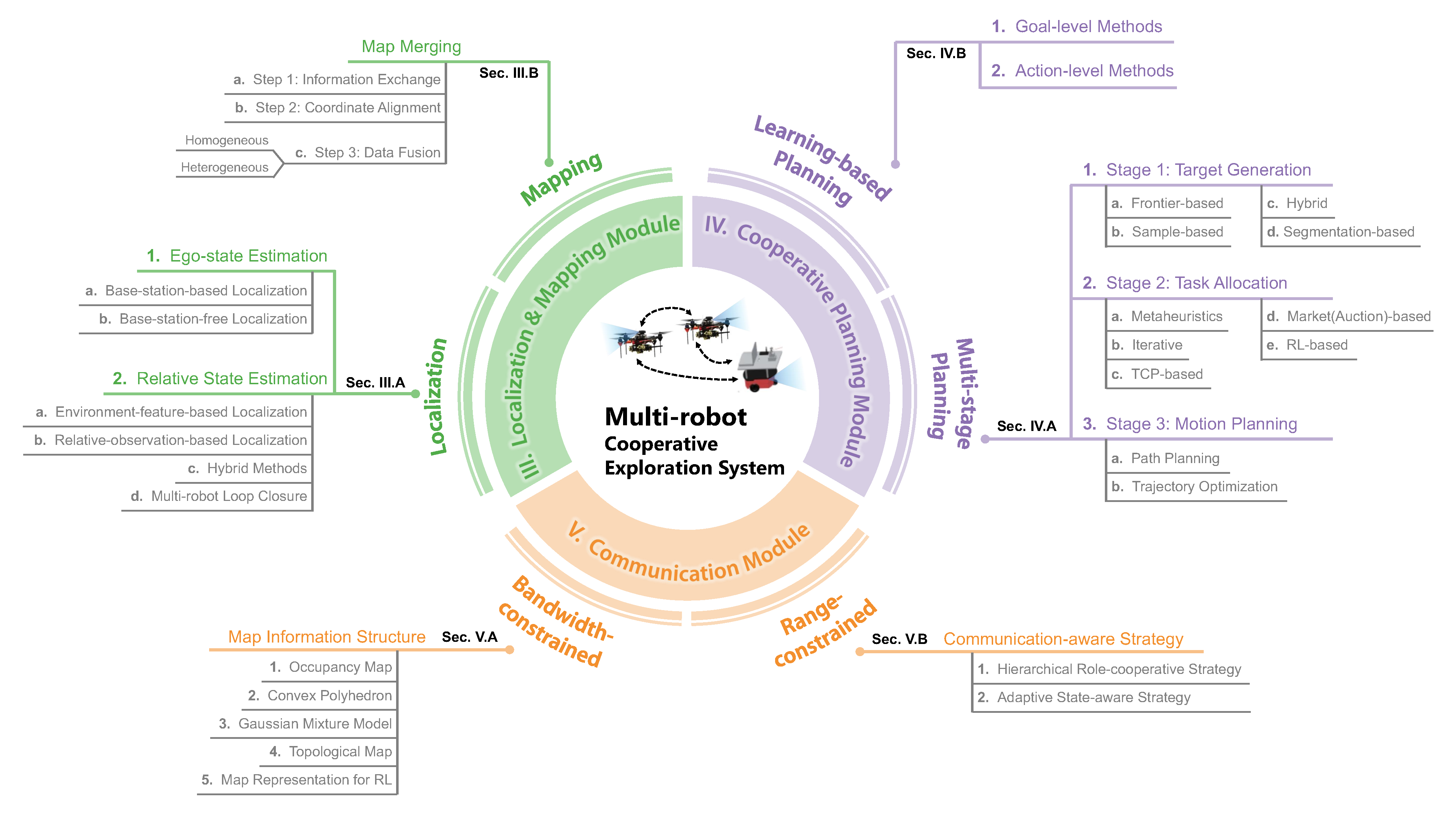}
    \caption{Overview structure of multi-robot cooperative exploration system in this survey.}
    \label{fig:structure}
\end{sidewaysfigure*}

Over the past few decades, researchers have worked hard to address the various challenges faced by multi-robot cooperative exploration, greatly advancing the development of this field. Therefore, this paper primarily surveys existing multi-robot cooperative exploration systems with a focus on map construction, providing the relevant literature on multi-robot cooperative exploration, aiming to comprehensively summarize the main concepts, current progress, unresolved issues, and emerging research trends in this field. We hope this survey will not only help newcomers quickly understand the current technical level but also provide valuable insights for experienced researchers, guiding future design choices and research directions. Since multi-robot cooperative exploration is a large-scale system involving multiple technical aspects, this paper focuses on the unique problems in multi-robot cooperative exploration; for areas overlapping with single-robot systems, such as target navigation and map construction, only brief overviews are provided. For more in-depth information on a single module, please refer to~\cite{ebadi2023present,Abujabal_Fareh_Sinan_Baziyad_Bettayeb_2023,MADRIDANO2021114660,khamis2015multi,skaltsis2021task,comm_survey}

The structure of the paper is as follows: Section \ref{sec: overview} provides an overview of the technical background of multi-robot cooperative exploration systems, reviews their development history, and discusses the related tasks in detail. Section \ref{sec: LM} introduces localization and map fusion methods used to extend the map boundaries in multi-robot cooperative exploration systems. Section \ref{sec: coop} presents the cooperative planning methods for multi-robot systems, including multi-stage planning and learning-based planning. Section \ref{sec: comm} discusses the robot interaction issues under limited communication range and bandwidth conditions. Section \ref{sec: fie} introduces the actual application of multi-robot cooperative exploration systems in field environments. Section \ref{sec: open} summarizes the current unresolved issues and explores potential future research directions. Finally, Section \ref{sec: conclu} concludes the paper by summarizing the research content. Fig. \ref{fig:structure} shows the overview structure of multi-robot cooperative exploration systems in this survey.

\begin{figure*}[htp]
    \centering
    \includegraphics[width=\linewidth]{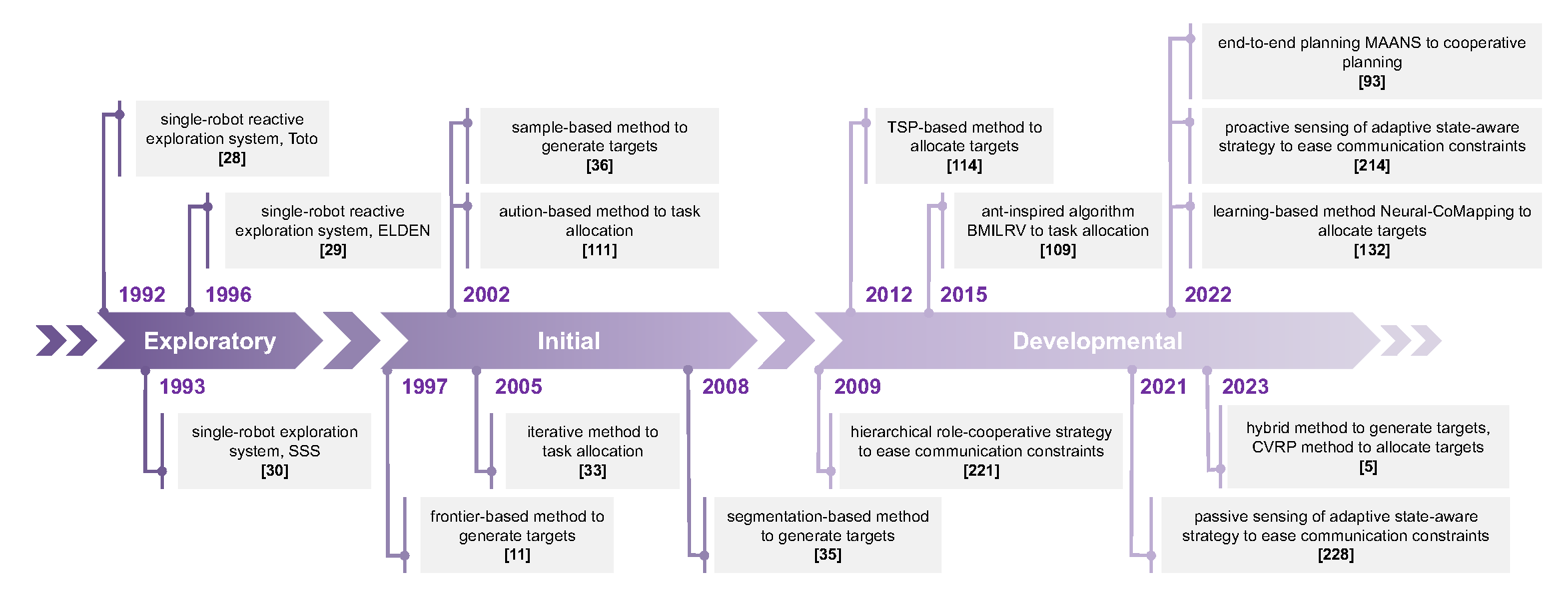}
    \caption{The evolutionary history of multi-robot cooperative exploration system.}
    \label{fig:history}
\end{figure*}

\section{Overview of the Multi-Robot System for Cooperative Exploration}
\label{sec: overview}

One of the best ways to understand a system is to first learn about its development history. The development history encompasses various aspects, including the research direction, technological composition, and application development of the system. This section aims to review the development history of multi-robot cooperative exploration systems (shown in Fig.~\ref{fig:history}), discuss their current technological structure, and highlight the challenges they face, providing a foundation for the subsequent research in this paper.

\vspace{2mm}
\noindent\textbf{II-A. Exploratory Period}
\vspace{2mm}

\label{sec: expperiod}
Research into robot autonomous exploration systems started in the 1980s. We define the period from then until the mid-1990s as the exploratory period. During this time, early studies mainly used simulations to test the proposed techniques. The first instances of autonomous robot exploration in real environments began in the early 1990s. Notably, the research conducted during this phase had already established the system's key objectives: enabling robots to navigate independently, explore confined areas, and ultimately create environmental maps. This period also saw the creation of techniques to tackle specific issues, including collision avoidance.

During this phase, research on robotic autonomous exploration systems predominantly focused on single-robot exploration, which led to an emphasis on the development of exploration strategies. Mataric~\cite{mataric1992integration} introduced the reactive exploration system, Toto, which operated in an office environment by employing elementary topological path planning, wall-following, and sonar-based autonomous obstacle avoidance techniques to explore the environment and construct a map. Yamauchi et al.~\cite{yamauchi1996spatial} proposed a single-robot exploration system known as ``Exploration and Learning in Dynamic ENvironments'' (ELDEN), which integrates reactive exploration with an incremental map-building approach, enabling adaptability to dynamic changes in the environment. The design of reactive exploration substantially enhances the robustness and safety of the system, thereby significantly extending its operational duration. Nevertheless, due to the limited range of sensors and the rudimentary motion planning methodologies employed, certain areas of the environment may become occluded by obstacles during exploration, resulting in instances where reactive exploration alone fails to completely map the environmental structure.

During the same period, Connell et al.~\cite{connell1993rapid} created an exploration system to show the ``servo, subsumption, symbolic'' (SSS) architecture they were working on, which, however, was limited to mapping environments where a corridor is perpendicular to doors or other corridors. Additionally, Thrun et al.~\cite{thrun1996integrating} introduced a system that integrated evidence grids with topological maps, capable of exploring corridor networks within large buildings. This system, however, required that walls be either parallel or perpendicular and remain unobstructed by any obstacles.

Constrained by sensor performance and the computational capabilities of the hardware at the time, the strategies developed during this period typically imposed stringent environmental requirements. Consequently, the versatility of autonomous robot exploration systems across diverse settings was significantly limited.

\vspace{2mm}
\noindent\textbf{II-B. Initial Period}
\vspace{2mm}

We define the period from the late 1990s to the late 2000s as the initial stage of the multi-robot cooperative exploration system.

To address the challenges outlined in Section \ref{sec: expperiod}, Yamauchi~\cite{yamauchi1997frontier} introduced a frontier-based exploration (FBE) strategy. This strategy leverages the boundary regions between the robot’s known open spaces and the unknown areas within the environment—referred to as frontiers—as primary exploration targets. By continuously advancing toward new frontiers, a robot can incrementally expand its map, ultimately surveying the entire accessible environment. The frontier concept liberates the exploration process from being confined to specific regions, enabling robots to autonomously select and navigate along the periphery of the known map boundaries in both open and cluttered spaces. This approach facilitates flexible and autonomous exploration, ensuring comprehensive map construction.

In an effort to optimize the exploration process, Yamauchi~\cite{yamauchi1998frontier} extended the FBE method to a multi-robot system. This system incorporates a probability-based grid mapping approach and robot path planning capabilities as described in \cite{yamauchi1997frontier}, where robots select the nearest frontier to explore and plan their paths accordingly. Additionally, the system includes communication functionalities that allow for map merging among robots and the planning of collision-free paths. However, due to the simplistic rules governing target point selection, the system is susceptible to redundant exploration of certain areas and scenarios where one robot may impede the exploration efforts of others. This multi-robot cooperative exploration system represents one of the earliest efforts to investigate cooperative exploration among multiple robots, marking the inception of research in this domain.

Building upon the frontier-based exploration strategy, Burgard et al.~\cite{burgard2005coordinated} introduced a method employing a utility function to calculate information gain and greedily assign exploration targets to different robots. This approach aims to balance the workload among robots and minimize redundant exploration. Extending Burgard et al.’s work~\cite{burgard2005coordinated}, Butzke et al.~\cite{butzke2011planning} enhanced the utility function to account for multiple objectives, enabling the adjustment of exploration priorities for individual robots as well as for the entire robotic fleet. While the utility function-based method is effective for assigning single target points to individual robots, its performance may decline when multiple targets need to be concurrently allocated among a team of robots. To address this limitation, Wurm et al.~\cite{wurm2008coordinated} proposed a novel task allocation approach within the frontier-based exploration framework. By segmenting the explored environment into multiple sections using a Voronoi graph, this method assigns robots to different local maps for exploration. This approach effectively mitigates the issues associated with utility function-based methods. However, it can result in uneven task distribution, where some robots may complete their exploration tasks earlier while others remain engaged in more complex areas.

In addition to frontier-based strategies, González-Baños et al.~\cite{gonzalez2002navigation} proposed a pioneering sampled-based approach known as the next-best-view (NBV)~\cite{connolly1985determination} exploration strategy. This method involves randomly sampling points within the largest unobstructed regions of the environment. For each sampled point, the strategy evaluates information gain and movement cost, guiding the robot to the position that offers the highest information gain with the lowest movement cost. This facilitates the continuous updating of the map.

Overall, this period primarily focuses on developing exploration strategies and task allocation issues. In exploration tasks, the selection of target points for local or global map updates by robots has become an autonomous process, no longer constrained by specific environmental requirements. Additionally, research has begun addressing the allocation of target points among different robots to reduce redundant exploration, optimize load balancing, and enhance exploration efficiency. However, while this period has developed a foundational understanding of these issues, the technical integration of multi-robot cooperative exploration systems remains incomplete, often resulting in exploration efficiency that falls short of expectations.

\vspace{2mm}
\noindent\textbf{II-C. Developmental Period}
\vspace{2mm}

The period from the 2010s to the present marks the development stage of multi-robot cooperative exploration systems, characterized by a significant shift towards modularization in their research and development. During this stage, researchers have increasingly focused on decomposing system design into distinct yet interdependent modules. Specifically, the current research framework is broadly organized into three primary modules: the localization and mapping module, the cooperative planning module, and the communication module~\cite{lluvia2021active, queralta2020collaborative, comm_survey}. This modular approach enables targeted problem-solving within each module while ensuring seamless integration and compatibility across the system.

\vspace{2mm}
\noindent\textbf{II-C-1. Localization and Mapping Module}
\vspace{2mm}

The localization and mapping module plays a pivotal role in multi-robot cooperative exploration systems. Its primary function is to estimate each robot’s pose within the system and generate a global map through map merging techniques. The localization component primarily involves ego-state estimation and relative state estimation between robots. Thanks to the foundation of Simultaneous Localization and Mapping (SLAM), ego-state estimation is already well-established. Meanwhile, relative state estimation between robots, as a new challenge in multi-robot systems, helps mitigate drift in individual localization and further improves overall system accuracy, leading to extensive research in this area. We categorize the mainstream methods into three types: environment-feature-based, relative-observation-based, and hybrid. Additionally, we highlight the important technique of multi-robot loop closure. The map merging process mainly involves three steps: information exchange, coordinate alignment, and data fusion. The first two steps are inherently linked to the communication and localization modules, respectively. Data fusion can further be classified into homogeneous and heterogeneous map merging, depending on the types of robots and sensors in the system. Through the collaboration of localization and mapping, accurate robot pose information and a foundational map are provided to the cooperative planning module.

\vspace{2mm}
\noindent\textbf{II-C-2. Cooperative Planning Module}
\vspace{2mm}

The cooperative planning module is broadly categorized into multi-stage planning and learning-based planning. Multi-stage planning is further subdivided into three key components: target generation, task allocation, and motion planning.

The primary objective of target generation is to determine future target points for robot movement. 
Task allocation assigns target points to individual robots, minimizing redundant exploration and resource consumption to enhance overall efficiency. According to Gerkey et al.~\cite{gerkey2004formal}, task allocation in multi-robot cooperative exploration falls under the category of ``Single-task robots, multi-robot tasks, time-extended assignment'' (ST–MR–TA). This problem is inherently NP-hard and dynamic, with no mathematically optimal general solution. As a result, extensive research has been conducted to develop strategies for balancing robot workloads and enhancing exploration efficiency.
Motion planning ensures the safe execution of trajectories to connect these target points while minimizing unnecessary backtracking once new targets are assigned. 

With the rise of reinforcement learning (RL), researchers have increasingly applied RL to multi-robot cooperative exploration problems. RL has demonstrated significant potential in both learning-based planning and multi-stage planning, achieving notable progress in addressing the complexities of multi-robot planning~\cite{garaffa2021reinforcement}.

\vspace{2mm}
\noindent\textbf{II-C-3. Communication Module}
\vspace{2mm}

The communication module enables the exchange of map data, position information, and trajectory data between robots, supporting map fusion and coordinated motion planning. During multi-robot exploration, the transmission of large datasets, such as maps, imposes high demands on bandwidth. Communication may suffer from high latency or even disruptions due to obstacles or long distances~\cite{comm_survey}. As the localization and mapping module and the cooperative planning module heavily rely on the communication module, failures in communication can significantly degrade their performance and, in severe cases, result in the failure of the entire system. As research progresses, this module has received growing attention.

Overall, research on multi-robot cooperative exploration systems has transitioned from lateral expansion to in-depth vertical exploration in this phase. The focus has shifted towards refining existing methodologies and addressing emerging challenges. A notable trend in this research is modularization, where independent solutions are designed for specific system modules. Given the interdependence among modules, improvements in one module often lead to performance enhancements in others. While modularization allows for targeted solutions, integrating these modules into a cohesive system remains a major challenge in multi-robot exploration.

In the following sections, we will provide a detailed review of these three critical modules.
\section{Localization and Mapping Methods}
\label{sec: LM}

Localization and mapping are fundamental components of exploration, whether performed by a single robot or a multi-robot system. Localization involves ego-state estimation and relative state estimation. Ego-state estimation refers to determining the pose, i.e., position and orientation, of each robot, while relative state estimation focuses on calculating the relative pose between robots. Mapping, on the other hand, involves integrating data from multiple robots to create a comprehensive environmental map. 

This section examines the localization and mapping modules in multi-robot cooperative exploration, emphasizing the complexities unique to multi-robot systems. In Section \ref{sec: loc} on localization, we will discuss various methods of ego-state and relative state estimation, with a primary focus on the latter in the context of multi-robot systems. In Section \ref{sec: map} on mapping, we will address the unique challenges of map merging in multi-robot systems. It is important to clarify that, in this context, the term ``map'' refers to representations of an environment’s geometric and semantic information, typically for collaborative decision-making, visualization, and similar tasks, rather than feature maps used in localization. While the map types employed in most single-robot systems are also applicable to multi-robot systems, a key distinction is that in multi-robot systems, the global map is constructed from data gathered by multiple robots, which requires a complex communication architecture. Accordingly, we will explore specific map types and their respective advantages and limitations in Section \ref{sec: comm}.\newline

\vspace{2mm}
\noindent\textbf{III-A. Multi-robot Localization}
\vspace{2mm}
\label{sec: loc}

During cooperative exploration tasks, each robot must estimate its own position, known as self-localization, while also tracking the states of other robots~\cite{swarm-lio}. These processes are referred to as ego-state estimation and relative state estimation, respectively.\newline

\vspace{2mm}
\noindent\textbf{III-A-1. Ego-State Estimation}
\vspace{2mm}

The ego-state estimation process of each robot can be viewed as the localization component of SLAM in a single-robot system. This is tackled differently depending on the given platform~\cite{Rouek2021SystemFM}. It can generally be divided into two methods based on sensor equipment: base-station-based localization and base-station-free localization. These methods can be used in combination.

Base-station-based localization determines a robot’s absolute position in a global coordinate system by measuring its relative distance or angle to fixed base stations. Common sensors used include global positioning systems (GPS), RTK-GPS, base-station-based ultra-wideband (UWB), motion capture systems, and georeferenced markers. These methods offer good applicability and accuracy but often have specific environmental requirements. Therefore, current multi-robot cooperative exploration tasks often use or combine base-station-free localization methods to address localization problems.

Base-station-free localization refers to the technique of estimating the robot's pose in its initial coordinate system by analyzing its current pose changes or environmental features, while integrating historical information. Its key feature is that it does not rely on any external fixed base stations, making it suitable for environments where GPS is unavailable or unstable, and offering excellent flexibility and adaptability. However, these methods can accumulate errors and are susceptible to external dynamic interference, leading to localization deviations. Therefore, techniques such as relocalization or loop closure detection are needed to reduce accumulated errors and improve localization accuracy. Additionally, a new class of activate loop-closing single-robot localization methods~\cite{Lee2021REAL} has emerged in recent years, which uses proactive loop closure based on probability to significantly improve pose estimation performance.\newline

\vspace{2mm}
\noindent\textbf{III-A-2. Relative State Estimation}
\vspace{2mm}
\label{sec: RelStateEstimation}

Relative state estimation between robots is a task unique to multi-robot systems. By utilizing local maps and understanding the relative states between robots, each robot can apply specific algorithms to avoid redundant exploration of the same areas, thus significantly improving mapping efficiency~\cite{yu2020mapmerging}. While methods such as visual-inertial odometry (VIO), lidar-inertial odometry (LIO), and GPS are well-established for ego-state estimation in single-robot systems, they either suffer from estimation drift or lack sufficient accuracy in multi-robot scenarios. Therefore, relying solely on ego-state estimation for absolute localization and using initial positions and ego-state comparisons to estimate relative states often leads to unsatisfactory results, which can compromise the safety and reliability of the multi-robot system. Incorporating relative state estimation between robots can effectively mitigate these issues. Based on the current literature, as summarized in Table \ref{tab:relstate}, relative state estimation methods can be broadly classified into environment-feature-based methods, relative-observation-based methods, and hybrid approaches. Additionally, among the methods mentioned above, multi-robot loop closure is a key technique commonly used for relative state estimation between robots, and therefore we provide a detailed explanation.

\textbf{a. Environment-feature-based Method}

Environment-feature-based methods estimate relative poses by leveraging environmental features, typically through the following steps. First, each robot gathers environmental data and extracts key features from various sensor modalities, such as visual keyframes, LiDAR point clouds, and others. Next, the key features are exchanged and compared either between robots or between robots and a server to identify common features within the environment, which are then used to estimate their relative states. Collaborative SLAM (C-SLAM) is a typical example of these methods, such as Door-SLAM~\cite{door-slam}, DCL-SLAM~\cite{dcl-slam}, and CCM-SLAM~\cite{ccm-slam}. These methods generally represent environmental features using feature descriptors~\cite{netvlad, scancontext} or visual vocabulary indexes~\cite{cieslewski2017visualpr}. Then, a series of techniques such as multi-robot loop closure~\cite{dcl-slam}, outlier rejection~\cite{mangelson2018PCM}, and pose graph optimization (PGO)~\cite{Tian2019DistributedCC} are applied to obtain the relative states between robots. For example, D2SLAM~\cite{D2SLAM2024}  integrated two scenarios to enhance state estimation for aerial swarms. In the near-field scenario, it used visual overlap to obtain high-precision real-time local state estimation and relative state between UAVs; in the far-field scenario, where UAVs could not observe each other, it obtained globally consistent trajectory estimates through the PGO method.

\textbf{b. Relative-observation-based Method}

Relative-observation-based methods aim to determine relative states by directly observing surrounding robots. Common approaches include vision-based methods~\cite{omni-swarm,Niu2022qrtag,H-SwarmLoc2023Wang,TransformLoc2024Wang}, point-cloud-based methods~\cite{swarm-lio,Zhu2024swarm-lio2}, and UWB-based methods~\cite{omni-swarm,Zhou2022flyinwild}. 
Vision-based and point-cloud-based methods identify surrounding robots using visual cameras or LiDAR. 
For example, Omni-Swarm~\cite{omni-swarm} used Yolov4-tiny~\cite{bochkovskiy2020yolov4} for teammate recognition; Niu et al.~\cite{Niu2022qrtag} pre-placed QR codes on UGVs for UAV recognition and localization; and Swarm-LIO~\cite{swarm-lio} equipped UAVs with reflective tapes to identify teammates via reflectivity of point cloud. 
However, when multiple nearby teammates are present, these methods need to combine additional techniques to determine the ID numbers of each teammate, such as comparing robot tracking data with the trajectories of all nearby robots~\cite{swarm-lio,omni-swarm}. 
Another limitation of these methods is their reliance on environmental visibility. They fail when obstacles cause occlusions, and factors like water mist and dust can further degrade accuracy or clarity~\cite{cave2021petracek}.
UWB-based methods utilize base-station-free UWB technology to measure the distance between the transmitter and receiver. However, due to the low accuracy and only one-dimensional distance measurements, UWB is typically used as a supplementary source to enhance relative pose estimation~\cite{omni-swarm,Zhou2022flyinwild}.
For example, Zhou et al.~\cite{Zhou2022flyinwild} developed a decentralized drift correction algorithm to minimize relative distance errors in onboard UWB measurements.

\textbf{c. Hybrid Method}

Hybrid methods combine the approaches mentioned above or integrate them with additional information. These methods leverage the strengths of multiple techniques to overcome the limitations of individual approaches, improving the reliability and accuracy of relative state estimation. For example, Nguyen et al.~\cite{Nguyen2022virloc} combined UWB-based relative observations with additional odometry data from teammates, creating a localization solution that did not require detecting loop closures in robot trajectories. EGO-Swarm~\cite{ego-swarm} used the observed target position from VIO and the predicted position from the controller to address VIO drift, thereby estimating the relative state. Omni-Swarm~\cite{omni-swarm} combined vision- and UWB-based relative observation methods with environment-feature-based approaches, overcoming issues such as observability, initialization complexity, insufficient accuracy, and global consistency.

\begin{table*}
    \centering
    \begin{tabular}{ccc}
    \toprule
         \multicolumn{2}{c}{Environment-feature-based Methods} &~\cite{door-slam, dcl-slam, ccm-slam, omni-swarm, D2SLAM2024}\\ 
         \midrule
         \multirow{3}{*}{Relative-observation-based Methods} & Vision-based &~\cite{omni-swarm, Zhou2022flyinwild, ego-swarm}\\ \cline{2-3}
         & Point-cloud-based &~\cite{swarm-lio,Zhu2024swarm-lio2}\\ \cline{2-3}
         & UWB-based &~\cite{omni-swarm, Zhou2022flyinwild}\\ 
         \midrule
        \multicolumn{2}{c}{Hybrid Methods} &~\cite{omni-swarm, Zhou2022flyinwild}\\ 
    \bottomrule
    \end{tabular}
    \caption{Classification of common relative state estimation methods for robots.}
    \label{tab:relstate}
\end{table*}

\textbf{d. Multi-robot Loop Closure}

Multi-robot loop closure is an important and effective approach for estimating the relative pose between robots by detecting loop closures in their relative pose trajectories without the need for a base station~\cite{dcl-slam}. This method not only compensates for odometry drift~\cite{omni-swarm, dcl-slam}, but also integrates the trajectories of individual robots into a common reference frame, improving the accuracy of trajectory estimation~\cite{dcl-slam, door-slam}.

Whether in centralized or decentralized multi-robot systems, feature matching-based methods are commonly used. This method employs descriptors or visual bags of words extracted from sensor data, such as visual keyframes or LiDAR point clouds, to exchange and compare information for detecting potential loop closures through communication between robots or between robots and servers. The communication mechanism involved in this data exchange across a multi-robot system will be discussed in Section \ref{sec: comm}. Such methods have been widely adopted by many multi-robot systems~\cite{ccm-slam, door-slam, omni-swarm, dcl-slam}. 

Descriptors can be broadly categorized into two types: traditional feature-based descriptors and learning-based descriptors~\cite{netvlad, SegMap}. The former can be further divided into local descriptors~\cite{localdescriptor1999Lowe, localdescriptor2015Prakhya} and global descriptors~\cite{LiDARIris2020Wang, scancontext}. Local descriptors are typically detected using methods such as SIFT~\cite{SIFT2016} and ORB~\cite{ORB2011}, which extract information from the local neighborhood of visual features or keypoints. However, due to their limited descriptive power, local descriptor methods are not suitable for inter-robot loop closure detection, though they can be used to validate putative loop closures and estimate relative pose in geometric verification~\cite{dcl-slam}. In contrast, global descriptors, which describe the overall properties of an image or an entire dataset, have superior descriptive power and are thus more widely applied. Furthermore, with the advancement of data-driven technologies, learning-based methods leverage large amounts of training data to train more powerful feature descriptors using convolutional neural networks, such as NetVLAD~\cite{netvlad} and SegMap~\cite{SegMap}. It is worth mentioning that in some scenarios, some learning-based descriptors~\cite{Uy2018PointNetVLADDP, Chen2020OverlapNetLC} may result in perspective differences and incomplete observations for 3D representations of the same area. To address this issue, AutoMerge~\cite{AutoMerge2023} proposed an attention-enhanced multi-view fusion descriptor to improve robustness under both translation and orientation differences simultaneously, and also integrated an adaptive loop closure detection mechanism in a dual-mode setting. 

The aforementioned methods directly employ descriptors, which require substantial communication bandwidth. Consequently, some methods have been proposed to reduce the amount of information exchanged. Tardioli et al.~\cite{Visualdata2015Tardioli} proposed using visual vocabulary indexes instead of feature descriptors. Other methods stored regions of full-image descriptors~\cite{FullImageDescriptors2017Cieslewski} or a pretrained visual bag of words~\cite{DistributedInvertedIndex2017Cieslewski} locally on each robot. These methods minimized the required bandwidth and could be scaled according to the number of robots, but they were designed for fully connected teams~\cite{door-slam}. All of these can be considered feature matching-based matching methods.

In addition to feature matching-based approaches, some multi-robot systems used an agent or base station to receive all the data needed for loop closure detection, including keyframes~\cite{ccm-slam} and maps~\cite{LAMP2, 3DLiDARSonlineMRSLAM2017Dubé}, and detected loop closures between robots by comparing and merging local maps or map data, which is more commonly seen in centralized systems. In other words, these methods usually directly transmit the local maps generated by each robot to a server for centralized computation, which requires high computational capacity and reliable communication.\newline

\vspace{2mm}
\noindent\textbf{III-B. Multi-robot Mapping}
\vspace{2mm}
\label{sec: map}

Cooperative mapping in multi-robot systems refers to the process where multiple robots collaborate to perceive their environment and merge the data collected by each into a single, consistent global map. This process is a core element of multi-robot exploration, particularly in coverage tasks. Each robot continuously updates its local map, a process known as individual mapping, while also integrating the maps of other robots. The update of each robot's local map follows the same principles as single-robot SLAM, but the key distinction in multi-robot systems lies in the integration of these local maps into a unified global map.

Map merging, or map fusion, involves combining sensor data or local map information from different robots to create a cohesive global map. While the map representations used in single-robot systems are generally applicable in multi-robot systems, the latter offers the added capability of fusing multiple local maps to form a larger global map. This not only facilitates navigation but, when combined with efficient exploration algorithms, can greatly enhance cooperative efficiency by reducing redundant exploration and optimizing task allocation to high-priority areas.

Map merging typically involves three key steps: information exchange, coordinate alignment, and data fusion. Information exchange refers to the sharing of sensor data or local maps between robots. This process depends on the system communication architecture discussed in Section \ref{sec: comm}, affecting whether robots share a global map and if it is consistent across robots. Coordinate alignment aims to align the coordinate systems of robots to ensure global map consistency. When the initial positions of robots are known in the global coordinate system, ego-state estimation can align their positions. If unknown, alignment can be achieved by calculating coordinate transformations between local and global maps, by comparing historical movement trajectories~\cite{swarm-lio,ccm-slam,Zhu2024swarm-lio2,H-DrunkWalk2020Chen,DrunkWalk2015Chen} or observations~\cite{zhang2022mr,Dong2022mr-gmm,MUITARE2023,LVCPLT2024Jian}. The specific methods for coordinate alignment are tied to multi-robot localization, as discussed in Section \ref{sec: RelStateEstimation}. We will focus on data fusion tasks.

Data fusion is a critical challenge in map merging, involving the merging or stitching of local maps or sensor data from multiple robots based on an aligned coordinate system. This process becomes even more complex when the robot swarm includes different types of robots or sensors, which poses significant challenges for map merging in multi-robot systems. In such cases, robots may produce maps of the same type but with varying attributes, such as size or accuracy, or even entirely different types of maps. Merging maps under these conditions, characterized by perceptual and sensor heterogeneity, is referred to as heterogeneous map merging. However, solutions for commonly used map types in heterogeneous scenarios remain limited~\cite{Andersone2019heterogeneousmap}. In contrast, when all robots generate maps of the same type and with similar attributes, the process is referred to as homogeneous map merging, which typically involves fewer complexities.

\textbf{a. Homogeneous Map Merging Method}

Homogeneous map merging methods have been widely explored in the literature, with various techniques developed to handle occupancy grid maps, topological maps, and Gaussian mixture models (GMMs). For occupancy grid maps, Andrew~\cite{Andrew2006particle} proposed using particle filters to intuitively compute new occupancy probabilities for each cell in the global map, based on state information between robots and the occupancy probabilities in their local grids. Similarly, Li and Nashashibi~\cite{Li2012gridmerge} employed a genetic algorithm to optimize an occupancy likelihood-based objective function, ensuring consistency in the fusion of occupancy grid maps across multiple vehicles. For topological maps, Zhang et al.~\cite{zhang2022mr} used visual feature descriptors to represent vertices in each robot's local topological map, computing similarity through the inner product of descriptor vectors and merging vertices with high similarity. Meanwhile, Dong et al.~\cite{Dong2022mr-gmm} addressed GMM submap fusion using an adaptive model selection strategy to dynamically merge similar Gaussian components, facilitating efficient fusion across robots. Yu et al.~\cite{yu2020mapmerging} provided a detailed classification of fusion methods for the three main map types—occupancy grid maps, feature-based maps, and topological maps—offering a comprehensive overview of their applications and limitations. AutoMerge~\cite{AutoMerge2023} solved the issues of relative perspective differences and temporal differences, and proposed a framework capable of merging multi-robot 3D maps at scales larger than 100 km.

\textbf{b. Heterogeneous Map Merging Method}

Heterogeneous map merging remains more challenging and algorithms are typically developed for specific scenarios. Lin et al.~\cite{Lin2021gridmerge} framed the fusion of grid maps with differing resolutions as a point set registration problem, incorporating scale information into a context-based descriptor to achieve consistent fusion. Gawel et al.~\cite{Gawel2016vision-laser} used a structural descriptor that captured the surrounding context of key points to match lidar point cloud maps from LIO with sparse visual keypoint maps from VIO, enabling cross-sensor map merging. The CSIRO Data61 team~\cite{CSIRO2022Heterogeneous} deployed a heterogeneous team composed of both aerial and ground robots, where each robot received information from others via point-to-point communication, using a common starting area as an overlap reference to build their global maps easily. This approach was validated during the DARPA Subterranean Challenge. Andersone~\cite{Andersone2019heterogeneousmap} provided an extensive review of state-of-the-art methods for both homogeneous and heterogeneous map merging, emphasizing the need for further advancements in heterogeneous scenarios.

Additionally, it is important to note that while explainable maps with visual, geometric, or semantic representations are essential in certain application scenarios, they are not universally required. For example, some cooperative localization tasks can be effectively accomplished using feature maps alone, as demonstrated in~\cite{Lajoie2021TowardsCS}. Similarly, map merging is not always necessary in tasks such as target tracking, where robots primarily exchange trajectory information and rely on local maps for obstacle avoidance and tracking, without the need for map data exchange~\cite{Zhou2022flyinwild}. These examples highlight that the choice of map type and the decision to perform map merging should be guided by the specific requirements of the task and the available resources.

\section{Multi-robot Cooperative Planning}
\label{sec: coop}

In single-robot exploration, the planning design primarily focuses on generating targets to gain new information about the unknown environment and planning optimal trajectories to connect these targets with the robot’s current position~\cite{cimurs2021goal}. Beyond this, multi-robot cooperative exploration introduces an additional component. After generating targets, the component allocates these targets to individual robots, aiming to balance workloads and avoid redundant exploration.
The robots then plan subsequent trajectories based on these assigned targets~\cite{vincent2008distributed, alitappeh2022multi}. 
Such a process in traditional multi-robot cooperative exploration can be summarized in three stages: generating targets, task allocation~\cite{sariel2006efficient}, and motion planning. 
We refer to this hierarchical approach as multi-stage planning, a major branch of cooperative planning.

\begin{figure}[h]
    \centering
    \includegraphics[width=\linewidth]{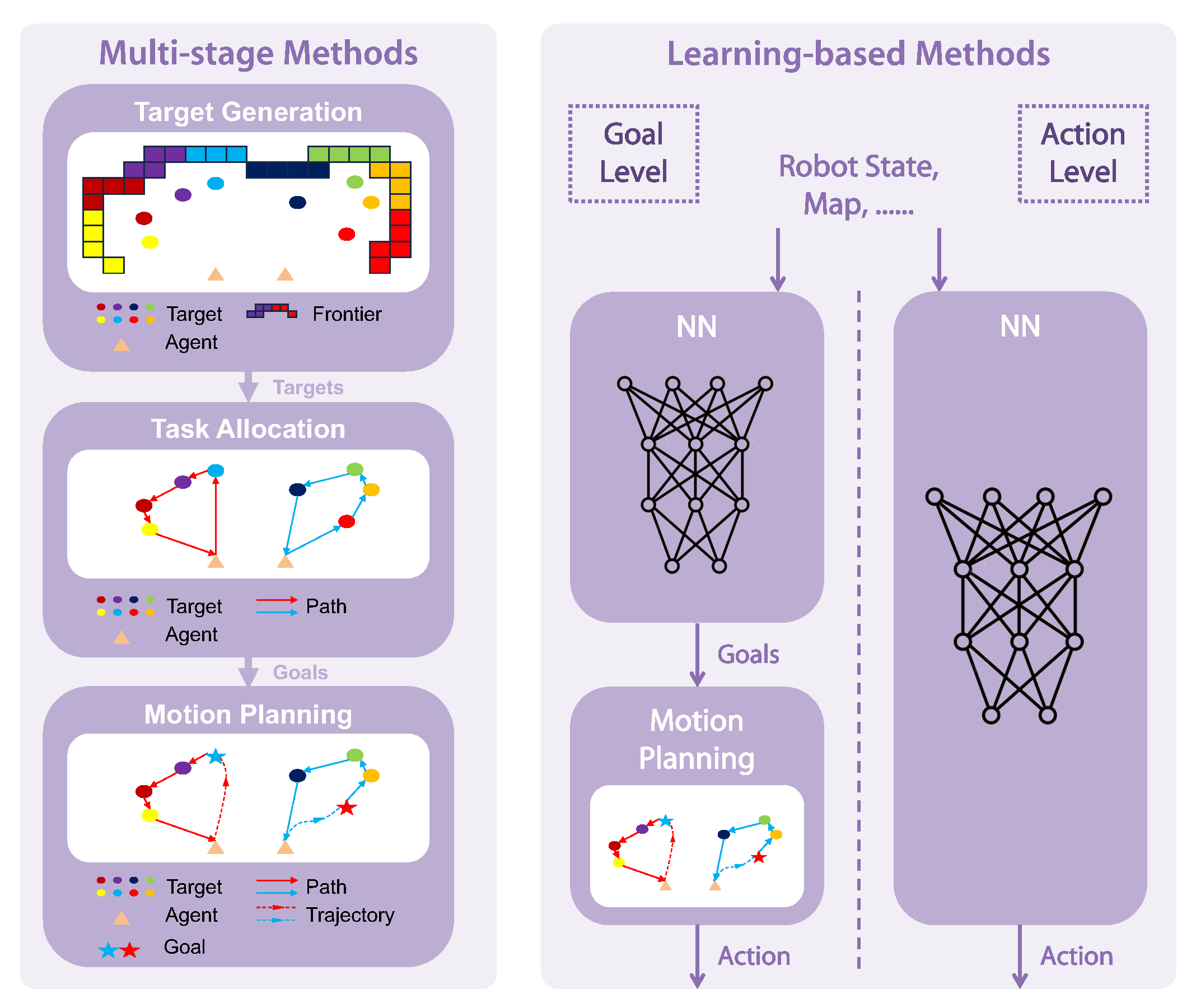}
    \caption{Comparison of two planning methods in the cooperative planning module: the multi-stage method and the learning-based method. The multi-stage method consists of three stages: target generation, task allocation, and path planning. In contrast, the learning-based method directly maps observations to actions and can be further categorized into goal-level and action-level approaches.}
    \label{fig:motion}
\end{figure}

With the development of RL, some researchers have explored learning-based planning approaches~\cite{chen2019end, MAANS, liang2024hdplanner}. These methods aim to directly generate motion commands from the map information, avoiding the complex hierarchical structure with gaps between modules. Fig.~\ref{fig:motion} illustrates the concepts of multi-stage planning and learning-based planning methods. 

This section introduces the development of the cooperative planning module, discussing multi-stage and learning-based planning methods in detail and offering insights into current trends and potential improvements.

\vspace{2mm}
\noindent\textbf{IV-A. Multi-stage Planning Methods}
\vspace{2mm}

Multi-robot systems require complex coordination among robots, creating a highly intricate problem space. Multi-stage planning addresses this complexity by breaking down the overall task into simpler sub-tasks, enhancing both efficiency and accuracy. Additionally, this approach allows for periodic reevaluation of environmental conditions, enabling adjustments to the plan that accommodate changes and specifically cater to the demands of multi-robot 
cooperative exploration tasks.

\vspace{2mm}
\noindent\textbf{IV-A-1. Target Generation}
\vspace{2mm}
\begin{figure*}[ht]
    \centering
    \includegraphics[width=\linewidth]{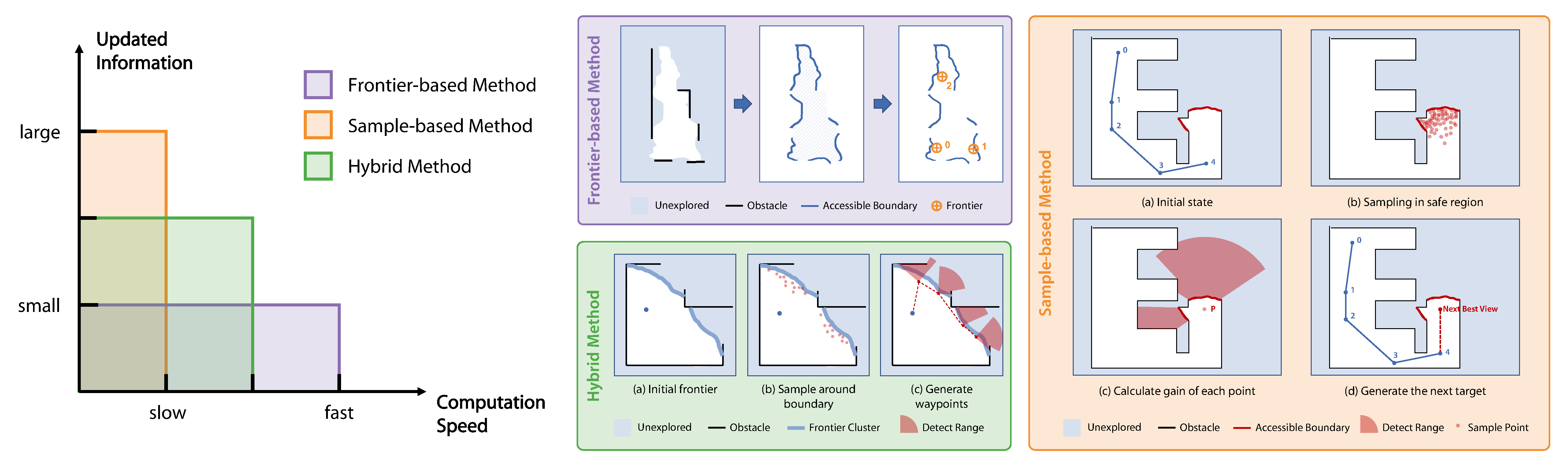}
    \caption{Comparison of different target generation methods. The figure illustrates how each method generates targets, with a focus on computation speed and updated information. The frontier-based method exhibits faster computation speed but provides less updated information, while the sample-based method offers more updated information at the cost of slower computation speed. The hybrid method strikes a balance between the two, with computation speed and updated information lying between the frontier-based and sample-based methods.}
    \label{fig:target-generation}
\end{figure*}

As the first stage of the multi-stage planning approach, target generation focuses on identifying targets that direct robots toward the boundary between known and unknown areas of the map for exploration. 

Existing specific target generation methods can be broadly classified into \textbf{\textit{frontier-based methods}}, \textbf{\textit{sample-based methods}}, and \textbf{\textit{hybrid methods}} that combine or improve both approaches. Fig.~\ref{fig:target-generation} provides a comparison of different target generation methods and illustrates how each method generates targets. For target generation methods, we design two metrics: one is \textit{computation speed}, which reflects the computational resources required for the method to generate targets; the other is \textit{updated information}, which represents the number of map updates that can be obtained from the targets generated by the method. The frontier-based method can generate targets directly from the frontier cluster based on indicators such as the shortest Euclidean distance, thereby achieving extremely fast computation speed. However, it may also result in less updated information. The sample-based method, by randomly sampling near the boundary between explored and unknown areas to calculate the detect range, can target a higher amount of updated information, but it may lead to excessive computational loads, thereby reducing the computation speed. The hybrid method combines the aforementioned two methods, maintaining a certain amount of updated information while possessing fast computation speed.

Additionally, here is another kind of method referred to as \textbf{\textit{segmentation-based methods}}. The methods focus on dividing the map into several subregions using specific rules. Each subregion is then assigned to different robots for exploration. While segmentation-based methods don’t directly create the targets for motion planning, they essentially produce larger objectives (the subregions within the exploration zone). For this reason, we consider them as one of the methods for target generation.

\textbf{a. Frontier-based Method}

The frontier-based exploration method originated from the work of Yamauchi~\cite{yamauchi1997frontier}. In his initial formulation, a ``frontier'' refers to the boundary between explored and unexplored regions within a given exploration range. This boundary is generally accessible to the robot, meaning that it excludes inaccessible points such as obstacles or walls. To determine targets for the robot’s movement, Yamauchi proposed that any explored, open cell adjacent to an unexplored cell be labeled as a frontier edge cell, and that adjacent frontier edge cells collectively form a frontier region. Any frontier region exceeding a certain size is then defined as a frontier (an example shown in Fig.~\ref{fig:target-generation}).

By this definition, frontier-based exploration (FBE) relies on two key properties of the map: \textit{divisibility} and \textit{continuity}. \textit{Divisibility} is needed to clearly distinguish between explored and unexplored areas, as well as to identify boundaries that are reachable or unreachable, in order to define the frontiers. \textit{Continuity} is necessary to ensure that the robot can generate a feasible path on the map, thereby determining whether a given frontier is usable.

Yamauchi~\cite{yamauchi1998frontier} later extended his approach to cooperative multi-robot exploration, where all robots share a global map and operate independently. However, this work only considers a simple mechanism that assigns a frontier to the nearest robot, leading to redundant exploration. This issue is essentially related to task allocation, inspiring extensive subsequent research. Based on the frontier-based method, numerous task allocation methods have been proposed to facilitate efficient collaboration. We will provide an in-depth introduction to these task allocation methods in Section IV-A-2.

\textbf{b. Sample-based Method}

In robot exploration tasks, sample-based methods heavily rely on the next-best-view (NBV) method~\cite{connolly1985determination} in practical implementations, and therefore, the two are often considered the same. The NBV method in the field of robot exploration was first introduced by González-Baños et al.~\cite{gonzalez2002navigation}. They defined the concept of a safe region, which is the largest area where the robot can guarantee the absence of obstacles based on sensor data. They then randomly generated a set of potential candidate positions within the current safe region, which the robot could reach via a collision-free path. Each candidate position was evaluated based on information gain, map overlap, and movement cost, with the highest-scoring position selected as the robot’s next target point for movement (an example shown in Fig.~\ref{fig:target-generation}).

Other studies may employ NBV algorithms that differ in specific processes, but the basic steps of collision checking, random point selection, gain calculation, and target point determination are largely similar. Although NBV algorithms are efficient in updating the map with new information, they involve significant computational overhead due to the need for random point selection and gain calculation, and they require high-performance sensors. Therefore, in large-scale environments, simple NBV algorithms may face challenges in completing the task.

\textbf{c. Hybrid Method}

Due to the simplicity of computation in frontier-based methods, which, however, cannot guarantee the updating of large amounts of information, and the extensive information acquisition but the heavy computational burden of sample-based methods, some researchers have attempted to combine the two to leverage their respective strengths. The goal is to generate movement targets for robots that can acquire a significant amount of updates under a certain computational load. In this survey, they are referred to as hybrid methods (an example shown in Fig.~\ref{fig:target-generation}).

Charrow et al.\cite{charrow2015information} use a frontier-based method to generate frontiers, then cluster these frontiers, selecting the path that can reach and observe the cluster the fastest, while sampling the robot’s control space to obtain local short paths. Meng et al.\cite{meng2017two} use a frontier-based sampler to select a set of candidate viewpoints from the map, calculate the cost matrix between the candidate viewpoints based on Euclidean distance and information gain, and generate an approximate optimal path covering all candidate viewpoints by solving the fixed-start open traveling salesman problem (FSOTSP). Zhou et al.~\cite{zhou2023racer} proposed the concept of Frontier Information Structure (FIS), which stores information about all known free voxels and their adjacent unknown voxels in the exploration space, grouping them into frontier clusters. Each time the map is updated, it checks for the appearance of new frontier clusters or the disappearance of old ones. Each frontier cluster contains all its voxels, average position, axis-aligned bounding box (AABB), and candidate viewpoints (VPi), which are uniformly sampled points in cylindrical coordinates and optimized based on the sensor model to maximize the coverage of the frontier cluster. Subsequently, a graph search method considers multiple viewpoints for each frontier cluster and selects the best combination to further improve exploration efficiency.

\textbf{d. Segmentation-based Method}

\begin{figure}
     \centering
     \includegraphics[width=\linewidth]{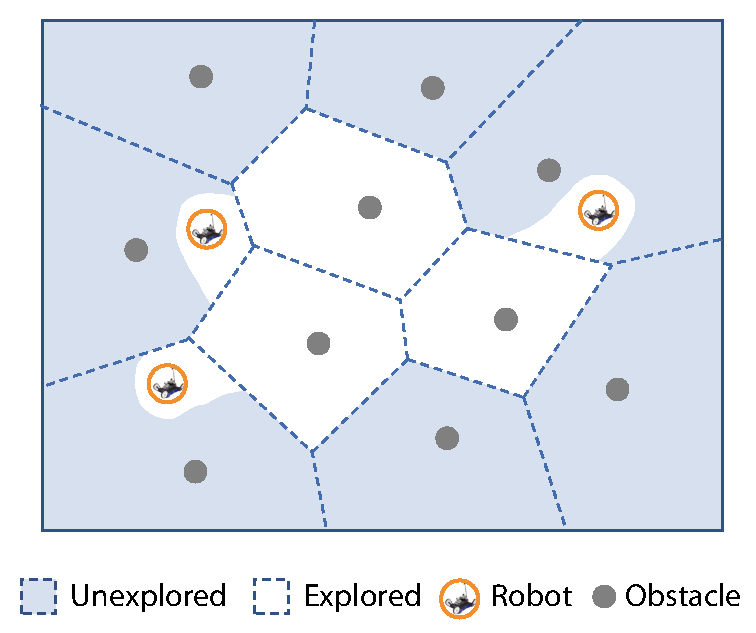}
     \caption{An example of segmentation-based methods dividing the map into subregions.}
     \label{fig:seg}
\end{figure}

Segmentation-based methods differ from the aforementioned approaches by not creating direct targets but focusing on dividing the map into subregions according to certain rules, such as Voronoi diagrams~\cite{aurenhammer1991voronoi} and clustering algorithms~\cite{xu2005survey} (an example shown in Fig.~\ref{fig:seg}). Besides, the process of generating specific targets and the segmentation-based method are not sequentially dependent. One approach is to first generate specific targets and then segment them using clustering algorithms. Alternatively, one can first carry out Voronoi diagrams, allocate subregions to individual robots, and then generate specific targets.

Wurm et al.~\cite{wurm2008coordinated} employed Voronoi diagrams to partition the exploration area into multiple sub-regions by identifying critical points in the diagram to determine the subregion boundaries. Subsequently, the Hungarian algorithm was used to assign robots to these subregions instead of individual targets, ensuring a more uniform robot distribution in the environment. Similarly, Karapetyan et al.~\cite{karapetyan2017efficient} proposed a clustering algorithm inspired by breadth-first search (BFS)~\cite{bundy1984breadth} to group decomposed regions and assign them to robots. For each region, they applied the coverage path planning (CPP)~\cite{galceran2013survey} algorithm to calculate the optimal path for complete coverage.
Dong et al.~\cite{dong2024fast} introduced the concept of multi-robot dynamic topological graph and performed graph Voronoi partition, addressing the unreasonable load distribution caused by conventional space-based Euclidean Voronoi partition, which ignores obstacles. Zhang et al.~\cite{zhang2024leces} tackled the issue of load balancing between robots by utilizing a variant of the K-means algorithm~\cite{likas2003global} to cluster all targets and assign them to robots. They further applied the asymmetric traveling salesman problem (ATSP)~\cite{oncan2009comparative} method to navigate robots through their assigned target clusters.

Segmentation-based methods facilitate workload distribution among robots, mitigate redundant exploration, and decompose multi-robot exploration tasks into a series of independent single-robot subproblems, thereby enhancing scalability and computational efficiency in large-scale environments~\cite{dong2024fast}. However, by effectively reducing the problem dimensionality, these methods may compromise global path optimality. Moreover, in environments with uneven complexity, naive segmentation strategies can result in imbalanced task allocation, leading to inefficiencies in exploration.

\vspace{2mm}
\noindent\textbf{IV-A-2. Task Allocation}
\vspace{2mm}

Once targets are generated, the system allocates them to robots following specific rules. Robots then plan and execute paths to reach these targets. As exploration progresses, target generation and task allocation are updated dynamically.

Gerkey et al.~\cite{gerkey2004formal} classify task allocation in multi-robot cooperative exploration systems as the ``Single-task robots, multi-robot tasks, time-extended assignment'' (ST–MR–TA) type. This task allocation problem is NP-Hard, meaning no unified optimal solution exists at present. To address this challenge, researchers have introduced various constraints to simplify the problem, leading to improved solutions under specific scenarios. Based on the approaches employed in the current literature, existing task allocation methods can be categorized into iterative methods, metaheuristic methods, market-based and auction-based methods, TSP-based methods, and RL-based methods. Table~\ref{tab:task_allocation} provides a concise overview of the strengths and limitations of various task allocation methods.

Besides, it’s important to understand the difference between centralized and distributed methods in task allocation. In centralized methods, a single central decision-making unit holds all the information about the robots and their environment, making all the decisions and planning. On the other hand, distributed methods involve placing decision-making units on each robot, allowing them to make their own decisions while having access to information about other robots and the environment~\cite{lajoie2022towards}.

As the size of the area to be explored and the number of robots increase, centralized methods face challenges. They struggle with the communication and computational load, which slows down the process of transmitting planning results to all robots. This can significantly reduce the efficiency of exploration. That’s why, in large-scale environments, the trend is shifting towards using distributed methods for better performance.

\textbf{a. Iterative Method}

The iterative method primarily involves iteratively updating the utility of targets and continuously selecting the optimal movement target for each robot. 

Burgard et al.~\cite{burgard2005coordinated} proposed a method based on frontiers, where for each robot, the cost of reaching each frontier cell is calculated, and the target with the best utility is selected for the robot through iterative updates. 
Butzke et al.~\cite{butzke2011planning} introduced a method that considers multiple objective factors such as information gain, regional bias, distance bias, and repeat rewards to calculate the utility of targets generated based on the frontier-based method, and the robot-target pair with the highest score is selected for target allocation. Huang et al.~\cite{Huang2024Autonomous} generated targets based on the hybrid method and constructed a utility function considering map uncertainty, task allocation efficiency, and target point distance to iteratively calculate the optimal target point for robots. 

While the iterative method effectively assigns a single target point to robots, it struggles when a single robot needs to handle multiple targets simultaneously. Consequently, this method often yields better performance when combined with other approaches, rather than being used in isolation.

\textbf{b. Metaheuristic Method}

The metaheuristic method employs heuristic algorithms for task allocation and motion planning, simulating the cooperation and competition observed in biological populations to guide robot exploration behavior. Common techniques include 
swarm intelligence algorithms and potential field-based methods.

Wang et al.~\cite{wang2011frontier} proposed a multi-robot map exploration algorithm based on PSO. Initially, the system divides the exploration area into multiple subregions and assigns them to robots. Each robot iteratively selects the nearest frontier point within its subregion and plans its movement path using the PSO algorithm. Once all frontiers in a subregion are explored, the PSO algorithm reassigns the robot to another subregion that contains unexplored frontiers and is not assigned to other robots.
Andries et al.~\cite{andries2015multi} developed an exploration approach based on an ant-inspired algorithm referred to as Brick\&Mortar Improved Long Range Vision (BMILRV), where robots operate independently without direct communication or shared exploration goals. Each robot makes decisions based on local information within its field of view, leaving traces during exploration. These traces guide other robots to avoid already explored areas and focus on unexplored regions. 
Yu et al.~\cite{yu2021smmr} introduced the multi-robot multi-target potential field (MMPF) method, which determines targets for robots based on factors like information gain and travel cost. Additionally, this method incorporates a repulsive potential field for each robot to ensure they explore distinct areas and avoid interference with one another.

A key limitation of inspiration-based metaheuristic methods is that they can provide feasible solutions for task allocation and exploration but do not guarantee these solutions are globally optimal. While these methods are capable of completing exploration tasks, the lack of optimality may result in additional resource consumption and extended time requirements. Nevertheless, their ability to generate feasible solutions in complex and dynamic scenarios highlights their practical value in multi-robot exploration systems.

\textbf{c. Market-based and Auction-based Method}

Market-based and auction-based methods for multi-robot cooperative exploration leverage concepts from economics, particularly market mechanisms and auction theory, to coordinate the behavior of robots during exploration tasks. These methods aim to optimize resource allocation by distributing resources to those robots or tasks that need them most or can provide the highest value under defined rules.

Zlot et al.~\cite{zlot2002multi} proposed a method where targets are generated using the sample-based method and inserted into robots’ tour paths. The paths are then optimized based on the principle of cost minimization. Robots subsequently auction these targets to others, allowing robots to bid based on their own circumstances, with the highest bidder acquiring the target.
Berhault et al.~\cite{berhault2003robot} addressed the limitations of single-target auctions, which fail to account for the synergistic effects between targets, by proposing a combinatorial auction method. They introduced various combination strategies and demonstrated that combinatorial auctions significantly reduce travel distances compared to single-target auctions.

Market-based and auction-based methods enable systems to adapt to dynamically changing environments and tasks, as they eliminate the need for a central control point, thereby reducing the risk of a single point of failure. However, these mechanisms may not ensure fair task allocation opportunities for all robots. Robots might engage in strategic behavior to maximize their individual benefits, potentially leading to a decrease in overall efficiency.

\textbf{d. TSP-based Method}

The traveling salesman problem (TSP)~\cite{gutin2006traveling} is a well-known combinatorial optimization problem that seeks the shortest Hamiltonian circuit within a given set of cities—a path that visits each city exactly once and returns to the starting city. Faigl et al.~\cite{faigl2012goal} characterized the task allocation problem in multi-robot cooperative exploration as the multiple traveling salesman problem (MTSP). MTSP extends TSP by routing multiple robots through multiple targets while optimizing for the shortest path or other criteria. This formulation has allowed numerous methods originally designed for TSP and its variants~\cite{applegate2006concorde,helsgaun2000effective,helsgaun2017extension} to be adapted in multi-robot systems~\cite{hardouin2020next}.

Faigl et al. \cite{faigl2012goal} introduced a method that applies a variant of the K-means clustering algorithm to group targets into clusters, which are then allocated to individual robots based on the traveling salesman problem (TSP) distance cost. Zhou et al.\cite{zhou2023racer} proposed the RApid Collaborative ExploRation (RACER) system, which leverages hgrid decomposition and pairwise interaction to incorporate connection costs and capacity constraints, following the capacitated vehicle routing problem (CVRP) formulation~\cite{dantzig1959truck}. This approach aims to minimize the total coverage path (CP) length while ensuring a balanced distribution of unexplored space among drones. The robots then use the CVRP-generated CPs as high-level guidance to plan their paths from their current viewpoints to the centers of subsequent CP cells.

TSP-based methods are capable of finding the shortest or near-shortest paths to visit all targets, balancing tasks among multiple robots, and extending to scenarios where multiple targets are assigned to multiple robots simultaneously. These features make them suitable for large-scale area exploration~\cite{zhou2023racer,bartolomei2023fast}. However, as these methods aim to solve NP-hard problems, the computational complexity increases dramatically with the number of targets~\cite{chauhan2012survey}, making real-time updates in dynamic environments challenging.

\begin{table*}[ht]
\centering

\begin{tabular}{l|p{2cm}|p{10cm}}
\toprule
\multirow{4}{*}{Iterative} 
& \textbf{Key Idea} & Iteratively update target utility and allocate tasks based on optimized target selection. \\ \cline{2-3}
& \textbf{Strengths} & Well-suited for single-target allocation; effectively reduces redundant exploration. \\ \cline{2-3}
& \textbf{Limitations} & Inefficient for multi-target allocation; often requires integration with other methods. \\ \cline{2-3}
& \textbf{References} & \cite{burgard2005coordinated}, \cite{butzke2011planning}, \cite{Huang2024Autonomous} \\ \hline

\multirow{4}{*}{Metaheuristic} 
& \textbf{Key Idea} & Utilize bio-inspired algorithms (e.g., Particle Swarm Optimization, Ant Colony Optimization) to iteratively optimize task allocation. \\ \cline{2-3}
& \textbf{Strengths} & Adaptable to dynamic environments and complex exploration tasks. \\ \cline{2-3}
& \textbf{Limitations} & No guarantee of global optimality; potential inefficiency in resource utilization. \\ \cline{2-3}
& \textbf{References} & \cite{wang2011frontier}, \cite{andries2015multi}, \cite{yu2021smmr} \\ \hline

\multirow{4}{*}{Market/Auction-based} 
& \textbf{Key Idea} & Employ market-driven approaches, such as auctions, to dynamically allocate tasks among robots based on bidding strategies. \\ \cline{2-3}
& \textbf{Strengths} & Enables decentralized decision-making and adapts well to dynamic environments. \\ \cline{2-3}
& \textbf{Limitations} & Susceptible to strategic bidding behavior, which may compromise fairness and efficiency. \\ \cline{2-3}
& \textbf{References} & \cite{zlot2002multi}, \cite{berhault2003robot} \\ \hline

\multirow{4}{*}{TSP-based} 
& \textbf{Key Idea} & Formulate task allocation as a Multi-Traveling Salesman Problem (MTSP) or Capacitated Vehicle Routing Problem (CVRP) to optimize task assignment and path planning. \\ \cline{2-3}
& \textbf{Strengths} & Effectively balances task distribution and minimizes total travel distance. \\ \cline{2-3}
& \textbf{Limitations} & Computationally expensive, making real-time applications challenging. \\ \cline{2-3}
& \textbf{References} & \cite{faigl2012goal}, \cite{zhou2023racer},\cite{hardouin2020next} \\ \hline

\multirow{4}{*}{RL-based} 
& \textbf{Key Idea} & Utilize reinforcement learning (RL) for adaptive task allocation in either centralized or decentralized frameworks. \\ \cline{2-3}
& \textbf{Strengths} & Capable of handling complex strategies with minimal inference overhead after training. \\ \cline{2-3}
& \textbf{Limitations} & High training complexity; centralized approaches face scalability challenges due to reliance on global information. \\ \cline{2-3}
& \textbf{References} & \cite{yang2023learning}, \cite{zhang2022centralized}, \cite{marchesini2021centralizing}, \cite{luo2019multi}, \cite{zhou2019bayesian}, \cite{jin2019efficient}, \cite{elfakharany2021end}, \cite{paul2021learning}, \cite{lu2024reinforcement}, \cite{park2021cooperative}, \cite{NeuralCoMapping}, \cite{dou2015genetic} \\ 
\bottomrule
\end{tabular}
\caption{Overview of different task allocation methods, highlighting their strengths and limitations in multi-robot cooperative exploration.}
\label{tab:task_allocation}
\end{table*}

\textbf{e. RL-based Method}

Classical task allocation methods rely on prior knowledge and have limited expressiveness in modeling complex strategies. Additionally, they suffer from high computational costs and challenges related to hyperparameter tuning. Reinforcement learning (RL) has demonstrated significant potential in multi-robot task allocation~\cite{task_allo_rl1,task_allo_rl2,task_allo_rl3} due to its strong representational capabilities for complex strategies and low inference overhead once policies are well-trained.

RL-based task allocation methods can be broadly classified into centralized and decentralized approaches, depending on whether decision-making is independent. In the centralized setting, the RL policy~\cite{yang2023learning,zhang2022centralized,marchesini2021centralizing} receives global information from all agents and makes decisions for the entire robot team. In this scenario, the RL policy has the potential to achieve a globally near-optimal solution. However, centralized methods face scalability limitations, particularly in scenarios with agent failures or unstable communication, as they heavily depend on global information.
To address these challenges, recent research has increasingly focused on decentralized approaches~\cite{luo2019multi,zhou2019bayesian,jin2019efficient,elfakharany2021end,paul2021learning,lu2024reinforcement,park2021cooperative}, where each robot operates with an independently trained RL policy, making decisions based on partial observations. Decentralized methods are more robust to agent failures and communication instabilities, making them better suited for real-world multi-robot systems.

Among decentralized methods, NeuralCoMapping~\cite{NeuralCoMapping} introduced a multiplex graph neural network to predict the neural distance between frontier nodes and agents. Based on this neural distance, it assigned each agent a frontier node at every global step in the decentralized setting. To enhance flexibility in goal selection, Luo et al.~\cite{luo2019multi} used general graph nodes instead of frontier nodes, transforming the environment exploration problem into an abstract graph domain. A graph convolutional network (GCN) was then applied to allocate exploration targets to agents based on the graph structure.
Additionally, some studies incorporated prior knowledge from classical methods into reinforcement learning to improve policy learning during early training stages. For example, Dou et al.~\cite{dou2015genetic} introduced a hybrid approach for decentralized multi-robot task allocation and scheduling in intelligent warehouse environments, combining genetic algorithms for global scheduling optimization with reinforcement learning for task allocation.
Furthermore, to better handle real-world scenarios where unexpected disruptions could occur, some approaches~\cite{zhou2019bayesian,jin2019efficient} made decentralized decisions based on probabilistic models of the environment. For instance, Zhou et al.~\cite{zhou2019bayesian} modeled the uncertain environment as a Bayes-adaptive transition-decoupled partially observable Markov decision process (POMDP) and proposed a scalable decentralized online learning method by extending Monte Carlo tree search.

RL-based methods have demonstrated significant potential in task allocation due to their adaptability to diverse and complex scenarios. Compared to centralized approaches, decentralized RL methods enhance robustness against agent failures and communication uncertainties, as each robot operates with an independently trained policy. However, in large-scale multi-robot systems, RL-based approaches face challenges in efficiently exploring the vast solution space, often struggling to achieve near-optimal task allocations. Moreover, their generalization capability remains limited—these methods~\cite{dasari2020robonet,cai2024transformer,howell2024generalization} may fail in unseen scenarios if the deployment conditions deviate substantially from the training distribution.

\vspace{2mm}
\noindent\textbf{IV-A-3. Motion Planning}
\vspace{2mm}

After the multi-robot cooperative exploration system completes task allocation, all robots receive a series of targets. The final step in multi-stage planning — motion planning — aims to enable each robot to connect its assigned targets by generating feasible and safe trajectories. These trajectories must adhere to the platform's physical limitations, and environmental constraints, and ensure that robots satisfy the Field of View (FOV) constraints at different targets along the trajectory.

Most motion planning comprises two primary steps — path planning and trajectory optimization. The path planning (front-end) identifies a path that integrates geometric constraints, and determines the trajectory’s homotopy class, which will be discussed first in this subsection. Subsequently, we will discuss the trajectory optimization (back-end), refining the path to satisfy kinematic/dynamic constraints and control-input requirements. 
The aspects of this work related to single-robot motion planning are not the focus of this subsection. Here, we provide only a brief overview. For detailed information, please refer to~\cite{paden2016survey,liang2014review}. Our survey focuses primarily on multi-robot cooperative motion planning, emphasizing methodologies for coordinating trajectories, resolving inter-robot conflicts, and ensuring collective efficiency while adhering to dynamic constraints.

\textbf{a. Path Planning}

In multi-robot cooperative exploration systems, the objective of path planning is to generate collision-free paths within fully or partially known environments for all robots. These paths must connect the starting point of each robot to all targets allocated for them while adhering to state and FOV constraints at each target. They should also satisfy predefined performance metrics and adapt to both dynamic and static environmental conditions. Additionally, path planning can be further subdivided into local planning (for dynamic obstacle avoidance and real-time adjustments) and global planning (for long-term optimal path generation) in multi-robot cooperative exploration systems. Common path-planning algorithms can be divided into four main categories: search-based methods, sampling-based methods, heuristic-based methods, and learning-based methods.

Both search-based methods~\cite{xu2024Cost,hart1968,kim2023multi,bramblett2022coordinated,yu2021market} and sample-based methods~\cite{lavalle1998rapidly} are widely utilized in path-planning algorithms, but they are often hindered by the curse of dimensionality and struggle with dynamic or unknown obstacles~\cite{mac2016heuristic}. Heuristic-based approaches leverage empirical rules to simplify search processes and reduce computational complexity~\cite{warren1989global,dorigo2006ant}. Besides, learning-based methods, such as imitation learning (IL)~\cite{loquercio2021learning, li2020aggressive, li2018oil}, reinforcement learning (RL)~\cite{penicka2022learning, stachowicz2023fastrlap, zhang2024npe}, and both of them~\cite{song2023learning, xing2024bootstrapping,ross2011reduction}, are also applied to path planning.

There have been numerous studies in the field of Multi-Agent Path Finding (MAPF).~\cite{stern2019multi,sharon2015conflict,vcap2015prioritized,yu2013multi,contreras2017distributed,zhou2019robust,zhou2023racer,zhou2021raptor,zhou2020ego,zhou2021ego,burger2018cooperative}, addressing the efficiency of cooperative exploration and mitigate potential collision risks. Conflict-based search (CBS)~\cite{sharon2015conflict} is one of the most popular methods. This algorithm treats path conflicts as constraints and gradually resolves them through hierarchical search. Another representative is the prioritized planning algorithm~\cite{vcap2015prioritized}, which assigns an order to the robots and plans for each in turn. Each subsequent robot designs its trajectory with awareness of the previously planned paths, thereby avoiding collisions. The flow-based approach~\cite{yu2013multi} models multi-robot path planning as a network flow problem and obtains collision-free paths by solving it using linear programming or integer programming. Contreras et al.~\cite{contreras2017distributed} introduced a bee-inspired approach, which solves the problem of robot path conflicts by mimicking the way bees take turns obtaining food in nature, effectively reducing conflicts among robots and shortening task execution time. Li et al.~\cite{li2022tract} modeled the UAV swarm path planning problem as a policy matrix searching problem and implemented UAV swarm path planning by using the Simulated Annealing Algorithm (SA) to search for the optimal policy matrix. Chen et al.~\cite{chen2024soscheduler} designed a multi-UAV cooperative scheduling framework named SOScheduler, which employs a sequential allocation scheme and utilizes graph search algorithms to plan long-term paths for each UAV, making it suitable for large-scale environments.

It is important to note that, in multi-robot cooperative exploration systems, the objective of global planning is to allocate exploration regions for each robot, ensuring balanced workloads (avoiding overburdened or underutilized agents) and identifying optimal paths connecting all targets. This minimizes total travel distance and enhances exploration efficiency. Local planning, in contrast, focuses on generating collision-free trajectories from the robot’s current position to the next target. These trajectories prioritize optimal FOV alignment and optimal state at targets while avoiding obstacles and coordinating with other robots’ movements~\cite{zhou2023racer,dong2024fast,xu2024Cost}.

\textbf{b. Trajectory Optimization}

In practical multi-robot cooperative exploration tasks, the trajectories to be executed must adhere to the kinematic/dynamic constraints of all types of robots in the systems (e.g., velocity, acceleration limits) and be temporally discretized to enable precise tracking by low-level controllers. This ensures smooth motion, avoids actuator saturation, and guarantees real-time responsiveness in dynamic environments. To address this, trajectory optimization incorporates the robot’s dynamic characteristics and operational capabilities, producing paths that are more efficient, safer in real-world operation, and aligned with predefined optimization objectives. Broadly speaking, trajectory optimization can be categorized into hard-constraint~\cite{mellinger2011minimum,gao2018online,ding2019efficient,ren2022bubble,zhou2021decentralized,han2021fast,quan2022distributed,wang2022geometrically,tordesillas2021mader,tordesillas2022minvo} and soft-constraint approaches~\cite{gao2017gradient,zhou2019robust,usenko2017real,zhou2020ego}.

In current research, many works have extended trajectory optimization to multi-robot systems. Zhou et al.~\cite{zhou2021ego} extended their gradient-based replanning method~\cite{zhou2020ego} to a team of quadrotors, which has exploited the properties of B-splines and used topological guiding paths and active perception to achieve aggressive flight in complex scenes. Multi-robot trajectory optimization is also commonly used in autonomous driving. Burger et al.~\cite{burger2018cooperative} introduced a Mixed-Integer-Quadratic-Programming (MIQP) formulation for the motion planning of multiple vehicles, achieving efficient traffic flow through trajectory optimization.

\vspace{2mm}
\noindent\textbf{IV-B. Learning-based Planning Methods}
\vspace{2mm}

Multi-stage planning methods consist of three stages: target generation, task allocation, and motion planning. However, it is difficult to promise the execution of each stage achieves optimality, resulting in error accumulation across stages. The multi-stage planning methods have limited scalability to complex cooperative exploration scenarios since they lack a globally optimized strategy that accounts for interdependencies between stages. Besides, their sequential execution across stages introduces time delays, which is unfavorable in the real world. In contrast, learning-based approaches model the entire exploration process by training a single neural network to find a flexible strategy without prior knowledge. In multi-robot exploration tasks, robots need to adjust their strategy timely to better adapt to the changes of environments. RL policies are optimized via direct interaction with environments~\cite{chen2019end,yu2023asynchronous,yue2019reinforcement,chen2024ddl,chen2022deliversense}, which can learn a flexible strategy with a strong representation for complex and dynamic environments. Therefore, RL has emerged as a prominent paradigm in multi-robot exploration. 

Learning-based methods for multi-robot exploration can be categorized into two main types, as shown in Fig.~\ref{fig:motion}, based on the stages that are jointly optimized: goal-level methods and action-level methods. Goal-level methods generate a navigational goal for each robot, combining target generation and task allocation into a learning-based policy. Action-level methods, on the other hand, directly generate environmental actions for each robot, replacing all three stages with an end-to-end policy. The challenge in developing a learning-based policy for multi-robot exploration lies not only in optimizing policies over vast search spaces but also in addressing the credit assignment problem~\cite{wei2021multi}. This problem becomes particularly challenging when dealing with heterogeneous robots, as differences in capabilities require sophisticated strategies for workload distribution~\cite{liu2024heterogeneous}. Additionally, multi-robot exploration tasks are marked by partial observability and dynamic, uncertain environments, necessitating policies that can predict global observations and environmental dynamics based on limited sensor data.

\textbf{1) The Goal-Level Methods}

In recent advancements in multi-robot exploration, a novel approach has emerged that directly generates long-term navigational goals for each agent without relying on target generation and task allocation in multi-stage methods~\cite{liu2024heterogeneous,wang2023maddpg,liang2024hdplanner}. This approach allows for flexibility by generating goals that can be specific coordinates on the exploration map or latent states representing regions of interest. However, the exploration space expands to encompass the entire map rather than being confined to node candidates generated from target generation, making it more challenging for an RL policy to figure out a nearly optimal strategy. 

To address this challenge, several studies have proposed methods to narrow down the exploration space, thereby improving the efficiency and effectiveness of goal selection. For instance, Yu et al.~\cite{MAANS} divided the entire metric map into multiple submaps and proposed an RL-based multi-agent planning module, multi-agent spatial planner~(MSP) to first choose a submap and then select a coordinate within it. This approach significantly reduced the complexity of the exploration space. Nonetheless, Yu et al.~\cite{MAANS} utilized metric maps as spatial representation, which may vary significantly between scenes, limiting their generalization ability across different environments. Moreover, metric maps are resource-intensive; they generate high communication traffic and require substantial memory storage, which poses scalability issues for larger or more complex environments. Yang et al.~\cite{MANTM} adopted topological maps instead of metric maps, which captured abstract yet crucial connectivity information between locations, requiring less bandwidth for communication and being more robust to changes in scene structure. The topological maps in \cite{MANTM} were augmented with ``ghost nodes'', representing potential exploration points. These ghost nodes served as long-term goal candidates for an RL-based hierarchical topological planner~(HTP), further refining the search space and facilitating exploration. Besides, hierarchical reinforcement learning (HRL) has also been employed to tackle the problem of large exploration space. HRL frameworks~\cite{yang2023learning,cai2013combined,liang2021hierarchical} in the multi-robot exploration consist of a high-level policy for task decomposition and a low-level policy for task allocation. Liang et al.~\cite{liang2021hierarchical} developed a hierarchical reinforcement learning approach, called COM-cooperative HRL for multi-robot cooperation in a partially observable environment. Specifically, COM-cooperative HRL addressed the above gaps by introducing a partner selector to learn high-level communication strategy and a low-level controller to select goals based on shared information and individual observation. 

The credit assignment problem in the goal-level learning-based methods is analogous to the task allocation stage in the multi-stage methods, where agents must distribute cooperative workloads effectively through goal assignments. To enhance multi-agent cooperation, some works introduced graph-based models~\cite{zhang2022h2gnn,tolstaya2021multi,li2020graph,he2023multi} or attention mechanisms~\cite{chen2024transformer,chen2023transformer,wang2024multi} to capture the relationships between agents, enabling more coordinated and efficient exploration. 
Zhang et al.~\cite{zhang2022h2gnn} presented the hierarchical-hops graph neural networks~(H2GNN) to enable robots to targetedly integrate the key information of the graph-represented environment, which distinguished the importance of information from different hops around robots based on the multi-head attention mechanism. Wang et al.~\cite{wang2024multi} devised a transformer-based representation network to infer and align latent spatial and temporal dependencies among robots, facilitating cooperative behaviors. Moreover, the goal-level methods show superior performance in scalability, as the goal generation is less sensitive to dynamic changes in the environment. Low sensitivity to real-time environmental dynamics also reduces the need for frequent updates to the exploration strategy, thereby improving the robustness and efficiency of multi-robot teams.

\textbf{2) The Action-Level Methods}

In this method, robots receive observations from the environment and directly yield environmental actions via RL. The action space of the action-level method is related to the available action of each robot, which is often continuous due to the motor systems equipped on the robots. The continuous action space provides exacter control of the robot's movements but introduces complexity in training and policy optimization. Deng et al.~\cite{deng2022multi} utilized deep deterministic policy gradient (DDPG) algorithms to address the training problem of continuous action spaces by estimating both the policy and value functions simultaneously. Chen et al.~\cite{chen2019end} combined convolutional neural networks to process multi-channel visual inputs, curriculum-based learning, and proximal policy optimization (PPO) algorithm for motivation-based reinforcement learning in end-to-end multi-robot exploration tasks. Mete et al.~\cite{mete2023coordinated} developed a COMA-inspired actor-critic architecture to facilitate effective coordination and novelty-based intrinsic team rewards to promote exploration.

The action-level approach~\cite{lin2019end,jun2019goal,shi2019end} must balance both local planning tasks, such as obstacle avoidance, with global planning objectives, like coordinating multi-robot cooperation. Local planning ensures that robots can navigate safely within their surroundings, while global planning facilitates efficient exploration and task completion. Integrating these two levels of planning requires sophisticated policies that can dynamically adjust priorities based on real-time conditions. Cao et al.~\cite{cao2021multi} constructed the policy with Bayesian Linear Regression based on a neural network (called BNL) to compute the state-action value uncertainty efficiently for safe planning. It combined distributional reinforcement learning to estimate the intrinsic uncertainty of the state-action value globally and more accurately. Huang et al.~\cite{huang2024collision} implemented a curriculum and a replay buffer of the clipped collision episodes to improve performance in obstacle-rich environments. It further applied an attention mechanism to attend to the neighbor robots and obstacle interactions.

Applying the action-level methods to heterogeneous robots is another challenge since heterogeneous robots may have diverse observation and action spaces, making it difficult to apply a shared policy across all robots. Bettini et al.~\cite{bettini2023heterogeneous} introduced heterogeneous graph neural network proximal policy optimization (HetGPPO), a paradigm for training heterogeneous MARL policies that leveraged a graph neural network to learn heterogeneous behaviors. Seraj et al.~\cite{seraj2024heterogeneous} proposed heterogeneous policy networks~(HetNet) to learn efficient and diverse communication models for coordinating cooperative heterogeneous teams, building on heterogeneous graph-attention networks.

\section{Multi-robot Communication Strategies in Unpredictable Environments}
\label{sec: comm}

In multi-robot cooperative exploration systems, whether using distributed or centralized communication, robots typically need to share map information and planned paths to ensure efficient, safe, and rapid exploration. Early research on multi-robot cooperative exploration, such as~\cite{yamauchi1998frontier, burgard2000collaborative, ko2003practical}, assumed absolutely reliable communication among all robots, ignoring specific communication limitations. In practice, however, achieving fully reliable communication is nearly impossible. Studies~\cite{comm_survey} have identified bandwidth and range limitations as primary constraints, which we have summarized in Section \ref{sec: intro}.

To address these limitations, various methods have been proposed, focusing on improving map information structures and communication-aware strategies to mitigate the effects of bandwidth and range constraints. Map information structures involve the shared representation of environmental data among robots, with optimizations aimed at reducing data volume and enhancing transmission efficiency. Communication-aware strategies determine how robots organize and exchange information, with improvements playing a critical role in increasing communication reliability.

This section focuses on works addressing these communication constraints, introducing methods to optimize map information structures and communication-aware strategies under bandwidth and range limitations.\newline

\noindent\textbf{V-A. Map Information Structure}

In multi-robot cooperative exploration tasks, a map has two primary functions.
First, it \textit{represents known free space}, allowing the system to distinguish between explored and unexplored regions, which drives robots to continuously gather information about unknown areas.
Second, it \textit{represents occupied space}, capturing environment structures and obstacle distributions, which support task allocation among robots and enable collision avoidance during exploration.

Besides the above fundamental functions, an ideal map representation for exploration tasks should also achieve \textit{high accuracy} to fully capture the information necessary for cooperative planning. 
However, higher accuracy often comes at the cost of increased computational demands, posing challenges for resource-constrained onboard platforms in practice.
Therefore, it is essential that the mapping method also maintains \textit{low computational complexity} to ensure real-time performance.
Moreover, richer global environment information enables better cooperative planning, necessitating frequent map sharing and fusion. 
However, in real-world scenarios, limited communication bandwidth can cause data delays or losses during transmission. 
Thus, the ideal map representation should also ensure a \textit{small data transmission volume} for efficient and reliable information sharing.

\begin{figure*}
    \centering
    \includegraphics[width=\linewidth]{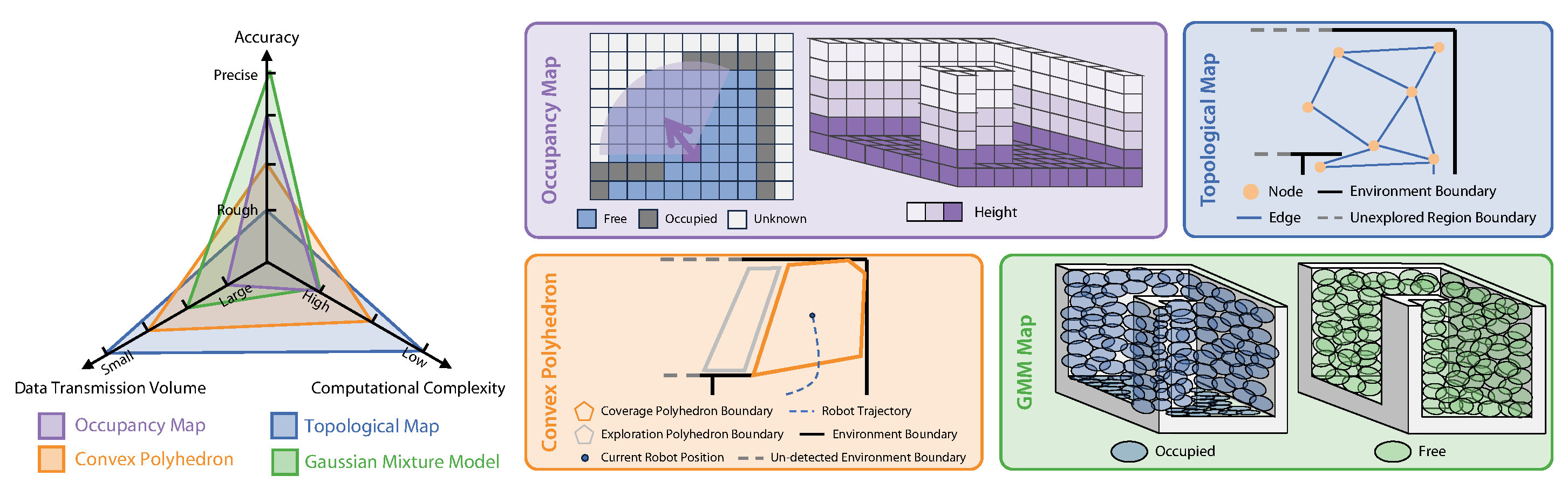}
    \caption{Comparison of different types of map representation. Accuracy represents the precision of the environmental representation, data transmission volume refers to the amount of data required for communication during map fusion, and computational complexity indicates the resources consumed by the computation.
    }
    \label{fig:comm_map}
\end{figure*}

To achieve a map that closely approximates the ideal, current researchers have developed various representation methods, including occupancy maps, convex polyhedra, Gaussian mixture models (GMM), and topological maps. Fig.~\ref{fig:comm_map} shows a comparison of different types of map representations. Although occupancy maps can express the environment with high precision, they require calculating large data because of the high computational complexity, leading to both high memory storage space requirement and large data transmission volume when shared. Convex polyhedron has low computational complexity to assess obstacles and small data transmission volume to share, but its accuracy is rough. GMM has extremely high precision in map representation and appropriate data transmission volume, but the computation is complex, requiring more resources to calculate. Topological maps are very rough in expression but can be computed with fewer resources because of their low computational complexity and minimal data transmission volume. Additionally, RL-based multi-robot exploration systems have driven the creation of novel map representations tailored to specific training requirements.\newline

\noindent\textbf{V-A-1. Occupancy Map}

Early works~\cite{yamauchi1998frontier, burgard2000collaborative, burgard2005coordinated}, used the occupancy maps to represent the environments, which employ grids or voxels, with cells marked as free or occupied. The multi-robot cooperative exploration system can easily identify the occupied cells as obstacles, the free cells as explored areas, and the unknown cells as unexplored areas, which helps in cooperative planning.
Zhou et al.~\cite{zhou2023racer} proposed an online exploration space hgrid decomposition method to represent exploration tasks, broadcasting hgrid blocks to share exploration information. The advantage of this approach lies in its ability to achieve high map detail by reducing the size of the cubes.

However, because of the large data of occupancy maps, the system requires transmitting large volumes of cell data, which poses challenges under bandwidth constraints. To address these issues, methods such as OctoMap~\cite{hornung2013octomap} and UFOMap~\cite{duberg2020ufomap} were developed, using Octree structure dividing the 3D space into octants at various levels of resolution.
Zhang et al.~\cite{zhang2024leces} introduced a system where each unmanned aerial vehicle (UAV) constructs a locally improved OctoMap. When the number of newly discovered voxels exceeds a predefined threshold, a sub-map sharing mechanism is triggered. Upon receiving a new map, a UAV integrates it with its existing map using the ID and vertex information of the blocks.

In a multi-robot cooperative exploration system, the occupancy map can adjust the size of the grid or voxel cells to accommodate different scales of environments and varying accuracy requirements. However, a high-precision occupancy map requires substantial computing resources for updating and fusion, and a large bandwidth for transmission, meaning that a high-precision occupancy map has high computational complexity and large data transmission volume. Therefore, during multi-robot communication, there is a high chance of encountering map data transmission issues, such as incomplete or failed data transfer. This makes it challenging to match the features of different sub-maps, ultimately resulting in unsuccessful map fusion.\newline

\noindent\textbf{V-A-2. Convex Polyhedron}

Convex polyhedra provide a compact representation of free space in the environment using the process of convex decomposition. 
In multi-robot cooperative exploration tasks, using convex polyhedra to represent free and unobstructed space effectively reduces data transmission volume and computational complexity, and has enough accuracy in a static environment.

Yang et al. \cite{yang2021graph} utilized convex polyhedra to represent 3D free space in a single-robot exploration system. By employing Euclidean distance clustering algorithms, point cloud data was constructed into multiple convex polyhedra, each representing an independent exploration region within the boundary space. Visibly connected regions were merged into larger convex polyhedra. 
Gao et al. \cite{gao2022meeting} used lightweight star-convex polytopes to represent the known free space, and used meshes to represent frontier.
When robots meet, sub-robots transmit their star-convex polytopes information and frontier meshes with viewpoints to the host robot, which merges all received data and removes frontiers located within the free space. The final merged map is then shared back with all meeting robots.

However, the modeling process of convex polyhedrons involves constructing multiple convex polyhedra and combining them using Boolean operations when faced with complex environments. This approach can readily result in a loss of map accuracy, leading to cooperative planning that falls short of the method’s optimal performance. Furthermore, in dynamic environments, each change necessitates the reconstruction of the convex polyhedra, which increases computational complexity and data transmission volume. These factors pose challenges to real-time communication and map integration for multi-robot system.\newline

\noindent\textbf{V-A-3. Gaussian mixture model}

The Gaussian mixture model (GMM) is a statistical model used to represent a probability distribution composed of multiple Gaussian distributions. The expectation-maximization (EM) algorithm is typically employed during the modeling process to estimate the density function, effectively compressing large data into a few parameters, which reduces the data transmission volume needed for robot-to-robot sub-map exchange in a multi-robot cooperative exploration system. However, GMM outputs probability densities rather than geometric features, it necessitates post-processing for use in cooperative planning.

Corah et al. \cite{corah2019communication} used GMM to model the density function of independently and identically distributed point cloud data sampled from 3D space. The GMM components were parameterized by mixture coefficients, means, and covariances, and learned through the EM algorithm. Each robot maintained its own GMM independently and shared GMM maps by transmitting the parameters. Upon receiving shared data, robots updated their local GMMs accordingly. Similarly, Meadhra et al. \cite{o2018variable} utilized GMM to compactly encode information about occupied surfaces and free spaces. They employed a Bernoulli distribution to model the evidence for occupancy and free spaces, calculating the occupancy probability for each cell. To ensure maximum resolution, they compared the area under the curve (AUC) scores between the reconstructed occupancy map and the true occupancy map.

GMM can apply consistent probability density modeling across various sensor data, including LiDAR and cameras, which enables cooperative mapping among diverse types of robots in the multi-robot system. However, GMM’s reliance on incremental parameter updates can lead to delayed responses to rapidly changing dynamic obstacles, such as pedestrians, making it less effective in dynamic environments.\newline

\noindent\textbf{V-A-4. Topological Map}

The topological maps simplify the environments into nodes and edges. Nodes denote significant locations in space and edges describe the connectivity between these locations. Unlike maps that explicitly depict obstacles, topological maps only emphasize the spatial relationships between nodes. This simplicity speeds up algorithms like Dijkstra search~\cite{dijkstra2022note}, enhancing cooperative planning efficiency in multi-robot exploration tasks and making it well-suited for large-scale environment exploration. 

Bayer et al. \cite{bayer2021decentralized} utilized topological maps to guide a multi-robot cooperative exploration system. As a robot travels a certain distance, it adds its current position as a new node to the topological map. Newly created nodes are broadcasted to other robots to minimize the communication load during map expansion. Zhang et al. \cite{zhang2022mr} employed panoramic cameras to gather environmental data and used image retrieval algorithms for node identification. New nodes are added only when the robot encounters unexplored areas, avoiding redundancy. Moreover, only the newly created nodes and edges are shared between robots, significantly reducing communication data. The integration of old and new maps is achieved through node descriptors. Dong et al.~\cite{dong2024fast} proposed a dynamic multi-robot topological map framework, where each robot cooperatively maintains the map. They introduced the concepts of historical nodes and explorable regions of interest (EROI). Edges, traditionally representing visibility connections, are redefined as weighted paths indicating traversable routes between historical nodes and EROI, with weights based on path length. A Dijkstra tree is employed for efficient path queries, further decreasing communication overhead.

For topological maps, when multi-robot collaborate in mapping, only the data changes in nodes and edges need to be communicated, which reduces the data transmission volume. In dynamic environments, if an edge on the topological map becomes impassable due to obstacle changes, the edge can be easily removed, eliminating the need to reconstruct the entire map. 
However, topological maps lack the capability to represent detailed structures, making them likely to be ineffective in complex, unstructured environments like forests or ruins. When multiple robots explore the same area simultaneously, the topological map may end up with multiple similar nodes being generated repeatedly. Furthermore, task allocation that relies on nodes can result in an uneven distribution of workload among the robots.\newline

\noindent\textbf{V-A-5. Map representation for RL}

When applying RL techniques within the coordinated motion module, maps play an essential role as inputs to RL policies. To facilitate effective policy learning, these maps must offer rich state representations. However, traditional map types—such as occupancy grid maps and topological maps—often fall short of representing the historical trajectory information of agents. To overcome this limitation, RL policies often design new map representations that more efficiently encode environmental states and robot motion data, while simultaneously optimizing data transmission and storage.

Wang et al. \cite{wang2021spatial} represented the robot’s state using a combination of a top-view map, a robot position map, a visit frequency map (VFM), and a shortest path map to guide the agent in exploring unknown areas. The robot’s actions were encoded using a spatial action map (SAM), where each pixel in the VFM represented the number of times that area was scanned by a depth sensor, indicating its degree of visitation. SAM encoded potential action commands as pixels, with each pixel corresponding to a possible action point. However, Chen et al. \cite{chen2023efficient} stated that Wang et al.’s approach required transmitting a large amount of data due to its four-channel state representation and was only suitable for single-agent systems. To address this limitation, they proposed the i-VFM method, which applies a sigmoid function to the VFM channel values and sets all pixel values of the top-view map and robot position map to be higher than the maximum sigmoid output. This transformation reduces the four-channel input to a two-channel input, significantly decreasing the data volume transmitted during communication and mitigating bandwidth-related constraints.\newline

\noindent\textbf{V-B. Communication-aware Strategy}

\begin{figure*}    
    \centering
    \includegraphics[width=\linewidth]{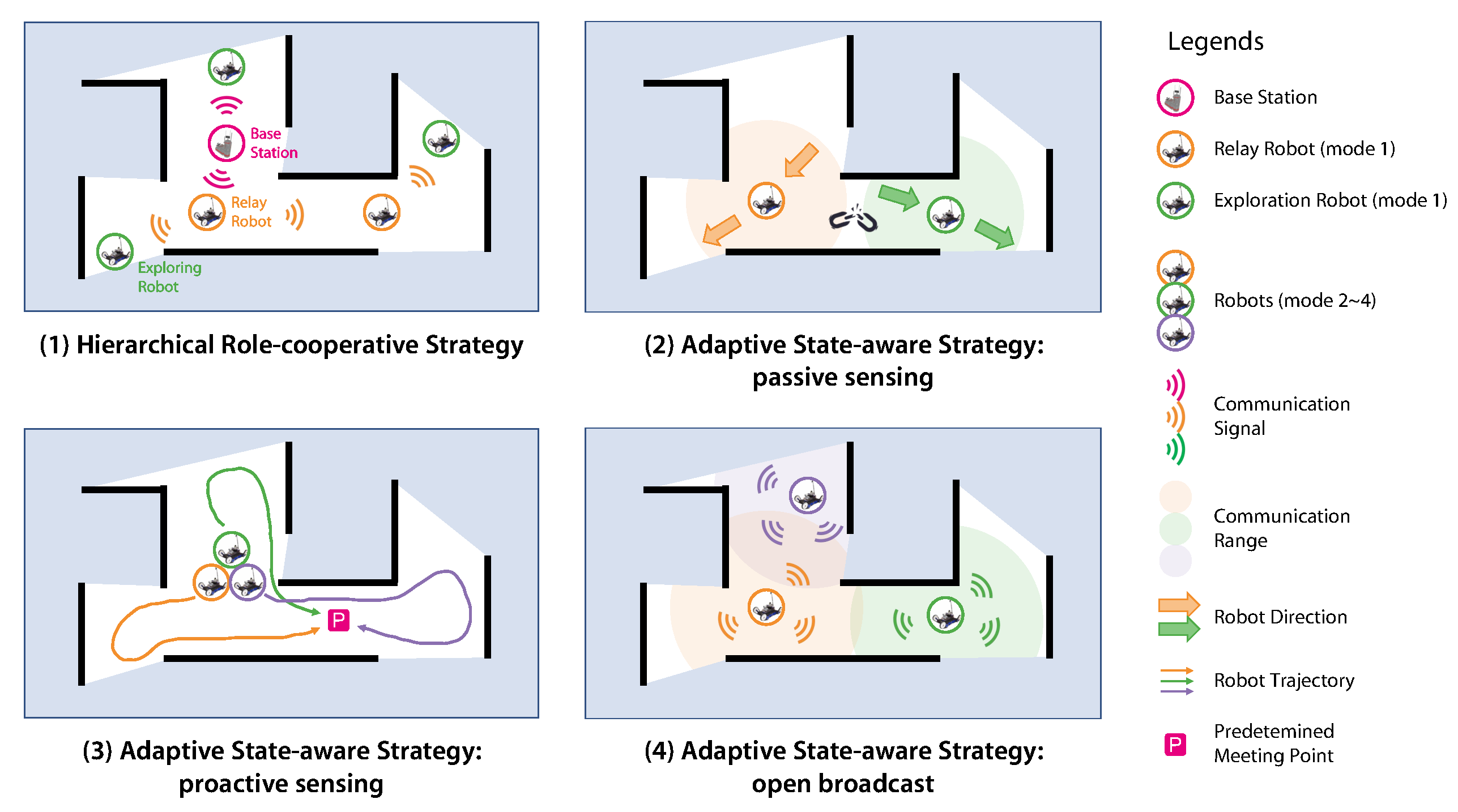}
    \caption{Examples of two different communication-aware strategy and their subdivided communication mechanisms. In the hierarchical role-cooperative strategy, a base station coordinates two types of robots: one for exploration and the other for relaying information back to the base station. The proactive sensing mechanism of adaptive state-aware strategy involves robots moving and communicating according to predefined rules, allowing them to actively disconnect communication and reconnect at specific times and locations. In the passive sensing mechanisms of adaptive state-aware strategy, robots can continue executing tasks even when they leave the communication range. The open broadcast mechanism of adaptive state-aware strategy relies on broadcast information to guide robots as they collaborate and perform map fusion.}
    \label{fig:communication}
\end{figure*}

In multi-robot cooperative exploration tasks, practical communication is limited by the range constraints of equipment. These limitations require robots to follow certain communication-aware strategies to enable localization and sub-map information transmission, thereby achieving cooperative planning and map fusion. Based on the differences between centralized and distributed approaches in task allocation and map transmission methods, we categorize the primary communication-aware strategies into two types: hierarchical role-cooperative strategy and adaptive state-aware strategy. Fig.~\ref{fig:communication} depicts the examples of two strategies.

\noindent\textbf{V-B-1. Hierarchical Role-cooperative Strategy}

In multi-robot cooperative exploration systems, the hierarchical role-cooperative strategy comes from communication methods that rely on a base station or central computing platform fixed in one place. Due to the limited communication range of the fixed platform and the requirement to relay pending information back to it, a communication-aware strategy has emerged, in which some robots or other equipment actively serve as relays to extend the effective communication range, and the other robots serve as explorers to keep exploring or tasks. The hierarchical role-cooperative strategy creates a layered communication architecture with base stations at the top, relay robots as intermediaries, and exploration robots at the operational tier, which has been applied in multi-robot cooperative exploration for search and rescue missions, enabling a central command to make new decisions based on freshly gathered information.

De Hoog et al. \cite{de2009role} introduced a hierarchical role-cooperative exploration framework in which robots were categorized as explorers and relayers, coordinating their actions via designated meeting points. Explorers advanced into unknown areas to collect information, while relayers shuttled between explorers and the base station to facilitate data transmission. Building on this concept, Cesare et al.~\cite{cesare2015multi} refined the framework by defining four distinct robot states: explore, meet, sacrifice, and relay. Robots in the explore state employed frontier-based exploration algorithms to autonomously navigate the environment. When their exploration duration exceeded a predefined threshold or their battery levels became insufficient for a safe return to the base station, they transitioned into the meet state, seeking other robots to negotiate role transitions. Robots assuming the sacrifice role continued exploring until their batteries were depleted, whereas those in the relay role landed and remained stationary to serve as relay nodes, ensuring continuous data flow. Lillian et al.~\cite{clark2022propem} proposed the PropEM-L method, which integrates 3D geometric perception with machine learning to coordinate multi-robot collaboration centered around a base station, dynamically deploying relay nodes to extend signal coverage. This method employs online neural networks to predict signal attenuation, thereby significantly enhancing communication reliability and adaptability between base stations and relays in complex scenarios. Xia et al.~\cite{xia2023relink} propose a multi-robot deployment framework called REal-time LINE-of-sight-based deployment frameworK (RELINK). This framework constructs a guidance tree to connect base stations with relay nodes, employs tree pruning algorithms to eliminate redundant relay paths, integrates star-shaped convex polygon optimization to transform visibility constraints into differentiable penalty terms, and utilizes the L-BFGS solver~\cite{flab2021} for low-energy communication link planning among base stations, relays, and mobile clients while ensuring real-time performance and connectivity.

The hierarchical role-cooperative strategy effectively mitigates the challenges posed by the limited communication range in centralized multi-robot cooperative exploration systems. While the frequent meetings it necessitates can result in numerous non-informative movements, leading to resource wastage and reduced exploration efficiency, this strategy remains a viable solution for addressing communication constraints in base-station-dependent exploration strategies.\newline

\noindent\textbf{V-B-2. Adaptive State-aware Strategy}

The adaptive state-aware strategy is a distributed communication-aware strategy that employs predefined rules for communication awareness in multi-robot cooperative exploration systems. These rules specify that robots should proactively monitor their own communication status to determine appropriate times and locations for initiating communication exchanges, as well as define actionable protocols for recovering communication in case of unexpected failures, thereby ensuring the seamless progression of system-wide cooperative planning and map fusion processes. The adaptive state-aware strategy can be further categorized into two communication mechanisms. One is \textit{proactive sensing}, the other is \textit{passive sensing}. 

In proactive sensing, robots adaptively initiate or suppress communication-based on real-time changes in exploration, adapting their actions to optimize network efficiency.
Sheng et al.~\cite{sheng2004multi} introduced a proximity measurement index based on inter-robot distances to evaluate their ability to maintain communication links. By adjusting the weight of this proximity measure in the decision-making process, the system can control the degree of robot clustering. This approach prevents issues like network congestion caused by excessive clustering or communication failures due to robots being too far apart. Similarly, Masaba et al.~\cite{masaba2021gvgexp} designed a method that minimizes communication frequency. Under their strategy, communication only occurs during initial deployments, after exploring gate areas, when connecting these areas to known parts of the environment, and when robots reach leaf nodes that lead into unknown regions. This selective communication reduces bandwidth load. Bramblett et al.~\cite{bramblett2022coordinated} proposed a communication framework that integrates intentional disconnection with dynamic assembly. Through intentional disconnection, robots can explore distinct areas without maintaining constant communication, thus speeding up exploration. Dynamic assembly ensures that robots later share the gathered information, coordinate tasks, and reallocate duties as needed. Gao et al.~\cite{gao2022meeting} further introduced a task-based communication protocol with alternating centralized planning and decentralized exploration phases. During the decentralized phase, robots deliberately disconnect to conserve communication energy and broaden the exploration radius. They then reconvene at predetermined locations and times to engage in centralized planning, sharing information, and recalibrating their strategies.

In passive sensing, upon disconnection, robots adjust their behavior adaptively using predicted state information to restore connectivity, ensuring continuous system coordination and data synchronization.
Building on segmentation and distributed auction algorithms, Woosley et al.~\cite{woosley2021bid} proposed a system in which, when communication is interrupted, robots employ a bidding prediction model to anticipate the bids of other team members. This predictive mechanism enables robots to complete distributed auctions and maintain their exploration activities despite the loss of communication.
Similarly, Schack et al.~\cite{schack2024sound} introduced a framework that allows robots to schedule meetings or conduct parallel exploration by predicting the likely positions of their teammates. When a robot becomes disconnected, it interprets the absence of communication as an observation, using a Bayesian filter to update the probability distribution of its teammates’ states. By simulating these state changes and inferring information about unexplored areas, the robot selects the optimal target for further exploration or communication through a bidding mechanism.

It is worth noting that in distributed multi-robot cooperative exploration systems, during sub-map transmission, some systems may employ open broadcast mechanisms, where robots continuously emit newly generated sub-map information to all peers. As exploration progresses, at least one robot will receive this incremental sub-map data from another robot and integrate it with its own maintained map through map fusion. Importantly, such information remains broadcast persistently until all other robots in the system have ultimately acquired it. In the system devised by Dong et al.~\cite{dong2024fast}, the robots will broadcast the updated segments of the multi-robot dynamic topological graph (MR-DTG), which comprise new history nodes, explorable regions of interest and connecting edges, to all other robots in the network. Upon receipt of this information, the respective robots update their local data and subsequently engage in graph Voronoi partition and motion planning, guided by the newly updated MR-DTG.

In distributed multi-robot cooperative exploration systems, the adaptive state-aware strategy enhances communication efficiency by proactively monitoring and adjusting robot interactions based on predefined rules. However, its reliance on manually crafted regulations introduces vulnerabilities in dynamic or extreme environments. Such systems may struggle to adapt to unpredictable scenarios, leading to delayed information exchange between robots. This delay not only wastes resources through unnecessary rendezvous attempts but also causes prolonged disconnections, gradually amplifying predictive errors in shared environmental models. Over time, these cumulative errors degrade the consistency of collaborative planning, forcing robots to operate with outdated or conflicting knowledge. Consequently, the overall exploration efficiency diminishes as suboptimal decisions are made due to incomplete or stale information, highlighting a critical trade-off between rule-based adaptability and robustness in highly variable contexts.
\section{The Field Application of Multi-robot Cooperative Exploration Systems}
\label{sec: fie}

Unlike laboratory research that seeks innovation and superior performance, field multi-robot cooperative exploration systems emphasize robustness, stability, and efficiency. Indoor environments have sparse obstacles, so single-type robot exploration systems can explore them effectively. However, in extreme and complex field settings like underground mines, tunnels, or underwater areas, challenges such as limited visibility, communication constraints, rough terrain, and GNSS-denied make single-type robot systems inefficient. This drives the need for heterogeneous robot systems. heterogeneous robots have different motion models, dynamics, payload capacities, computational resources, and endurance. Therefore, tasks must be allocated according to each robot’s strengths to enhance exploration efficiency. 

This section selects the work of excellent teams in the US Defense Advanced Research Projects Agency (DARPA) Subterranean (SubT) Challenge~\cite{DARPA2022,orekhov2022darpa}, including MARBLE~\cite{ohradzansky2022multi}, Explorer~\cite{scherer2022resilient}, CTU-CRAS-NORLAB~\cite{roucek2021system}, CSIRO Data61~\cite{hudson2022heterogeneous}, CoSTAR~\cite{agha2022nebula}, and CERBERUS~\cite{tranzatto2022cerberus}, and discusses the design of heterogeneous multi-robot cooperative exploration systems and their applications for exploring extreme, complex, and large-scale field environments.

\vspace{2mm}
\noindent\textbf{VI-A. Heterogeneous Robots in Multi-robot Cooperative Exploration Systems}
\vspace{2mm}

A heterogeneous multi-robot system is a team of robots that differ in type, with differences in shape, mobility, and sensor configuration~\cite{rizk2019cooperative}. This diversity enables the system to perform tasks efficiently in complex, unknown, and dangerous field environments through complementary advantages. In the DARPA SubT Challenge, many teams demonstrated how to use heterogeneous multi-robot cooperative exploration systems to deal with various challenges in underground environments. Based on the mobility of robots, these systems mainly consist of three types: UGV (unmanned ground vehicles), legged, and UAVs. UGV can be further divided into wheeled and tracked categories.

Wheeled robots usually have high speed and long endurance, making them ideal for moving quickly on flat or slightly rough terrain. Their higher payload capacity allows them to carry more sensors and equipments. They can rapidly cover large areas to obtain environmental information and deploy communication nodes to enhance the network. This makes them well-suited for perception, mapping, and communication node deployment tasks. Examples include the Husky robot used by CoSTAR and CTU-CRAS-NORLAB, Armadillo from CERBERUS, and R1/R2 from Explorer.

When moving over very rough terrain, wheeled robots often struggle due to their limited obstacle - crossing ability, which has led to the need for tracked robots. Tracked robots can handle rough, muddy, or soft ground well. They are stable, can carry many sensors and devices, and are suitable for long - term tasks in complex terrain, such as perception, mapping, and communication node deployment. Examples are the Absolem from CTU - CRAS - NORLAB, the BIA5 ATR from CSIRO Data61, and the SuperDroid LT2 - F. However, since tracked robots are generally slower than wheeled ones, it's important to analyze and decide which robot to use for exploring different terrains to ensure efficient exploration.

Legged robots like quadrupedal and hexapodal robots can stably walk on stairs, slopes, and uneven or muddy terrain due to their strong adaptability. Their flexibility and mobility make them suitable for complex and narrow areas. With a certain payload capacity, they can carry multiple sensors and computing units, enabling highly autonomous navigation and exploration. Examples include CTU-CRAS-NORLAB's PhantomX Mark II,CSIRO Data61's Hexapod, Ghost Robotics' Vision60, and CERBERUS's ANYmal B300.

UAVs can move quickly in the air, unrestricted by ground obstacles. Equipped with light sensors like cameras and LiDAR, they offer an aerial perspective. They help the team understand the overall environment, assist other robots in path planning, and explore narrow or dangerous areas such as shafts and crevices. UAVs can also act as communication relay nodes to expand network coverage. To address their limited endurance, they can be transported on wheeled or legged robots and deployed when needed, as demonstrated by Explorer's DS.

The multi-robot cooperative exploration system with heterogeneous robots can combine the strengths of different robot types during exploration, reduce the weaknesses of single-type robots, and enhance system stability, robustness, and exploration efficiency. However, it also causes problems like ensuring consistent localization estimation, aligning multi-modal perception and maps, planning multi-agent collaboration, and designing a reliable communication network. Existing possible solutions to these problems have been described in Section~\ref{sec: LM}, ~\ref{sec: coop}, ~\ref{sec: comm}, so won't repeat here.

\vspace{2mm}
\noindent\textbf{VI-B. The Field Application of Multi-robot Cooperative Exploration Systems}
\vspace{2mm}

In the DARPA Subterranean Challenge, heterogeneous multi-robot cooperative exploration systems have unique advantages in dealing with complex and variable outdoor environments such as tunnels, urban, and caves because of the limits of these environments. In such environments, balancing robustness, stability, and efficiency of multi-robot cooperative exploration systems is crucial because the environments are full of uncertainties and challenges. Over-relying on a useful but complex module can limit system performance due to the buckets effect. Therefore, each module should be simple and practical while ensuring close cooperation. In addition, tasks should be allocated according to the characteristics of heterogeneous robots to give full use to their strengths.

The CoSTAR framework's Networked Belief-aware Perceptual Autonomy (NeBula) system comprises the following modules: a localization and mapping module consisting of Heterogeneous and Resilient Odometry estimator (HeRO)~\cite{santamaria2019towards} and Large-scale Autonomous Mapping and Positioning (LAMP)~\cite{badi2020lamp}, a cooperative planning module Probabilistic Local and Global Reasoning on Information roadMaps(PLGRIM)~\cite{kim2021plgrim}, and a communication module Collaborative High-bandwidth Operations with Radio Droppables (CHORD)~\cite{ginting2021chord}.
HeRO enhances localization resilience through parallel processing of multiple sensors and algorithms, such as integratig LiDAR-inertial (LIO), visual-inertial (VIO), thermal-inertial (TIO) kinematic-inertial (KIO), contact-inertial (CIO) and RaDAR-inertial (RIO), while considering redundancy and heterogeneity. This architecture enables fault detection and system adaptation to ensure accurate and stable positioning in GNSS-denied field environments.
LAMP serves as a factor graph-based SLAM solution designed for large-scale, perceptually degraded environments, such as those encountered in the DARPA Subterranean Challenge. It achieves consistent global environmental representation by fusing multi-sensor inputs to detect loop closures and mitigate accumulated errors.
PLGRIM employs a hierarchical Information RoadMap (IRM) representation and planning architecture to enable efficient exploration in expansive environments. The system models autonomous exploration as a partially observable Markov decision process (POMDP), enabling uncertainty-aware decision-making. By integrating local and global planning strategies, PLGRIM dynamically adjusts robotic exploration paths to maximize information gain while minimizing action costs.
CHORD maintains high-bandwidth communication links among multiple robots, ensuring effective command execution, autonomous operations, and data collection in communication-constrained field environments. The module extends wireless mesh networks through deployable communication nodes, enabling extended-range connectivity between robots and base stations.
Besides, NeBula implements mission-specific role allocation based on robotic platform capabilities and environmental characteristics: tracked and hybrid-powered robots navigate complex mine and tunnel terrains, while UAVs perform rapid data relay or vertical shaft exploration. This heterogeneous coordination leverages each platform's unique advantages to optimize system performance.

In the localization and mapping module, the CERBERUS system employs a complementary hierarchical multi-modal sensor fusion approach (CompSLAM)~\cite{khattak2020complementary}, integrating data from visual/thermal cameras, LiDAR, and inertial sensors to achieve robust robot pose estimation. This fusion strategy ensures reliable localization even under degraded perceptual conditions like geometric self-similarity, darkness, or obscurants. Additionally, CERBERUS adopts a centralized multi-robot multi-modal mapping (M3RM)~\cite{schneider2018maplab} optimization method, where submaps from individual robots are aggregated and refined at the base station using factor graphs with visual, LiDAR, and prior constraints. This approach not only enhances the localization accuracy of individual robots but also improves the system’s environmental perception consistency, enabling artifact position refinement in the DARPA coordinate frame.
For the cooperative planning module, CERBERUS uses Graph-based exploration path planner (GBPlanner)~\cite{dang2020graph} - a unified graph-search-based exploration path planning method applicable to both legged and aerial robots. This method guides robots to autonomously explore complex environments efficiently through hierarchical local-global planning with volumetric information gain metrics. During planning, the system generates optimal exploration paths based on occupancy maps, traversability constraints from elevation maps, and frontier vertices from a global graph. The planning framework also incorporates geofencing for obstacle avoidance and accounts for robot motion constraints to ensure safe navigation. Furthermore, CERBERUS implements automated homing based on remaining mission time and supports multi-robot coordination through frontier sharing in the global graph.
The communication module of CERBERUS employs multiple strategies to maintain reliable connectivity between robots and the base station. For underground environments, wheeled robots equipped with high-gain antennas and fiber-optic cables establish stable communication links with the base station. Meanwhile, legged robots dynamically deploy wireless relay nodes at critical locations to extend the communication network’s coverage. This creates an adaptive, dynamic communication network during exploration missions.
CERBERUS also optimizes task allocation based on robot capabilities and environmental demands. The ANYmal quadrupedal robot serves as the primary exploration platform due to its outstanding endurance and complex terrain traversal capabilities, enabling stable navigation in rugged underground environments. The Aerial Scouts multi-rotor UAVs leverage their rapid flight agility to explore confined spaces like narrow passages or vertical shafts and perform fast reconnaissance. The ArmadiIIo wheeled robot specializes in communication node deployment and object detection, providing network support via its high-gain antennas and fiber-optic cables while contributing to environmental sensing tasks. 

In the localization and mapping module, CSIRO Data61 developed the Wildcat SLAM system~\cite{ramezani2022wildcat}, which enables each robot to generate a shared global map and achieve efficient environmental mapping and localization through multi-sensor data fusion. The Wildcat SLAM system generates SLAM frames and shares them across robots, allowing each robot to construct a globally consistent map based on both its own data and inputs from other agents. While absolute consistency across all robots’ global maps cannot be guaranteed in complex field environments, the overlapping of frames from a shared initial region enables each robot to incrementally build its global solution using all available data.
For the cooperative planning module, CSIRO Data61 designed a Multi-Robot Task Allocation (MRTA) method. This approach employs a market auction mechanism, where robots bid for tasks based on expected rewards, enabling rational task distribution. Robots dynamically adjust task priorities and bidding strategies by evaluating task locations, their own states, and communication conditions with other robots. Additionally, CSIRO Data61 introduced a priority zone tool, allowing human operators to influence task allocation, thereby enhancing system flexibility and adaptability.
In the communication module, CSIRO Data61 designed the Mule communication system, which utilizes a peer-to-peer network to enable robots to share data within communication range while maintaining task execution during outages. The Mule system dynamically discovers other robots in the network and establishes point-to-point connections for efficient data sharing. This strategy addresses communication challenges in subterranean environments and ensures continued collaboration post-recovery through persistent data storage and synchronization mechanisms.
CSIRO Data61 assigns roles based on robotic capabilities. The BIA5 ATR robot, with its strong obstacle-crossing capability and high payload capacity, is tasked with carrying communication nodes and UAVs, while the smaller SuperDroid LT2-F robot leverages its agility and compact size to explore narrow spaces. UAVs, benefiting from their high speed and maneuverability, rapidly survey large areas or hard-to-reach zones.

The system designed by CTU-CRAS-NORLAB employs a robust Iterative Closest Point (ICP)~\cite{chen1992object} based localization and mapping method, applied across all robot types to ensure stability and reliability in complex environments. In Urban scenarios, characterized by vertically diverse structures, the team adopted the Advanced implementation of the original Lidar Odometry and Mapping~\cite{zhang2014loam} (A-LOAM) algorithm to process 3D LiDAR data. This method eliminates the need for prior motion estimation, making it more suitable for navigating intricate building interiors. Although challenges such as map warping and drift persist, adjustments to the algorithm and integration of auxiliary measurements reduce localization errors, thereby enhancing the robustness of the localization and mapping module.  
Additionally, CTU-CRAS-NORLAB developed vision-based, radio-based, and olfactory-based target localization methods, alongside adaptive traversal and information entropy-driven exploration strategies. These approaches enable robots to plan paths and explore unknown regions based on environmental data and onboard sensor inputs. By refining frontier-based exploration, the system reduces computational demands while improving environmental coverage efficiency through camera model-aware exploration waypoint generation.  
Multi-robot coordination is primarily achieved via 4G Mobilicom mesh transceivers, allowing operators to directly control robots or set navigation goals. Robots share positional data through low-bandwidth links to avoid redundant exploration. Furthermore, CTU-CRAS-NORLAB implemented a multi-network communication strategy combining short-range Wi-Fi, Mobilicom mid-range links, and Mote long-range links. These provide varying levels of reliability, range, and bandwidth to meet data transmission demands in challenging field environments. In Tunnel settings, larger Mobilicom modules extend communication range, while smaller modules are deployed in Urban environments, enabling all robotic platforms to act as communication nodes and enhance network coverage. Mote modules facilitate critical data sharing—such as status, location, and detection information—among robots, ensuring peer-to-peer communication even when base station connectivity is lost.  
CTU-CRAS-NORLAB emphasizes task allocation based on robotic capabilities: large wheeled robots excel in long-distance exploration, while compact hexapod robots navigate confined spaces. Larger robots also carry additional equipment, such as communication relays, to support the system’s communication infrastructure. This heterogeneous task distribution optimizes exploration efficiency by leveraging the unique strengths of each platform.

The multi-robot cooperative exploration system developed by Explorer ensures robustness and stability in complex field environments through modular design and rational allocation of tasks among heterogeneous robots. In the localization and mapping module, Super Odometry~\cite{zhao2021super} employs an IMU as the primary sensor, integrating constraints from vision, thermal imaging, and LiDAR odometry. A selective fusion strategy automatically prioritizes effective constraints while rejecting erroneous data associations, significantly enhancing real-time performance and enabling rapid recovery from potential sensor failures. By decomposing large factor graphs into multiple sub-factor graphs such as IMU, LIO, and VIO, each sub-factor graph tightly couples raw sensor measurements internally while maintaining loose coupling between sub-factor graphs through pose constraint transmission. This architecture ensures exceptional robustness in harsh environments, allowing the system to continue operating via redundant sensor constraints even if one sensor fails.  
For cooperative planning, a hierarchical exploration planner manages sparse global-level information to generate coarse exploration routes, while maintaining dense local-level information to ensure comprehensive sensor coverage. The global planner operates on low-resolution maps to define exploration directions, whereas the local planner utilizes high-resolution maps for detailed path planning, achieving full coverage of localized areas.  
The communication module incorporates an autonomous node deployment algorithm and a ledger mechanism. The deployment algorithm evaluates communication quality between robots and deployed nodes using local point cloud data. A ray-casting method from the robot to existing nodes determines "line-of-sight visible" or "non-line-of-sight visible" status. New communication nodes are autonomously deployed when distances to existing nodes exceed a threshold, ensuring network coverage. The ledger, a distributed data structure for efficient map and object detection data sharing, records all submap metadata and broadcasts updates. Robots can request specific submaps via the ledger, which also optimizes bandwidth usage by tracking submap availability and local storage status.  
In heterogeneous robot task allocation, the system categorizes robots into UGVs(R1, R2, R3) and UAV(DS) based on perception and mobility capabilities. UGVs excel at long-distance, high-speed exploration and communication network deployment. In tunnel environments, R1 and R2 explore extended passages while establishing communication backbones. The UAV DS specializes in three-dimensional exploration of confined spaces. In urban settings, DS rapidly maps multi-story structures and narrow passages within buildings. 

In the localization and mapping module, MARBLE utilizes tools and methodologies such as Google Cartographer~\cite{hess2016real} and Voxblox~\cite{oleynikova2017voxblox} to support robot positioning and mapping in unknown and complex environments. Google Cartographer, renowned for its real-time loop closure correction, ease of use, and comprehensive documentation, is employed to generate probabilistic occupancy grid representations for ground robots, enabling global path planning and situational awareness for human supervisors. Voxblox provides aerial robots with sufficient path clarity by generating a signed distance field (SDF) representation of 3D environments. For mapping, the MARBLE team leverages the open-source OctoMap package to create sparse voxel-based environmental representations and has developed a custom mapping package that incrementally integrates map segments from other robots through direct merge functionality during map creation, thereby conserving computational resources.  
In the cooperative planning module, MARBLE designs a metric-topological graph-based planner~\cite{ohradzansky2020reactive} and a continuous frontier-based planner to adapt to diverse underground environments. Additionally, MARBLE implements an internal market-based multi-agent coordination algorithm~\cite{riley2021assessment}, enabling robots to coordinate without relying on continuous bidirectional communication. Even under message loss or communication outages, robots autonomously proceed to their lowest-cost target points as if uncoordinated. This approach mitigates system-wide efficiency degradation caused by communication issues, ensuring stability and robustness in cooperative planning.  
For the communication module, MARBLE integrates communication beacons, a UDP-Mesh communication network, and prioritized message handling. Communication beacons extend communication ranges and facilitate map and target point sharing among robots. The UDP-Mesh network addresses challenges of packet loss and transient connectivity in complex environments by providing discovery services and reliable User Datagram Protocol (UDP) datagram transmission. Compatible with any underlying Layer 2 mesh network solution, it requires no modifications to Robot Operating System (ROS) nodes or diagnostic tools. This module enables efficient sharing of maps, target points, and other critical data to support multi-robot cooperative planning. It also prioritizes data types, ensuring high-value data (e.g., artifact discovery reports) are transmitted first, enhancing system efficiency and robustness.  
MARBLE optimizes heterogeneous robot task allocation through its multi-agent coordination algorithm, which facilitates data sharing and target deconfliction. For instance, the Clearpath Husky and Superdroid HD2 ground robots, with their high payload capacity and endurance, are suited for long-duration exploration in complex terrains. In contrast, the Lumenier QAV500 aerial robot excels in rapid reconnaissance and maneuvering through confined spaces. 

In addressing the numerous challenges faced by multi-robot cooperative exploration systems in comprehensive field applications, many teams have contributed unique solutions through the design of heterogeneous multi-robot cooperative exploration systems in the DARPA SubT Challenge. The technologies employed extensively reference existing mature engineering-proven solutions, thereby enhancing system stability, robustness, and operational efficiency. However, these representative works still commonly confront challenges including communication stability dependency in extreme field environments, high computational complexity, and insufficient adaptability to dynamic scenarios. Future efforts need to focus on optimizing lightweight communication protocols, enhancing computing capabilities, and exploring adaptive cooperative algorithms to balance system efficiency and robustness. 
\section{Open Problems and Possible Future Research Directions}
\label{sec: open}

Current multi-robot cooperative exploration systems leveraging multi-stage planning have shown strong performance in general environments. Therefore, this section focuses on open problems and future directions for exploration in extreme, dynamic, and large-scale environments. Additionally, it discusses the challenges of replacing multi-stage planning with learning-based planning in such systems.

\vspace{2mm}
\noindent\textbf{VII-A-1. Exploring in Extreme Environments}
\vspace{2mm}

Multi-robot cooperative exploration in extreme environments encounters challenges such as environmental hazards, platform heterogeneity, communication fragility, system resilience, and hardware limitations. Tackling these issues necessitates coordinated progress in resilient planning, adaptive communication architectures, multimodal sensing, and strategic robot team composition to enable reliable operation in GNSS-denied, unstructured, and high-risk scenarios.

\textbf{Environmental Challenges in Extreme Environments:} In extreme environments, such as underground caves, underwater environments, and mine caves, robots are highly prone to collisions during motion due to uneven surfaces, GNSS denial, and potential localization drift caused by uneven terrain and insufficient lighting, which can easily lead to multiple robot failures~\cite{azpurua2023survey}. 

\textbf{Robot Type Selection Challenges:} A key challenge for multi-robot cooperative exploration in extreme environments is selecting the appropriate robot types. Due to scenario complexity, it is crucial to deploy a well-balanced combination of platforms with complementary functionalities. In the DARPA SubT Challenge, many participating works featured multiple types of robots collaborating in exploration, with the majority achieving significant results~\cite{chung2023into}.

\textbf{Communication Constraints:} Communication constraints are a significant challenge. In extreme environments, reliable and continuous communication is frequently disrupted by complex geographical conditions. In highly constrained spaces, network architectures should incorporate cooperative strategies to optimize communication opportunities. However, no universally effective architecture currently exists to address bandwidth and range limitations. An ideal architecture should maximize efficiency in communication-constrained scenarios while maintaining a robust minimum performance threshold in highly degraded conditions~\cite{gielis2022critical}. 

\textbf{System Resilience Requirements:} In extreme scenarios, communication instability can result in complete connectivity loss and even failure of individual robots. Consequently, single-robot failures are inevitable. To address this, a robust cooperative planning framework is needed to sustain exploration capabilities despite communication loss or malfunctions. Such a framework should integrate mechanisms for communication recovery and dynamic replanning using the remaining operational robots, presenting a significant challenge for resilient cooperative planning development.

\vspace{2mm}
\noindent\textbf{VII-A-2. Exploring in Dynamic Environments}
\vspace{2mm}

Currently, many exploration tasks are set in static environments, such as forests~\cite{bartolomei2023fast}. However, when the environment to be explored is dynamic, such as in a factory where objects on conveyor belts are usually in motion, or on city streets with a large number of vehicles and people, tasks involving localization, mapping, and cooperative planning become particularly challenging~\cite{ullah2024mobile}.

\textbf{Map Representation:} Dynamic objects disrupt the feature association and tracking process. When features attached to moving objects change positions, disappear, or become occluded, the reliability of feature tracking is affected, leading to incorrect feature matching and subsequent erroneous pose estimation. Additionally, dynamic objects tend to cause constant map updates, resulting in the creation of inaccurate or transient map elements, which reduces the map’s usefulness for navigation and further environmental understanding, especially in large-scale scenarios~\cite{wang2024survey}. Consequently, in a dynamic environment, it is necessary to design a map representation method that can rapidly reconstruct the environment while maintaining high map accuracy to assist the system in robot localization. 

\textbf{Real-time Planning:} On the other hand, multi-robot systems need dynamic planning to avoid dynamic obstacles, which poses a challenge to the real-time design of the cooperative planning method of the system~\cite{placed2023survey}.

\vspace{2mm}
\noindent\textbf{VII-A-3. Exploring in Large-scale Environments}
\vspace{2mm}

\textbf{Compact Representation of Cooperative Information:} For general large-scale environments (such as cities~\cite{yin2023automerge} or forests~\cite{baril2021kilometer}), it is almost impossible to complete exploration tasks using only a single robot, so multi-robot systems are typically employed for exploration. Larger environments require more robots to improve exploration efficiency. For instance, when exploring a large-scale environment that necessitates expanding the number of robots to 100, centralized methods quickly encounter bottlenecks in terms of communication and processing at the base station. In such cases, distributed methods are relatively easier to design for scalable communication~\cite{ebadi2023present}. However, the amount of map data that needs to be exchanged among robots increases with the number of robots, leading to a decrease in the efficiency of map fusion and cooperative planning. Hence, a compact representation of the environment and tasks—both lightweight and rich in cooperative information—is essential.

\textbf{Cooperative Planning:} A major challenge in large-scale multi-robot systems is cooperative planning. As the number of robots increases, the complexity of task allocation algorithms also increases, which in turn increases the complexity of motion planning, resulting in a decrease in planning speed~\cite{poudel2022task}. This can lead to consistency issues in navigation directions, thereby imposing requirements on the system’s real-time performance.

\vspace{2mm}
\noindent\textbf{VII-A-4. Learning-based Planning}
\vspace{2mm}

Learning-based planning methods have demonstrated great potential in multi-robot coordination tasks, but their development is still in its early stages, with limited real-world applications. Currently, the field faces several key open issues:

\textbf{Simulator and Training Efficiency Issues:} Learning-based planning methods, especially reinforcement learning, heavily rely on interactive policy optimization with the environment, requiring simulators to have both high fidelity and fast computational capabilities. Online policy algorithms, such as Proximal Policy Optimization (PPO)~\cite{ schulman2017ppo }, require parallel processing of large volumes of sensor or image data. However, mainstream robotics simulators, such as Gazebo~\cite{koenig2004design}, lack support for parallel training, severely limiting training efficiency. Although researchers have developed high-parallel simulators, such as Flightmare~\cite{song2021flightmare} and Omnidrones~\cite{xu2024omnidrones}, which support reinforcement learning, they still have significant shortcomings in modeling scenarios for multi-robot outdoor collaborative exploration tasks. In addition, collaborative exploration tasks based on visual input face rendering speed bottlenecks, and the simulation-to-reality transfer issue has not yet been effectively addressed. It is worth noting that combining traditional environmental representation methods, such as grid maps, with reinforcement learning can effectively improve training speed and mitigate the reality gap through perceptual abstraction.

\textbf{Limitations of Action Space:} Existing reinforcement learning-based methods often use discrete action spaces to reduce training difficulty~\cite{liu2024heterogeneous, wang2023maddpg, liang2024hdplanner}, but this simplification has significant drawbacks in real-world physical systems. For example, in drone motion control, the motor throttle parameters are inherently continuous variables. Discretization not only reduces control accuracy but can also lead to system stability issues. Recent research has made progress in indoor scenarios using hierarchical architectures or continuous action space methods, but faces significant challenges in large-scale outdoor environments. Complex dynamic environments demand more flexible action strategies, yet balancing exploration efficiency and training stability in continuous action spaces remains difficult.

\textbf{Challenges of Large-Scale Clusters and Environments:} As the number of robots and the scale of the exploration space increase, the dimensionality of the state-action space grows exponentially, making it difficult for traditional methods to find near-optimal solutions. Existing research mainly reduces problem complexity through spatial decomposition~\cite{yang2023learning, zhang2022h2gnn, ding2018hierarchical} or local interaction constraints~\cite{tang2023autonomous, zhang2022multi}, but these methods may overlook the collaborative potential of ungrouped or distant robots. More importantly, existing simulators face a significant drop in computational efficiency when scaled to large robot clusters. The communication overhead of interaction decisions grows nonlinearly with the increase in the number of robots, severely limiting the scalability of real-world systems.

\textbf{Insufficient Generalization Ability:} Reinforcement learning (RL) strategies are prone to overfitting to specific training environments, leading to significantly worse performance than traditional planning methods when faced with unknown agent numbers or new scenarios. This limitation stems from the strong dependence of RL strategies on the parameter distribution of the training environment, while real-world applications often involve high uncertainty. Current improvements include using dynamic topology modeling frameworks (such as Transformer-based~\cite{MAANS} or graph neural network-based methods~\cite{MANTM, NeuralCoMapping}) for adaptable population scaling, constructing scene topological maps~\cite{lee2024multi} for environment-independent representations, and applying domain randomization techniques~\cite{mozian2020learning, horvath2022object} to enhance robustness to environmental disturbances. However, these methods are still limited to small- and medium-scale scenarios. In complex and dynamic large-scale environments, issues such as the real-time updating of dynamic topologies and the combinatorial complexity of randomization parameters hinder the generalization ability and robustness of the strategies, making it difficult to meet practical demands.
\section{CONCLUSIONS}
\label{sec: conclu}
This paper begins by defining the multi-robot cooperative exploration system, tracing its development history, and highlighting the growing trend toward modularization. It then presents a comprehensive classification of various algorithms and their variants across different modules, along with their practical applications. Additionally, the paper discusses the existing challenges in multi-robot cooperative exploration and suggests potential future research directions. By providing valuable insights for researchers and practitioners, this work serves as a reference to guide future advancements in the field. We believe that with continuous theoretical innovations and technological advancements, multi-robot cooperative exploration systems will achieve even greater progress.

\bibliographystyle{IEEEtran}
\bibliography{reference}

\begin{thebibliography}{100}
\providecommand{\url}[1]{#1}
\csname url@rmstyle\endcsname
\providecommand{\newblock}{\relax}
\providecommand{\bibinfo}[2]{#2}
\providecommand\BIBentrySTDinterwordspacing{\spaceskip=0pt\relax}
\providecommand\BIBentryALTinterwordstretchfactor{4}
\providecommand\BIBentryALTinterwordspacing{\spaceskip=\fontdimen2\font plus
\BIBentryALTinterwordstretchfactor\fontdimen3\font minus
  \fontdimen4\font\relax}
\providecommand\BIBforeignlanguage[2]{{%
\expandafter\ifx\csname l@#1\endcsname\relax
\typeout{** WARNING: IEEEtran.bst: No hyphenation pattern has been}%
\typeout{** loaded for the language `#1'. Using the pattern for}%
\typeout{** the default language instead.}%
\else
\language=\csname l@#1\endcsname
\fi
#2}}

\bibitem{Ackerman2022DarpaAutoRobo}
E.~Ackerman, ``Robots conquer the underground: What darpa's subterranean
  challenge means for the future of autonomous robots,'' \emph{IEEE Spectrum},
  vol.~59, no.~5, pp. 30--37, 2022.

\bibitem{Morrell2024CoSTAR}
B.~Morrell, K.~Otsu, A.~Agha, D.~D. Fan, S.-K. Kim, M.~F. Ginting, X.~Lei,
  J.~Edlund, S.~Fakoorian, A.~Bouman, F.~Chavez, T.~Kim, G.~J. Correa,
  M.~Saboia, A.~Santamaria-Navarro, B.~Lopez, B.~Kim, C.~Jung, M.~Sobue, O.~C.
  Peltzer, J.~Ott, R.~Trybula, T.~Touma, M.~Kaufmann, T.~S. Vaquero,
  T.~Pailevanian, M.~Palieri, Y.~Chang, A.~Reinke, M.~Anderson, F.~E.~T.
  Schöller, P.~Spieler, L.~M. Clark, A.~Archanian, K.~Chen, H.~Melikyan,
  A.~Dixit, H.~Delecki, D.~Pastor, B.~Ridge, N.~Marchal, J.~Uribe, S.~Dey,
  K.~Ebadi, K.~Coble, A.~N. Dimopoulos, V.~Thangavelu, V.~S. Varadharajan,
  N.~Palomo, A.~Rosinol, A.~Chatterjee, C.~Kanellakis, B.~Lindqvist, M.~Corah,
  K.~Strickland, R.~Stonebraker, M.~Milano, C.~E. Denniston, S.~Sahnoune,
  T.~Claudet, S.~Lee, G.~Salhotra, E.~Terry, R.~Musuku, R.~Schmid, T.~Tran,
  A.~Kourchians, J.~Schachter, H.~Azpurua, L.~Resende, A.~Kalantari, J.~Nash,
  J.~Lee, C.~Patterson, J.~G. Blank, K.~Patath, Y.~Kubo, R.~Alimo,
  Y.~Almalioglu, A.~Curtis, J.~Sly, T.~Wells, N.~T. Ho, M.~Kochenderfer,
  G.~Beltrame, G.~Nikolakopoulos, D.~Shim, L.~Carlone, Burdick, and Joel, ``An
  addendum to nebula: Toward extending team costar’s solution to larger scale
  environments,'' \emph{IEEE Transactions on Field Robotics}, vol.~1, pp.
  476--526, 2024.

\bibitem{CSIRO2022Heterogeneous}
N.~Hudson, F.~Talbot, M.~Cox, J.~Williams, T.~Hines, A.~Pitt, B.~Wood,
  D.~Frousheger, K.~Lo~Surdo, T.~Molnar, R.~Steindl, M.~Wildie, I.~Sa,
  N.~Kottege, K.~Stepanas, E.~Hernández, G.~Catt, W.~Docherty, and B.~Tidd,
  ``Heterogeneous ground and air platforms, homogeneous sensing: Team csiro
  data61’s approach to the darpa subterranean challenge,'' \emph{Field
  Robotics}, vol.~2, pp. 595--636, 03 2022.

\bibitem{Ebadi2020LAMP}
K.~Ebadi, Y.~Chang, M.~Palieri, A.~Stephens, A.~Hatteland, E.~Heiden,
  A.~Thakur, N.~Funabiki, B.~Morrell, S.~Wood, L.~Carlone, and A.-a.
  Agha-mohammadi, ``Lamp: Large-scale autonomous mapping and positioning for
  exploration of perceptually-degraded subterranean environments,'' in
  \emph{2020 IEEE International Conference on Robotics and Automation (ICRA)},
  2020, pp. 80--86.

\bibitem{zhou2023racer}
B.~Zhou, H.~Xu, and S.~Shen, ``Racer: Rapid collaborative exploration with a
  decentralized multi-uav system,'' \emph{IEEE Transactions on Robotics},
  vol.~39, no.~3, pp. 1816--1835, 2023.

\bibitem{door-slam}
P.-Y. Lajoie, B.~Ramtoula, Y.~Chang, L.~Carlone, and G.~Beltrame, ``Door-slam:
  Distributed, online, and outlier resilient slam for robotic teams,''
  \emph{IEEE Robotics and Automation Letters}, vol.~5, no.~2, pp. 1656--1663,
  2020.

\bibitem{queralta2020collaborative}
J.~P. Queralta, J.~Taipalmaa, B.~C. Pullinen, V.~K. Sarker, T.~N. Gia,
  H.~Tenhunen, M.~Gabbouj, J.~Raitoharju, and T.~Westerlund, ``Collaborative
  multi-robot search and rescue: Planning, coordination, perception, and active
  vision,'' \emph{Ieee Access}, vol.~8, pp. 191\,617--191\,643, 2020.

\bibitem{roman2006framework}
I.~Roman-Ballesteros and C.~F. Pfeiffer, ``A framework for cooperative
  multi-robot surveillance tasks,'' in \emph{Electronics, Robotics and
  Automotive Mechanics Conference (CERMA'06)}, vol.~2.\hskip 1em plus 0.5em
  minus 0.4em\relax IEEE, 2006, pp. 163--170.

\bibitem{xu2024Cost}
X.~Xu, M.~Cao, S.~Yuan, T.~H. Nguyen, T.-M. Nguyen, and L.~Xie, ``A
  cost-effective cooperative exploration and inspection strategy for
  heterogeneous aerial system,'' in \emph{2024 IEEE 18th International
  Conference on Control \& Automation (ICCA)}, 2024, pp. 673--678.

\bibitem{miller2020mine}
I.~D. Miller, F.~Cladera, A.~Cowley, S.~S. Shivakumar, E.~S. Lee,
  L.~Jarin-Lipschitz, A.~Bhat, N.~Rodrigues, A.~Zhou, A.~Cohen, \emph{et~al.},
  ``Mine tunnel exploration using multiple quadrupedal robots,'' \emph{IEEE
  Robotics and Automation Letters}, vol.~5, no.~2, pp. 2840--2847, 2020.

\bibitem{yamauchi1997frontier}
B.~Yamauchi, ``A frontier-based approach for autonomous exploration,'' in
  \emph{Proceedings 1997 IEEE International Symposium on Computational
  Intelligence in Robotics and Automation CIRA'97.'Towards New Computational
  Principles for Robotics and Automation'}.\hskip 1em plus 0.5em minus
  0.4em\relax IEEE, 1997, pp. 146--151.

\bibitem{ebadi2023present}
K.~Ebadi, L.~Bernreiter, H.~Biggie, G.~Catt, Y.~Chang, A.~Chatterjee, C.~E.
  Denniston, S.-P. Desch{\^e}nes, K.~Harlow, S.~Khattak, \emph{et~al.},
  ``Present and future of slam in extreme environments: The darpa subt
  challenge,'' \emph{IEEE Transactions on Robotics}, 2023.

\bibitem{Abujabal_Fareh_Sinan_Baziyad_Bettayeb_2023}
N.~Abujabal, R.~Fareh, S.~Sinan, M.~Baziyad, and M.~Bettayeb, ``A comprehensive
  review of the latest path planning developments for multi-robot formation
  systems,'' \emph{Robotica}, vol.~41, no.~7, p. 2079–2104, 2023.

\bibitem{MADRIDANO2021114660}
\BIBentryALTinterwordspacing
Ángel Madridano, A.~Al-Kaff, D.~Martín, and A.~{de la Escalera}, ``Trajectory
  planning for multi-robot systems: Methods and applications,'' \emph{Expert
  Systems with Applications}, vol. 173, p. 114660, 2021. [Online]. Available:
  \url{https://www.sciencedirect.com/science/article/pii/S0957417421001019}
\BIBentrySTDinterwordspacing

\bibitem{khamis2015multi}
A.~Khamis, A.~Hussein, and A.~Elmogy, ``Multi-robot task allocation: A review
  of the state-of-the-art,'' \emph{Cooperative robots and sensor networks
  2015}, pp. 31--51, 2015.

\bibitem{skaltsis2021task}
G.~M. Skaltsis, H.-S. Shin, and A.~Tsourdos, ``A survey of task allocation
  techniques in mas,'' in \emph{2021 International Conference on Unmanned
  Aircraft Systems (ICUAS)}, 2021, pp. 488--497.

\bibitem{comm_survey}
F.~Amigoni, J.~Banfi, and N.~Basilico, ``Multirobot exploration of
  communication-restricted environments: A survey,'' \emph{IEEE Intelligent
  Systems}, vol.~32, no.~6, pp. 48--57, 2017.

\bibitem{mataric1992integration}
M.~Mataric, ``Integration of representation into goal-driven behavior-based
  robots,'' \emph{IEEE Transactions on Robotics and Automation}, vol.~8, no.~3,
  pp. 304--312, 1992.

\bibitem{yamauchi1996spatial}
B.~Yamauchi and R.~Beer, ``Spatial learning for navigation in dynamic
  environments,'' \emph{IEEE Transactions on Systems, Man, and Cybernetics,
  Part B (Cybernetics)}, vol.~26, no.~3, pp. 496--505, 1996.

\bibitem{connell1993rapid}
J.~H. Connell and S.~Mahadevan, ``Rapid task learning for real robots,'' in
  \emph{Robot Learning}.\hskip 1em plus 0.5em minus 0.4em\relax Springer, pp.
  105--139.

\bibitem{thrun1996integrating}
S.~Thrun and A.~B{\"u}cken, ``Integrating grid-based and topological maps for
  mobile robot navigation,'' in \emph{Proceedings of the national conference on
  artificial intelligence}, 1996, pp. 944--951.

\bibitem{yamauchi1998frontier}
B.~Yamauchi, ``Frontier-based exploration using multiple robots,'' in
  \emph{Proceedings of the second international conference on Autonomous
  agents}, 1998, pp. 47--53.

\bibitem{burgard2005coordinated}
W.~Burgard, M.~Moors, C.~Stachniss, and F.~E. Schneider, ``Coordinated
  multi-robot exploration,'' \emph{IEEE Transactions on robotics}, vol.~21,
  no.~3, pp. 376--386, 2005.

\bibitem{butzke2011planning}
J.~Butzke and M.~Likhachev, ``Planning for multi-robot exploration with
  multiple objective utility functions,'' in \emph{2011 IEEE/RSJ International
  Conference on Intelligent Robots and Systems}.\hskip 1em plus 0.5em minus
  0.4em\relax IEEE, 2011, pp. 3254--3259.

\bibitem{wurm2008coordinated}
K.~M. Wurm, C.~Stachniss, and W.~Burgard, ``Coordinated multi-robot exploration
  using a segmentation of the environment,'' in \emph{2008 IEEE/RSJ
  International Conference on Intelligent Robots and Systems}.\hskip 1em plus
  0.5em minus 0.4em\relax IEEE, 2008, pp. 1160--1165.

\bibitem{gonzalez2002navigation}
H.~H. Gonz{\'a}lez-Banos and J.-C. Latombe, ``Navigation strategies for
  exploring indoor environments,'' \emph{The International Journal of Robotics
  Research}, vol.~21, no. 10-11, pp. 829--848, 2002.

\bibitem{connolly1985determination}
C.~Connolly, ``The determination of next best views,'' in \emph{Proceedings.
  1985 IEEE International Conference on Robotics and Automation}, vol.~2.\hskip
  1em plus 0.5em minus 0.4em\relax IEEE, 1985, pp. 432--435.

\bibitem{lluvia2021active}
I.~Lluvia, E.~Lazkano, and A.~Ansuategi, ``Active mapping and robot
  exploration: A survey,'' \emph{Sensors}, vol.~21, no.~7, p. 2445, 2021.

\bibitem{gerkey2004formal}
B.~P. Gerkey and M.~J. Matari{\'c}, ``A formal analysis and taxonomy of task
  allocation in multi-robot systems,'' \emph{The International journal of
  robotics research}, vol.~23, no.~9, pp. 939--954, 2004.

\bibitem{garaffa2021reinforcement}
L.~C. Garaffa, M.~Basso, A.~A. Konzen, and E.~P. de~Freitas, ``Reinforcement
  learning for mobile robotics exploration: A survey,'' \emph{IEEE Transactions
  on Neural Networks and Learning Systems}, vol.~34, no.~8, pp. 3796--3810,
  2021.

\bibitem{swarm-lio}
F.~Zhu, Y.~Ren, F.~Kong, H.~Wu, S.~Liang, N.~Chen, W.~Xu, and F.~Zhang,
  ``Swarm-lio: Decentralized swarm lidar-inertial odometry,'' in \emph{2023
  IEEE International Conference on Robotics and Automation (ICRA)}, 2023, pp.
  3254--3260.

\bibitem{Rouek2021SystemFM}
\BIBentryALTinterwordspacing
T.~Rou{\v{c}}ek, M.~Pecka, P.~{\v{C}}{\'i}{\v{z}}ek,
  T.~Pet{\v{r}}{\'i}{\v{c}}ek, J.~Bayer, V.~{\v{S}}alansk{\'y}, T.~Azayev,
  D.~He{\v{r}}t, M.~Petrl{\'i}k, T.~B{\'a}{\v{c}}a, V.~Spurn{\'y},
  V.~Kr{\'a}tk{\'y}, P.~Petr{\'a}{\v{c}}ek, D.~Baril, M.~Vaidis, V.~Kubelka,
  F.~Pomerleau, J.~Faigl, K.~Zimmermann, M.~Saska, T.~Svoboda, and
  T.~Krajn{\'i}k, ``System for multi-robotic exploration of underground
  environments ctu-cras-norlab in the darpa subterranean challenge,''
  \emph{Field Robotics}, vol.~2, pp. 1779--1818, 2021. [Online]. Available:
  \url{https://api.semanticscholar.org/CorpusID:238634793}
\BIBentrySTDinterwordspacing

\bibitem{Lee2021REAL}
E.~M. Lee, J.~Choi, H.~Lim, and H.~Myung, ``Real: Rapid exploration with active
  loop-closing toward large-scale 3d mapping using uavs,'' in \emph{2021
  IEEE/RSJ International Conference on Intelligent Robots and Systems (IROS)},
  2021, pp. 4194--4198.

\bibitem{yu2020mapmerging}
\BIBentryALTinterwordspacing
S.~Yu, C.~Fu, A.~K. Gostar, and M.~Hu, ``A review on map-merging methods for
  typical map types in multiple-ground-robot slam solutions,'' \emph{Sensors},
  vol.~20, no.~23, 2020. [Online]. Available:
  \url{https://www.mdpi.com/1424-8220/20/23/6988}
\BIBentrySTDinterwordspacing

\bibitem{dcl-slam}
S.~Zhong, Y.~Qi, Z.~Chen, J.~Wu, H.~Chen, and M.~Liu, ``Dcl-slam: A distributed
  collaborative lidar slam framework for a robotic swarm,'' \emph{IEEE Sensors
  Journal}, vol.~24, no.~4, pp. 4786--4797, 2024.

\bibitem{ccm-slam}
\BIBentryALTinterwordspacing
P.~Schmuck and M.~Chli, ``Ccm-slam: Robust and efficient centralized
  collaborative monocular simultaneous localization and mapping for robotic
  teams,'' \emph{Journal of Field Robotics}, vol.~36, no.~4, pp. 763--781,
  2019. [Online]. Available:
  \url{https://onlinelibrary.wiley.com/doi/abs/10.1002/rob.21854}
\BIBentrySTDinterwordspacing

\bibitem{netvlad}
R.~Arandjelović, P.~Gronat, A.~Torii, T.~Pajdla, and J.~Sivic, ``Netvlad: Cnn
  architecture for weakly supervised place recognition,'' \emph{IEEE
  Transactions on Pattern Analysis and Machine Intelligence}, vol.~40, no.~6,
  pp. 1437--1451, 2018.

\bibitem{scancontext}
G.~Kim and A.~Kim, ``Scan context: Egocentric spatial descriptor for place
  recognition within 3d point cloud map,'' in \emph{2018 IEEE/RSJ International
  Conference on Intelligent Robots and Systems (IROS)}, 2018, pp. 4802--4809.

\bibitem{cieslewski2017visualpr}
T.~Cieslewski and D.~Scaramuzza, ``Efficient decentralized visual place
  recognition using a distributed inverted index,'' \emph{IEEE Robotics and
  Automation Letters}, vol.~2, no.~2, pp. 640--647, 2017.

\bibitem{mangelson2018PCM}
J.~G. Mangelson, D.~Dominic, R.~M. Eustice, and R.~Vasudevan, ``Pairwise
  consistent measurement set maximization for robust multi-robot map merging,''
  in \emph{2018 IEEE International Conference on Robotics and Automation
  (ICRA)}, 2018, pp. 2916--2923.

\bibitem{Tian2019DistributedCC}
\BIBentryALTinterwordspacing
Y.~Tian, K.~Khosoussi, D.~M. Rosen, and J.~P. How, ``Distributed certifiably
  correct pose-graph optimization,'' \emph{IEEE Transactions on Robotics},
  vol.~37, pp. 2137--2156, 2019. [Online]. Available:
  \url{https://api.semanticscholar.org/CorpusID:221274944}
\BIBentrySTDinterwordspacing

\bibitem{D2SLAM2024}
H.~Xu, P.~Liu, X.~Chen, and S.~Shen, ``$d^{2}$slam: Decentralized and
  distributed collaborative visual-inertial slam system for aerial swarm,''
  \emph{IEEE Transactions on Robotics}, vol.~40, pp. 3445--3464, 2024.

\bibitem{omni-swarm}
H.~Xu, Y.~Zhang, B.~Zhou, L.~Wang, X.~Yao, G.~Meng, and S.~Shen, ``Omni-swarm:
  A decentralized omnidirectional visual–inertial–uwb state estimation
  system for aerial swarms,'' \emph{IEEE Transactions on Robotics}, vol.~38,
  no.~6, pp. 3374--3394, 2022.

\bibitem{Niu2022qrtag}
G.~Niu, Q.~Yang, Y.~Gao, and M.-O. Pun, ``Vision-based autonomous landing for
  unmanned aerial and ground vehicles cooperative systems,'' \emph{IEEE
  Robotics and Automation Letters}, vol.~7, no.~3, pp. 6234--6241, 2022.

\bibitem{H-SwarmLoc2023Wang}
\BIBentryALTinterwordspacing
H.~Wang, X.~Chen, Y.~Cheng, C.~Wu, F.~Dang, and X.~Chen, ``H-swarmloc:
  Efficient scheduling for localization of heterogeneous mav swarm with deep
  reinforcement learning,'' in \emph{Proceedings of the 20th ACM Conference on
  Embedded Networked Sensor Systems}, ser. SenSys '22.\hskip 1em plus 0.5em
  minus 0.4em\relax New York, NY, USA: Association for Computing Machinery,
  2023, p. 1148–1154. [Online]. Available:
  \url{https://doi.org/10.1145/3560905.3568432}
\BIBentrySTDinterwordspacing

\bibitem{TransformLoc2024Wang}
H.~Wang, J.~Xu, C.~Zhao, Z.~Lu, Y.~Cheng, X.~Chen, X.-P. Zhang, Y.~Liu, and
  X.~Chen, ``Transformloc: Transforming mavs into mobile localization
  infrastructures in heterogeneous swarms,'' in \emph{IEEE INFOCOM 2024 - IEEE
  Conference on Computer Communications}, 2024, pp. 1101--1110.

\bibitem{Zhu2024swarm-lio2}
\BIBentryALTinterwordspacing
F.~Zhu, Y.~Ren, L.~Yin, F.~Kong, Q.~Liu, R.~Xue, W.~Liu, Y.~Cai, G.~Lu, H.~Li,
  and F.~Zhang, ``Swarm-lio2: Decentralized, efficient lidar-inertial odometry
  for uav swarms,'' 2024. [Online]. Available:
  \url{https://arxiv.org/abs/2409.17798}
\BIBentrySTDinterwordspacing

\bibitem{Zhou2022flyinwild}
X.~Zhou, X.~Wen, Z.~Wang, Y.~Gao, H.~Li, Q.~Wang, T.~Yang, H.~Lu, Y.~Cao,
  C.~Xu, and F.~Gao, ``Swarm of micro flying robots in the wild,''
  \emph{Science Robotics}, vol.~7, no.~66, p. eabm5954, 2022.

\bibitem{bochkovskiy2020yolov4}
A.~Bochkovskiy, C.-Y. Wang, and H.-Y.~M. Liao, ``Yolov4: Optimal speed and
  accuracy of object detection,'' \emph{arXiv preprint arXiv:2004.10934}, 2020.

\bibitem{cave2021petracek}
P.~Petráček, V.~Krátký, M.~Petrlík, T.~Báča, R.~Kratochvíl, and
  M.~Saska, ``Large-scale exploration of cave environments by unmanned aerial
  vehicles,'' \emph{IEEE Robotics and Automation Letters}, vol.~6, no.~4, pp.
  7596--7603, 2021.

\bibitem{Nguyen2022virloc}
T.~H. Nguyen, T.-M. Nguyen, and L.~Xie, ``Flexible and resource-efficient
  multi-robot collaborative visual-inertial-range localization,'' \emph{IEEE
  Robotics and Automation Letters}, vol.~7, no.~2, pp. 928--935, 2022.

\bibitem{ego-swarm}
X.~Zhou, J.~Zhu, H.~Zhou, C.~Xu, and F.~Gao, ``Ego-swarm: A fully autonomous
  and decentralized quadrotor swarm system in cluttered environments,'' in
  \emph{2021 IEEE International Conference on Robotics and Automation (ICRA)},
  2021, pp. 4101--4107.

\bibitem{SegMap}
\BIBentryALTinterwordspacing
R.~Dub\'{e}, A.~Cramariuc, D.~Dugas, H.~Sommer, M.~Dymczyk, J.~Nieto,
  R.~Siegwart, and C.~Cadena, ``Segmap: Segment-based mapping and localization
  using data-driven descriptors,'' \emph{Int. J. Rob. Res.}, vol.~39, no.
  2–3, p. 339–355, Mar. 2020. [Online]. Available:
  \url{https://doi.org/10.1177/0278364919863090}
\BIBentrySTDinterwordspacing

\bibitem{localdescriptor1999Lowe}
D.~Lowe, ``Object recognition from local scale-invariant features,'' in
  \emph{Proceedings of the Seventh IEEE International Conference on Computer
  Vision}, vol.~2, 1999, pp. 1150--1157 vol.2.

\bibitem{localdescriptor2015Prakhya}
S.~M. Prakhya, B.~Liu, and W.~Lin, ``B-shot: A binary feature descriptor for
  fast and efficient keypoint matching on 3d point clouds,'' in \emph{2015
  IEEE/RSJ International Conference on Intelligent Robots and Systems (IROS)},
  2015, pp. 1929--1934.

\bibitem{LiDARIris2020Wang}
Y.~Wang, Z.~Sun, C.-Z. Xu, S.~E. Sarma, J.~Yang, and H.~Kong, ``Lidar iris for
  loop-closure detection,'' in \emph{2020 IEEE/RSJ International Conference on
  Intelligent Robots and Systems (IROS)}, 2020, pp. 5769--5775.

\bibitem{SIFT2016}
M.~R. Kirchner, ``Automatic thresholding of sift descriptors,'' in \emph{2016
  IEEE International Conference on Image Processing (ICIP)}, 2016, pp.
  291--295.

\bibitem{ORB2011}
E.~Rublee, V.~Rabaud, K.~Konolige, and G.~Bradski, ``Orb: An efficient
  alternative to sift or surf,'' in \emph{2011 International Conference on
  Computer Vision}, 2011, pp. 2564--2571.

\bibitem{Uy2018PointNetVLADDP}
\BIBentryALTinterwordspacing
M.~A. Uy and G.~H. Lee, ``Pointnetvlad: Deep point cloud based retrieval for
  large-scale place recognition,'' \emph{2018 IEEE/CVF Conference on Computer
  Vision and Pattern Recognition}, pp. 4470--4479, 2018. [Online]. Available:
  \url{https://api.semanticscholar.org/CorpusID:4805033}
\BIBentrySTDinterwordspacing

\bibitem{Chen2020OverlapNetLC}
\BIBentryALTinterwordspacing
X.~Chen, T.~L{\"a}be, A.~Milioto, T.~R{\"o}hling, O.~Vysotska, A.~Haag,
  J.~Behley, and C.~Stachniss, ``Overlapnet: Loop closing for lidar-based
  slam,'' \emph{ArXiv}, vol. abs/2105.11344, 2020. [Online]. Available:
  \url{https://api.semanticscholar.org/CorpusID:219178115}
\BIBentrySTDinterwordspacing

\bibitem{AutoMerge2023}
P.~Yin, S.~Zhao, H.~Lai, R.~Ge, J.~Zhang, H.~Choset, and S.~Scherer,
  ``Automerge: A framework for map assembling and smoothing in city-scale
  environments,'' \emph{IEEE Transactions on Robotics}, vol.~39, no.~5, pp.
  3686--3704, 2023.

\bibitem{Visualdata2015Tardioli}
D.~Tardioli, E.~Montijano, and A.~R. Mosteo, ``Visual data association in
  narrow-bandwidth networks,'' in \emph{2015 IEEE/RSJ International Conference
  on Intelligent Robots and Systems (IROS)}, 2015, pp. 2572--2577.

\bibitem{FullImageDescriptors2017Cieslewski}
T.~Cieslewski and D.~Scaramuzza, ``Efficient decentralized visual place
  recognition from full-image descriptors,'' in \emph{2017 International
  Symposium on Multi-Robot and Multi-Agent Systems (MRS)}, 2017, pp. 78--82.

\bibitem{DistributedInvertedIndex2017Cieslewski}
------, ``Efficient decentralized visual place recognition using a distributed
  inverted index,'' \emph{IEEE Robotics and Automation Letters}, vol.~2, no.~2,
  pp. 640--647, 2017.

\bibitem{LAMP2}
Y.~Chang, K.~Ebadi, C.~E. Denniston, M.~F. Ginting, A.~Rosinol, A.~Reinke,
  M.~Palieri, J.~Shi, A.~Chatterjee, B.~Morrell, A.-a. Agha-mohammadi, and
  L.~Carlone, ``Lamp 2.0: A robust multi-robot slam system for operation in
  challenging large-scale underground environments,'' \emph{IEEE Robotics and
  Automation Letters}, vol.~7, no.~4, pp. 9175--9182, 2022.

\bibitem{3DLiDARSonlineMRSLAM2017Dubé}
R.~Dubé, A.~Gawel, H.~Sommer, J.~Nieto, R.~Siegwart, and C.~Cadena, ``An
  online multi-robot slam system for 3d lidars,'' in \emph{2017 IEEE/RSJ
  International Conference on Intelligent Robots and Systems (IROS)}, 2017, pp.
  1004--1011.

\bibitem{H-DrunkWalk2020Chen}
\BIBentryALTinterwordspacing
X.~Chen, C.~Ruiz, S.~Zeng, L.~Gao, A.~Purohit, S.~Carpin, and P.~Zhang,
  ``H-drunkwalk: Collaborative and adaptive navigation for heterogeneous mav
  swarm,'' \emph{ACM Trans. Sen. Netw.}, vol.~16, no.~2, Apr. 2020. [Online].
  Available: \url{https://doi.org/10.1145/3382094}
\BIBentrySTDinterwordspacing

\bibitem{DrunkWalk2015Chen}
X.~Chen, A.~Purohit, C.~R. Dominguez, S.~Carpin, and P.~Zhang, ``Drunkwalk:
  Collaborative and adaptive planning for navigation of micro-aerial sensor
  swarms,'' in \emph{Proceedings of the 13th ACM Conference on Embedded
  Networked Sensor Systems}, ser. SenSys '15.\hskip 1em plus 0.5em minus
  0.4em\relax New York, NY, USA: Association for Computing Machinery, 2015, p.
  295–308.

\bibitem{zhang2022mr}
Z.~Zhang, J.~Yu, J.~Tang, Y.~Xu, and Y.~Wang, ``Mr-topomap: Multi-robot
  exploration based on topological map in communication restricted
  environment,'' \emph{IEEE Robotics and Automation Letters}, vol.~7, no.~4,
  pp. 10\,794--10\,801, 2022.

\bibitem{Dong2022mr-gmm}
H.~Dong, J.~Yu, Y.~Xu, Z.~Xu, Z.~Shen, J.~Tang, Y.~Shen, and Y.~Wang,
  ``Mr-gmmapping: Communication efficient multi-robot mapping system via
  gaussian mixture model,'' \emph{IEEE Robotics and Automation Letters},
  vol.~7, no.~2, pp. 3294--3301, 2022.

\bibitem{MUITARE2023}
J.~Yan, X.~Lin, Z.~Ren, S.~Zhao, J.~Yu, C.~Cao, P.~Yin, J.~Zhang, and
  S.~Scherer, ``Mui-tare: Cooperative multi-agent exploration with unknown
  initial position,'' \emph{IEEE Robotics and Automation Letters}, vol.~8,
  no.~7, pp. 4299--4306, 2023.

\bibitem{LVCPLT2024Jian}
\BIBentryALTinterwordspacing
Z.~Jian, Q.~Li, S.~Zheng, X.~Wang, and X.~Chen, ``Lvcp: Lidar-vision tightly
  coupled collaborative real-time relative positioning,'' \emph{ArXiv}, vol.
  abs/2407.10782, 2024. [Online]. Available:
  \url{https://api.semanticscholar.org/CorpusID:271212580}
\BIBentrySTDinterwordspacing

\bibitem{Andersone2019heterogeneousmap}
\BIBentryALTinterwordspacing
I.~Andersone, ``Heterogeneous map merging: State of the art,'' \emph{Robotics},
  vol.~8, no.~3, 2019. [Online]. Available:
  \url{https://www.mdpi.com/2218-6581/8/3/74}
\BIBentrySTDinterwordspacing

\bibitem{Andrew2006particle}
A.~Howard, ``Multi-robot simultaneous localization and mapping using particle
  filters,'' \emph{The International Journal of Robotics Research}, vol.~25,
  no.~12, pp. 1243--1256, 2006.

\bibitem{Li2012gridmerge}
H.~Li and F.~Nashashibi, ``A new method for occupancy grid maps merging:
  Application to multi-vehicle cooperative local mapping and moving object
  detection in outdoor environment,'' in \emph{2012 12th International
  Conference on Control Automation Robotics and Vision (ICARCV)}, 2012, pp.
  632--637.

\bibitem{Lin2021gridmerge}
\BIBentryALTinterwordspacing
Z.~Lin, J.~Zhu, Z.~Jiang, Y.~Li, Y.~Li, and Z.~Li, ``Merging grid maps in
  diverse resolutions by the context-based descriptor,'' \emph{ACM Trans.
  Internet Technol.}, vol.~21, no.~4, July 2021. [Online]. Available:
  \url{https://doi.org/10.1145/3403948}
\BIBentrySTDinterwordspacing

\bibitem{Gawel2016vision-laser}
A.~Gawel, T.~Cieslewski, R.~Dubé, M.~Bosse, R.~Siegwart, and J.~Nieto,
  ``Structure-based vision-laser matching,'' in \emph{2016 IEEE/RSJ
  International Conference on Intelligent Robots and Systems (IROS)}, 2016, pp.
  182--188.

\bibitem{Lajoie2021TowardsCS}
\BIBentryALTinterwordspacing
P.-Y. Lajoie, B.~Ramtoula, F.~Wu, and G.~Beltrame, ``Towards collaborative
  simultaneous localization and mapping: a survey of the current research
  landscape,'' \emph{Field Robotics}, vol.~2, pp. 971--1000, 2021. [Online].
  Available: \url{https://api.semanticscholar.org/CorpusID:237213256}
\BIBentrySTDinterwordspacing

\bibitem{cimurs2021goal}
R.~Cimurs, I.~H. Suh, and J.~H. Lee, ``Goal-driven autonomous exploration
  through deep reinforcement learning,'' \emph{IEEE Robotics and Automation
  Letters}, vol.~7, no.~2, pp. 730--737, 2021.

\bibitem{vincent2008distributed}
R.~Vincent, D.~Fox, J.~Ko, K.~Konolige, B.~Limketkai, B.~Morisset, C.~Ortiz,
  D.~Schulz, and B.~Stewart, ``Distributed multirobot exploration, mapping, and
  task allocation,'' \emph{Annals of Mathematics and Artificial Intelligence},
  vol.~52, pp. 229--255, 2008.

\bibitem{alitappeh2022multi}
R.~J. Alitappeh and K.~Jeddisaravi, ``Multi-robot exploration in task
  allocation problem,'' \emph{Applied Intelligence}, vol.~52, no.~2, pp.
  2189--2211, 2022.

\bibitem{sariel2006efficient}
S.~Sariel and T.~R. Balch, ``Efficient bids on task allocation for multi-robot
  exploration.'' in \emph{FLAIRS}, 2006, pp. 116--121.

\bibitem{chen2019end}
Z.~Chen, B.~Subagdja, and A.-H. Tan, ``End-to-end deep reinforcement learning
  for multi-agent collaborative exploration,'' in \emph{2019 IEEE International
  Conference on Agents (ICA)}.\hskip 1em plus 0.5em minus 0.4em\relax IEEE,
  2019, pp. 99--102.

\bibitem{MAANS}
C.~Yu, X.~Yang, J.~Gao, H.~Yang, Y.~Wang, and Y.~Wu, ``Learning efficient
  multi-agent cooperative visual exploration,'' in \emph{European Conference on
  Computer Vision}.\hskip 1em plus 0.5em minus 0.4em\relax Springer, 2022, pp.
  497--515.

\bibitem{liang2024hdplanner}
J.~Liang, Y.~Cao, Y.~Ma, H.~Zhao, and G.~Sartoretti, ``Hdplanner: Advancing
  autonomous deployments in unknown environments through hierarchical decision
  networks,'' \emph{IEEE Robotics and Automation Letters}, 2024.

\bibitem{charrow2015information}
B.~Charrow, G.~Kahn, S.~Patil, S.~Liu, K.~Goldberg, P.~Abbeel, N.~Michael, and
  V.~Kumar, ``Information-theoretic planning with trajectory optimization for
  dense 3d mapping.'' in \emph{Robotics: Science and Systems}, vol.~11, 2015,
  pp. 3--12.

\bibitem{meng2017two}
Z.~Meng, H.~Qin, Z.~Chen, X.~Chen, H.~Sun, F.~Lin, and M.~H. Ang, ``A two-stage
  optimized next-view planning framework for 3-d unknown environment
  exploration, and structural reconstruction,'' \emph{IEEE Robotics and
  Automation Letters}, vol.~2, no.~3, pp. 1680--1687, 2017.

\bibitem{aurenhammer1991voronoi}
F.~Aurenhammer, ``Voronoi diagrams—a survey of a fundamental geometric data
  structure,'' \emph{ACM Computing Surveys (CSUR)}, vol.~23, no.~3, pp.
  345--405, 1991.

\bibitem{xu2005survey}
R.~Xu and D.~Wunsch, ``Survey of clustering algorithms,'' \emph{IEEE
  Transactions on neural networks}, vol.~16, no.~3, pp. 645--678, 2005.

\bibitem{karapetyan2017efficient}
N.~Karapetyan, K.~Benson, C.~McKinney, P.~Taslakian, and I.~Rekleitis,
  ``Efficient multi-robot coverage of a known environment,'' in \emph{2017
  IEEE/RSJ International Conference on Intelligent Robots and Systems
  (IROS)}.\hskip 1em plus 0.5em minus 0.4em\relax IEEE, 2017, pp. 1846--1852.

\bibitem{bundy1984breadth}
A.~Bundy and L.~Wallen, ``Breadth-first search,'' \emph{Catalogue of artificial
  intelligence tools}, pp. 13--13, 1984.

\bibitem{galceran2013survey}
E.~Galceran and M.~Carreras, ``A survey on coverage path planning for
  robotics,'' \emph{Robotics and Autonomous systems}, vol.~61, no.~12, pp.
  1258--1276, 2013.

\bibitem{dong2024fast}
Q.~Dong, H.~Xi, S.~Zhang, Q.~Bi, T.~Li, Z.~Wang, and X.~Zhang, ``Fast and
  communication-efficient multi-uav exploration via voronoi partition on
  dynamic topological graph,'' \emph{arXiv preprint arXiv:2408.05808}, 2024.

\bibitem{zhang2024leces}
T.~Zhang, H.~Shen, Y.~Yin, J.~Xu, J.~Yu, and Y.~Pan, ``Leces: A low-bandwidth
  and efficient collaborative exploration system with distributed multi-uav,''
  \emph{IEEE Robotics and Automation Letters}, 2024.

\bibitem{likas2003global}
A.~Likas, N.~Vlassis, and J.~J. Verbeek, ``The global k-means clustering
  algorithm,'' \emph{Pattern recognition}, vol.~36, no.~2, pp. 451--461, 2003.

\bibitem{oncan2009comparative}
T.~{\"O}ncan, I.~K. Alt{\i}nel, and G.~Laporte, ``A comparative analysis of
  several asymmetric traveling salesman problem formulations,'' \emph{Computers
  \& Operations Research}, vol.~36, no.~3, pp. 637--654, 2009.

\bibitem{lajoie2022towards}
P.-Y. Lajoie, B.~Ramtoula, F.~Wu, and G.~Beltrame, ``Towards collaborative
  simultaneous localization and mapping: A survey of the current research
  landscape,'' \emph{Field Robotics}, vol.~2, pp. 971--1000, 2022.

\bibitem{Huang2024Autonomous}
Y.~Huang, X.~Lin, and B.~Englot, ``Multi-robot autonomous exploration and
  mapping under localization uncertainty with expectation-maximization,'' in
  \emph{2024 IEEE International Conference on Robotics and Automation (ICRA)},
  2024, pp. 7236--7242.

\bibitem{wang2011frontier}
Y.~Wang, A.~Liang, and H.~Guan, ``Frontier-based multi-robot map exploration
  using particle swarm optimization,'' in \emph{2011 IEEE symposium on Swarm
  intelligence}.\hskip 1em plus 0.5em minus 0.4em\relax IEEE, 2011, pp. 1--6.

\bibitem{andries2015multi}
M.~Andries and F.~Charpillet, ``Multi-robot taboo-list exploration of unknown
  structured environments,'' in \emph{2015 IEEE/RSJ International Conference on
  Intelligent Robots and Systems (IROS)}.\hskip 1em plus 0.5em minus
  0.4em\relax IEEE, 2015, pp. 5195--5201.

\bibitem{yu2021smmr}
J.~Yu, J.~Tong, Y.~Xu, Z.~Xu, H.~Dong, T.~Yang, and Y.~Wang, ``Smmr-explore:
  Submap-based multi-robot exploration system with multi-robot multi-target
  potential field exploration method,'' in \emph{2021 IEEE International
  Conference on Robotics and Automation (ICRA)}.\hskip 1em plus 0.5em minus
  0.4em\relax IEEE, 2021, pp. 8779--8785.

\bibitem{zlot2002multi}
R.~Zlot, A.~Stentz, M.~B. Dias, and S.~Thayer, ``Multi-robot exploration
  controlled by a market economy,'' in \emph{Proceedings 2002 IEEE
  international conference on robotics and automation (Cat. No. 02CH37292)},
  vol.~3.\hskip 1em plus 0.5em minus 0.4em\relax IEEE, 2002, pp. 3016--3023.

\bibitem{berhault2003robot}
M.~Berhault, H.~Huang, P.~Keskinocak, S.~Koenig, W.~Elmaghraby, P.~Griffin, and
  A.~Kleywegt, ``Robot exploration with combinatorial auctions,'' in
  \emph{Proceedings 2003 IEEE/RSJ International Conference on Intelligent
  Robots and Systems (IROS 2003)(Cat. No. 03CH37453)}, vol.~2.\hskip 1em plus
  0.5em minus 0.4em\relax IEEE, 2003, pp. 1957--1962.

\bibitem{gutin2006traveling}
G.~Gutin and A.~P. Punnen, \emph{The traveling salesman problem and its
  variations}.\hskip 1em plus 0.5em minus 0.4em\relax Springer Science \&
  Business Media, 2006, vol.~12.

\bibitem{faigl2012goal}
J.~Faigl, M.~Kulich, and L.~P{\v{r}}eu{\v{c}}il, ``Goal assignment using
  distance cost in multi-robot exploration,'' in \emph{2012 IEEE/RSJ
  International Conference on Intelligent Robots and Systems}.\hskip 1em plus
  0.5em minus 0.4em\relax IEEE, 2012, pp. 3741--3746.

\bibitem{applegate2006concorde}
D.~Applegate, R.~Bixby, V.~Chvatal, and W.~Cook, ``Concorde tsp solver,'' 2006.

\bibitem{helsgaun2000effective}
K.~Helsgaun, ``An effective implementation of the lin--kernighan traveling
  salesman heuristic,'' \emph{European journal of operational research}, vol.
  126, no.~1, pp. 106--130, 2000.

\bibitem{helsgaun2017extension}
------, ``An extension of the lin-kernighan-helsgaun tsp solver for constrained
  traveling salesman and vehicle routing problems,'' \emph{Roskilde: Roskilde
  University}, vol.~12, pp. 966--980, 2017.

\bibitem{hardouin2020next}
G.~Hardouin, J.~Moras, F.~Morbidi, J.~Marzat, and E.~M. Mouaddib,
  ``Next-best-view planning for surface reconstruction of large-scale 3d
  environments with multiple uavs,'' in \emph{2020 IEEE/RSJ International
  Conference on Intelligent Robots and Systems (IROS)}.\hskip 1em plus 0.5em
  minus 0.4em\relax IEEE, 2020, pp. 1567--1574.

\bibitem{dantzig1959truck}
G.~B. Dantzig and J.~H. Ramser, ``The truck dispatching problem,''
  \emph{Management science}, vol.~6, no.~1, pp. 80--91, 1959.

\bibitem{bartolomei2023fast}
L.~Bartolomei, L.~Teixeira, and M.~Chli, ``Fast multi-uav decentralized
  exploration of forests,'' \emph{IEEE Robotics and Automation Letters}, 2023.

\bibitem{chauhan2012survey}
C.~Chauhan, R.~Gupta, and K.~Pathak, ``Survey of methods of solving tsp along
  with its implementation using dynamic programming approach,''
  \emph{International journal of computer applications}, vol.~52, no.~4, 2012.

\bibitem{yang2023learning}
X.~Yang, S.~Huang, Y.~Sun, Y.~Yang, C.~Yu, W.-W. Tu, H.~Yang, and Y.~Wang,
  ``Learning graph-enhanced commander-executor for multi-agent navigation,'' in
  \emph{Proceedings of the 2023 International Conference on Autonomous Agents
  and Multiagent Systems}, 2023, pp. 1652--1660.

\bibitem{zhang2022centralized}
Q.~Zhang, C.~Lu, A.~Garg, and J.~Foerster, ``Centralized model and exploration
  policy for multi-agent rl,'' in \emph{Proceedings of the 21st International
  Conference on Autonomous Agents and Multiagent Systems}, 2022, pp.
  1500--1508.

\bibitem{marchesini2021centralizing}
E.~Marchesini and A.~Farinelli, ``Centralizing state-values in dueling networks
  for multi-robot reinforcement learning mapless navigation,'' in \emph{2021
  IEEE/RSJ International Conference on Intelligent Robots and Systems
  (IROS)}.\hskip 1em plus 0.5em minus 0.4em\relax IEEE, 2021, pp. 4583--4588.

\bibitem{luo2019multi}
T.~Luo, B.~Subagdja, D.~Wang, and A.-H. Tan, ``Multi-agent collaborative
  exploration through graph-based deep reinforcement learning,'' in \emph{2019
  IEEE International Conference on Agents (ICA)}.\hskip 1em plus 0.5em minus
  0.4em\relax IEEE, 2019, pp. 2--7.

\bibitem{zhou2019bayesian}
X.~Zhou, W.~Wang, T.~Wang, Y.~Lei, and F.~Zhong, ``Bayesian reinforcement
  learning for multi-robot decentralized patrolling in uncertain
  environments,'' \emph{IEEE Transactions on Vehicular Technology}, vol.~68,
  no.~12, pp. 11\,691--11\,703, 2019.

\bibitem{jin2019efficient}
Y.~Jin, Y.~Zhang, J.~Yuan, and X.~Zhang, ``Efficient multi-agent cooperative
  navigation in unknown environments with interlaced deep reinforcement
  learning,'' in \emph{ICASSP 2019-2019 IEEE International Conference on
  Acoustics, Speech and Signal Processing (ICASSP)}.\hskip 1em plus 0.5em minus
  0.4em\relax IEEE, 2019, pp. 2897--2901.

\bibitem{elfakharany2021end}
A.~Elfakharany and Z.~H. Ismail, ``End-to-end deep reinforcement learning for
  decentralized task allocation and navigation for a multi-robot system,''
  \emph{Applied Sciences}, vol.~11, no.~7, p. 2895, 2021.

\bibitem{paul2021learning}
S.~Paul, P.~Ghassemi, and S.~Chowdhury, ``Learning to solve multi-robot task
  allocation with a covariant-attention based neural architecture,'' 2021.

\bibitem{lu2024reinforcement}
Z.~Lu, Y.~Wang, F.~Dai, Y.~Ma, L.~Long, Z.~Zhao, Y.~Zhang, and J.~Li, ``A
  reinforcement learning-based optimization method for task allocation of
  agricultural multi-robots clusters,'' \emph{Computers and Electrical
  Engineering}, vol. 120, p. 109752, 2024.

\bibitem{park2021cooperative}
B.~Park, C.~Kang, and J.~Choi, ``Cooperative multi-robot task allocation with
  reinforcement learning,'' \emph{Applied Sciences}, vol.~12, no.~1, p. 272,
  2021.

\bibitem{NeuralCoMapping}
K.~Ye, S.~Dong, Q.~Fan, H.~Wang, L.~Yi, F.~Xia, J.~Wang, and B.~Chen,
  ``Multi-robot active mapping via neural bipartite graph matching,'' in
  \emph{Proceedings of the IEEE/CVF conference on computer vision and pattern
  recognition}, 2022, pp. 14\,839--14\,848.

\bibitem{dou2015genetic}
J.~Dou, C.~Chen, and P.~Yang, ``Genetic scheduling and reinforcement learning
  in multirobot systems for intelligent warehouses,'' \emph{Mathematical
  Problems in Engineering}, vol. 2015, no.~1, p. 597956, 2015.

\bibitem{task_allo_rl1}
------, ``Genetic scheduling and reinforcement learning in multirobot systems
  for intelligent warehouses,'' \emph{Mathematical Problems in Engineering},
  vol. 2015, no.~1, p. 597956, 2015.

\bibitem{task_allo_rl2}
B.~Park, C.~Kang, and J.~Choi, ``Cooperative multi-robot task allocation with
  reinforcement learning,'' \emph{Applied Sciences}, vol.~12, no.~1, p. 272,
  2021.

\bibitem{task_allo_rl3}
M.~Strens and N.~Windelinckx, ``Combining planning with reinforcement learning
  for multi-robot task allocation,'' in \emph{Symposium on Adaptive Agents and
  Multi-agent Systems}.\hskip 1em plus 0.5em minus 0.4em\relax Springer, 2003,
  pp. 260--274.

\bibitem{dasari2020robonet}
S.~Dasari, F.~Ebert, S.~Tian, S.~Nair, B.~Bucher, K.~Schmeckpeper, S.~Singh,
  S.~Levine, and C.~Finn, ``Robonet: Large-scale multi-robot learning,'' in
  \emph{Conference on Robot Learning}.\hskip 1em plus 0.5em minus 0.4em\relax
  PMLR, 2020, pp. 885--897.

\bibitem{cai2024transformer}
Y.~Cai, X.~He, H.~Guo, W.-Y. Yau, and C.~Lv, ``Transformer-based multi-agent
  reinforcement learning for generalization of heterogeneous multi-robot
  cooperation,'' in \emph{2024 IEEE/RSJ International Conference on Intelligent
  Robots and Systems (IROS)}.\hskip 1em plus 0.5em minus 0.4em\relax IEEE,
  2024, pp. 13\,695--13\,702.

\bibitem{howell2024generalization}
P.~Howell, M.~Rudolph, R.~Torbati, K.~Fu, and H.~Ravichandar, ``Generalization
  of heterogeneous multi-robot policies via awareness and communication of
  capabilities,'' \emph{arXiv preprint arXiv:2401.13127}, 2024.

\bibitem{paden2016survey}
B.~Paden, M.~Čáp, S.~Z. Yong, D.~Yershov, and E.~Frazzoli, ``A survey of
  motion planning and control techniques for self-driving urban vehicles,''
  \emph{IEEE Transactions on Intelligent Vehicles}, vol.~1, no.~1, pp. 33--55,
  2016.

\bibitem{liang2014review}
L.~Yang, J.~Qi, J.~Xiao, and X.~Yong, ``A literature review of uav 3d path
  planning,'' in \emph{Proceeding of the 11th World Congress on Intelligent
  Control and Automation}, 2014, pp. 2376--2381.

\bibitem{hart1968}
P.~E. Hart, N.~J. Nilsson, and B.~Raphael, ``A formal basis for the heuristic
  determination of minimum cost paths,'' \emph{IEEE Transactions on Systems
  Science and Cybernetics}, vol.~4, no.~2, pp. 100--107, 1968.

\bibitem{kim2023multi}
S.~Kim, M.~Corah, J.~Keller, G.~Best, and S.~Scherer, ``Multi-robot multi-room
  exploration with geometric cue extraction and circular decomposition,''
  \emph{IEEE Robotics and Automation Letters}, 2023.

\bibitem{bramblett2022coordinated}
L.~Bramblett, R.~Peddi, and N.~Bezzo, ``Coordinated multi-agent exploration,
  rendezvous, \& task allocation in unknown environments with limited
  connectivity,'' in \emph{2022 IEEE/RSJ International Conference on
  Intelligent Robots and Systems (IROS)}.\hskip 1em plus 0.5em minus
  0.4em\relax IEEE, 2022, pp. 12\,706--12\,712.

\bibitem{yu2021market}
T.~Yu, B.~Deng, W.~Yao, X.~Zhu, and J.~Gui, ``Market-based robots cooperative
  exploration in unknown indoor environment,'' in \emph{2021 IEEE International
  Conference on Unmanned Systems (ICUS)}.\hskip 1em plus 0.5em minus
  0.4em\relax IEEE, 2021, pp. 414--419.

\bibitem{lavalle1998rapidly}
S.~LaValle, ``Rapidly-exploring random trees: A new tool for path planning,''
  \emph{Research Report 9811}, 1998.

\bibitem{mac2016heuristic}
T.~T. Mac, C.~Copot, D.~T. Tran, and R.~De~Keyser, ``Heuristic approaches in
  robot path planning: A survey,'' \emph{Robotics and Autonomous Systems},
  vol.~86, pp. 13--28, 2016.

\bibitem{warren1989global}
C.~W. Warren, ``Global path planning using artificial potential fields,'' in
  \emph{1989 IEEE International Conference on Robotics and Automation}.\hskip
  1em plus 0.5em minus 0.4em\relax IEEE Computer Society, 1989, pp. 316--317.

\bibitem{dorigo2006ant}
M.~Dorigo, M.~Birattari, and T.~Stutzle, ``Ant colony optimization,''
  \emph{IEEE computational intelligence magazine}, vol.~1, no.~4, pp. 28--39,
  2006.

\bibitem{loquercio2021learning}
A.~Loquercio, E.~Kaufmann, R.~Ranftl, M.~M{\"u}ller, V.~Koltun, and
  D.~Scaramuzza, ``Learning high-speed flight in the wild,'' \emph{Science
  Robotics}, vol.~6, no.~59, p. eabg5810, 2021.

\bibitem{li2020aggressive}
S.~Li, E.~{\"O}zt{\"u}rk, C.~De~Wagter, G.~C. De~Croon, and D.~Izzo,
  ``Aggressive online control of a quadrotor via deep network representations
  of optimality principles,'' in \emph{2020 IEEE International Conference on
  Robotics and Automation (ICRA)}.\hskip 1em plus 0.5em minus 0.4em\relax IEEE,
  2020, pp. 6282--6287.

\bibitem{li2018oil}
G.~Li, M.~Mueller, V.~Casser, N.~Smith, D.~L. Michels, and B.~Ghanem, ``Oil:
  Observational imitation learning,'' \emph{arXiv preprint arXiv:1803.01129},
  2018.

\bibitem{penicka2022learning}
R.~Penicka, Y.~Song, E.~Kaufmann, and D.~Scaramuzza, ``Learning minimum-time
  flight in cluttered environments,'' \emph{IEEE Robotics and Automation
  Letters}, vol.~7, no.~3, pp. 7209--7216, 2022.

\bibitem{stachowicz2023fastrlap}
K.~Stachowicz, D.~Shah, A.~Bhorkar, I.~Kostrikov, and S.~Levine, ``Fastrlap: A
  system for learning high-speed driving via deep rl and autonomous
  practicing,'' in \emph{Conference on Robot Learning}.\hskip 1em plus 0.5em
  minus 0.4em\relax PMLR, 2023, pp. 3100--3111.

\bibitem{zhang2024npe}
Y.~Zhang, C.~Yan, J.~Xiao, and M.~Feroskhan, ``Npe-drl: Enhancing perception
  constrained obstacle avoidance with non-expert policy guided reinforcement
  learning,'' \emph{IEEE Transactions on Artificial Intelligence}, 2024.

\bibitem{song2023learning}
Y.~Song, K.~Shi, R.~Penicka, and D.~Scaramuzza, ``Learning perception-aware
  agile flight in cluttered environments,'' in \emph{2023 IEEE International
  Conference on Robotics and Automation (ICRA)}.\hskip 1em plus 0.5em minus
  0.4em\relax IEEE, 2023, pp. 1989--1995.

\bibitem{xing2024bootstrapping}
J.~Xing, A.~Romero, L.~Bauersfeld, and D.~Scaramuzza, ``Bootstrapping
  reinforcement learning with imitation for vision-based agile flight,''
  \emph{arXiv preprint arXiv:2403.12203}, 2024.

\bibitem{ross2011reduction}
S.~Ross, G.~Gordon, and D.~Bagnell, ``A reduction of imitation learning and
  structured prediction to no-regret online learning,'' in \emph{Proceedings of
  the fourteenth international conference on artificial intelligence and
  statistics}.\hskip 1em plus 0.5em minus 0.4em\relax JMLR Workshop and
  Conference Proceedings, 2011, pp. 627--635.

\bibitem{stern2019multi}
R.~Stern, N.~Sturtevant, A.~Felner, S.~Koenig, H.~Ma, T.~Walker, J.~Li,
  D.~Atzmon, L.~Cohen, T.~Kumar, \emph{et~al.}, ``Multi-agent pathfinding:
  Definitions, variants, and benchmarks,'' in \emph{Proceedings of the
  International Symposium on Combinatorial Search}, vol.~10, no.~1, 2019, pp.
  151--158.

\bibitem{sharon2015conflict}
G.~Sharon, R.~Stern, A.~Felner, and N.~R. Sturtevant, ``Conflict-based search
  for optimal multi-agent pathfinding,'' \emph{Artificial intelligence}, vol.
  219, pp. 40--66, 2015.

\bibitem{vcap2015prioritized}
M.~{\v{C}}{\'a}p, P.~Nov{\'a}k, A.~Kleiner, and M.~Seleck{\`y}, ``Prioritized
  planning algorithms for trajectory coordination of multiple mobile robots,''
  \emph{IEEE transactions on automation science and engineering}, vol.~12,
  no.~3, pp. 835--849, 2015.

\bibitem{yu2013multi}
J.~Yu and S.~M. LaValle, ``Multi-agent path planning and network flow,'' in
  \emph{Algorithmic Foundations of Robotics X: Proceedings of the Tenth
  Workshop on the Algorithmic Foundations of Robotics}.\hskip 1em plus 0.5em
  minus 0.4em\relax Springer, 2013, pp. 157--173.

\bibitem{contreras2017distributed}
M.~A. Contreras-Cruz, J.~J. Lopez-Perez, and V.~Ayala-Ramirez, ``Distributed
  path planning for multi-robot teams based on artificial bee colony,'' in
  \emph{2017 IEEE congress on evolutionary computation (CEC)}.\hskip 1em plus
  0.5em minus 0.4em\relax IEEE, 2017, pp. 541--548.

\bibitem{zhou2019robust}
B.~Zhou, F.~Gao, L.~Wang, C.~Liu, and S.~Shen, ``Robust and efficient quadrotor
  trajectory generation for fast autonomous flight,'' \emph{IEEE Robotics and
  Automation Letters}, vol.~4, no.~4, pp. 3529--3536, 2019.

\bibitem{zhou2021raptor}
B.~Zhou, J.~Pan, F.~Gao, and S.~Shen, ``Raptor: Robust and perception-aware
  trajectory replanning for quadrotor fast flight,'' \emph{IEEE Transactions on
  Robotics}, vol.~37, no.~6, pp. 1992--2009, 2021.

\bibitem{zhou2020ego}
X.~Zhou, Z.~Wang, H.~Ye, C.~Xu, and F.~Gao, ``Ego-planner: An esdf-free
  gradient-based local planner for quadrotors,'' \emph{IEEE Robotics and
  Automation Letters}, vol.~6, no.~2, pp. 478--485, 2020.

\bibitem{zhou2021ego}
X.~Zhou, J.~Zhu, H.~Zhou, C.~Xu, and F.~Gao, ``Ego-swarm: A fully autonomous
  and decentralized quadrotor swarm system in cluttered environments,'' in
  \emph{2021 IEEE international conference on robotics and automation
  (ICRA)}.\hskip 1em plus 0.5em minus 0.4em\relax IEEE, 2021, pp. 4101--4107.

\bibitem{burger2018cooperative}
C.~Burger and M.~Lauer, ``Cooperative multiple vehicle trajectory planning
  using miqp,'' in \emph{2018 21st International Conference on Intelligent
  Transportation Systems (ITSC)}.\hskip 1em plus 0.5em minus 0.4em\relax IEEE,
  2018, pp. 602--607.

\bibitem{li2022tract}
Z.~Li, F.~Man, X.~Chen, B.~Zhao, C.~Wu, and X.~Chen, ``Tract: Towards
  large-scale crowdsensing with high-efficiency swarm path planning,'' in
  \emph{Adjunct Proceedings of the 2022 ACM International Joint Conference on
  Pervasive and Ubiquitous Computing and the 2022 ACM International Symposium
  on Wearable Computers}, 2022, pp. 409--414.

\bibitem{chen2024soscheduler}
X.~Chen, Z.~Xiao, Y.~Cheng, C.~Hsia, H.~Wang, J.~Xu, S.~Xu, F.~Dang, X.-P.
  Zhang, Y.~Liu, \emph{et~al.}, ``Soscheduler: Toward proactive and adaptive
  wildfire suppression via multi-uav collaborative scheduling,'' \emph{IEEE
  Internet of Things Journal}, 2024.

\bibitem{mellinger2011minimum}
D.~Mellinger and V.~Kumar, ``Minimum snap trajectory generation and control for
  quadrotors,'' in \emph{2011 IEEE international conference on robotics and
  automation}.\hskip 1em plus 0.5em minus 0.4em\relax IEEE, 2011, pp.
  2520--2525.

\bibitem{gao2018online}
F.~Gao, W.~Wu, Y.~Lin, and S.~Shen, ``Online safe trajectory generation for
  quadrotors using fast marching method and bernstein basis polynomial,'' in
  \emph{2018 IEEE International Conference on Robotics and Automation
  (ICRA)}.\hskip 1em plus 0.5em minus 0.4em\relax IEEE, 2018, pp. 344--351.

\bibitem{ding2019efficient}
W.~Ding, W.~Gao, K.~Wang, and S.~Shen, ``An efficient b-spline-based
  kinodynamic replanning framework for quadrotors,'' \emph{IEEE Transactions on
  Robotics}, vol.~35, no.~6, pp. 1287--1306, 2019.

\bibitem{ren2022bubble}
Y.~Ren, F.~Zhu, W.~Liu, Z.~Wang, Y.~Lin, F.~Gao, and F.~Zhang, ``Bubble
  planner: Planning high-speed smooth quadrotor trajectories using receding
  corridors,'' in \emph{2022 IEEE/RSJ International Conference on Intelligent
  Robots and Systems (IROS)}.\hskip 1em plus 0.5em minus 0.4em\relax IEEE,
  2022, pp. 6332--6339.

\bibitem{zhou2021decentralized}
X.~Zhou, Z.~Wang, X.~Wen, J.~Zhu, C.~Xu, and F.~Gao, ``Decentralized
  spatial-temporal trajectory planning for multicopter swarms,'' \emph{arXiv
  preprint arXiv:2106.12481}, 2021.

\bibitem{han2021fast}
Z.~Han, Z.~Wang, N.~Pan, Y.~Lin, C.~Xu, and F.~Gao, ``Fast-racing: An
  open-source strong baseline for {SE}(3) planning in autonomous drone
  racing,'' \emph{IEEE Robotics and Automation Letters}, vol.~6, no.~4, pp.
  8631--8638, 2021.

\bibitem{quan2022distributed}
L.~Quan, L.~Yin, C.~Xu, and F.~Gao, ``Distributed swarm trajectory optimization
  for formation flight in dense environments,'' in \emph{2022 International
  Conference on Robotics and Automation (ICRA)}.\hskip 1em plus 0.5em minus
  0.4em\relax IEEE, 2022, pp. 4979--4985.

\bibitem{wang2022geometrically}
Z.~Wang, X.~Zhou, C.~Xu, and F.~Gao, ``Geometrically constrained trajectory
  optimization for multicopters,'' \emph{IEEE Transactions on Robotics},
  vol.~38, no.~5, pp. 3259--3278, 2022.

\bibitem{tordesillas2021mader}
J.~Tordesillas and J.~P. How, ``Mader: Trajectory planner in multiagent and
  dynamic environments,'' \emph{IEEE Transactions on Robotics}, vol.~38, no.~1,
  pp. 463--476, 2021.

\bibitem{tordesillas2022minvo}
------, ``{MINVO} basis: Finding simplexes with minimum volume enclosing
  polynomial curves,'' \emph{Computer-Aided Design}, vol. 151, p. 103341, 2022.

\bibitem{gao2017gradient}
F.~Gao, Y.~Lin, and S.~Shen, ``Gradient-based online safe trajectory generation
  for quadrotor flight in complex environments,'' in \emph{2017 IEEE/RSJ
  international conference on intelligent robots and systems (IROS)}.\hskip 1em
  plus 0.5em minus 0.4em\relax IEEE, 2017, pp. 3681--3688.

\bibitem{usenko2017real}
V.~Usenko, L.~Von~Stumberg, A.~Pangercic, and D.~Cremers, ``Real-time
  trajectory replanning for mavs using uniform b-splines and a 3d circular
  buffer,'' in \emph{2017 IEEE/RSJ International Conference on Intelligent
  Robots and Systems (IROS)}.\hskip 1em plus 0.5em minus 0.4em\relax IEEE,
  2017, pp. 215--222.

\bibitem{yu2023asynchronous}
C.~Yu, X.~Yang, J.~Gao, J.~Chen, Y.~Li, J.~Liu, Y.~Xiang, R.~Huang, H.~Yang,
  Y.~Wu, \emph{et~al.}, ``Asynchronous multi-agent reinforcement learning for
  efficient real-time multi-robot cooperative exploration,'' in
  \emph{Proceedings of the 2023 International Conference on Autonomous Agents
  and Multiagent Systems}, 2023, pp. 1107--1115.

\bibitem{yue2019reinforcement}
W.~Yue, X.~Guan, and Y.~Xi, ``Reinforcement learning based approach for
  multi-uav cooperative searching in unknown environments,'' in \emph{2019
  Chinese Automation Congress (CAC)}.\hskip 1em plus 0.5em minus 0.4em\relax
  IEEE, 2019, pp. 2018--2023.

\bibitem{chen2024ddl}
X.~Chen, H.~Wang, Y.~Cheng, H.~Fu, Y.~Liu, F.~Dang, Y.~Liu, J.~Cui, and
  X.~Chen, ``Ddl: Empowering delivery drones with large-scale urban sensing
  capability,'' \emph{IEEE Journal of Selected Topics in Signal Processing},
  2024.

\bibitem{chen2022deliversense}
X.~Chen, H.~Wang, Z.~Li, W.~Ding, F.~Dang, C.~Wu, and X.~Chen, ``Deliversense:
  Efficient delivery drone scheduling for crowdsensing with deep reinforcement
  learning,'' in \emph{Adjunct proceedings of the 2022 ACM international joint
  conference on pervasive and ubiquitous computing and the 2022 ACM
  international symposium on wearable computers}, 2022, pp. 403--408.

\bibitem{wei2021multi}
Y.~Wei and R.~Zheng, ``Multi-robot path planning for mobile sensing through
  deep reinforcement learning,'' in \emph{IEEE INFOCOM 2021-IEEE Conference on
  Computer Communications}.\hskip 1em plus 0.5em minus 0.4em\relax IEEE, 2021,
  pp. 1--10.

\bibitem{liu2024heterogeneous}
X.~Liu, D.~Guo, X.~Zhang, and H.~Liu, ``Heterogeneous embodied multi-agent
  collaboration,'' \emph{IEEE Robotics and Automation Letters}, 2024.

\bibitem{wang2023maddpg}
R.~Wang, Y.~Xia, Y.~Wei, Z.~Pan, and J.~Li, ``Maddpg-based distributed
  cooperative search strategy for heterogeneous agents system,'' in
  \emph{Chinese Conference on Swarm Intelligence and Cooperative
  Control}.\hskip 1em plus 0.5em minus 0.4em\relax Springer, 2023, pp.
  292--305.

\bibitem{MANTM}
X.~Yang, Y.~Yang, C.~Yu, J.~Chen, J.~Yu, H.~Ren, H.~Yang, and Y.~Wang, ``Active
  neural topological mapping for multi-agent exploration,'' \emph{IEEE Robotics
  and Automation Letters}, vol.~9, no.~1, pp. 303--310, 2023.

\bibitem{cai2013combined}
Y.~Cai, S.~X. Yang, and X.~Xu, ``A combined hierarchical reinforcement learning
  based approach for multi-robot cooperative target searching in complex
  unknown environments,'' in \emph{2013 IEEE symposium on adaptive dynamic
  programming and reinforcement learning (ADPRL)}.\hskip 1em plus 0.5em minus
  0.4em\relax IEEE, 2013, pp. 52--59.

\bibitem{liang2021hierarchical}
Z.~Liang, J.~Cao, W.~Lin, J.~Chen, and H.~Xu, ``Hierarchical deep reinforcement
  learning for multi-robot cooperation in partially observable environment,''
  in \emph{2021 IEEE third international conference on cognitive machine
  intelligence (CogMI)}.\hskip 1em plus 0.5em minus 0.4em\relax IEEE, 2021, pp.
  272--281.

\bibitem{zhang2022h2gnn}
H.~Zhang, J.~Cheng, L.~Zhang, Y.~Li, and W.~Zhang, ``H2gnn: Hierarchical-hops
  graph neural networks for multi-robot exploration in unknown environments,''
  \emph{IEEE Robotics and Automation Letters}, vol.~7, no.~2, pp. 3435--3442,
  2022.

\bibitem{tolstaya2021multi}
E.~Tolstaya, J.~Paulos, V.~Kumar, and A.~Ribeiro, ``Multi-robot coverage and
  exploration using spatial graph neural networks,'' in \emph{2021 IEEE/RSJ
  International Conference on Intelligent Robots and Systems (IROS)}.\hskip 1em
  plus 0.5em minus 0.4em\relax IEEE, 2021, pp. 8944--8950.

\bibitem{li2020graph}
Q.~Li, F.~Gama, A.~Ribeiro, and A.~Prorok, ``Graph neural networks for
  decentralized multi-robot path planning,'' in \emph{2020 IEEE/RSJ
  international conference on intelligent robots and systems (IROS)}.\hskip 1em
  plus 0.5em minus 0.4em\relax IEEE, 2020, pp. 11\,785--11\,792.

\bibitem{he2023multi}
X.~He, X.~Shi, J.~Hu, and Y.~Wang, ``Multi-robot navigation with graph
  attention neural network and hierarchical motion planning,'' \emph{Journal of
  Intelligent \& Robotic Systems}, vol. 109, no.~2, p.~25, 2023.

\bibitem{chen2024transformer}
Q.~Chen, R.~Wang, M.~Lyu, and J.~Zhang, ``Transformer-based reinforcement
  learning for multi-robot autonomous exploration,'' \emph{Sensors}, vol.~24,
  no.~16, p. 5083, 2024.

\bibitem{chen2023transformer}
L.~Chen, Y.~Wang, Z.~Miao, Y.~Mo, M.~Feng, Z.~Zhou, and H.~Wang,
  ``Transformer-based imitative reinforcement learning for multirobot path
  planning,'' \emph{IEEE Transactions on Industrial Informatics}, vol.~19,
  no.~10, pp. 10\,233--10\,243, 2023.

\bibitem{wang2024multi}
W.~Wang, L.~Mao, R.~Wang, and B.-C. Min, ``Multi-robot cooperative
  socially-aware navigation using multi-agent reinforcement learning,'' in
  \emph{2024 IEEE International Conference on Robotics and Automation
  (ICRA)}.\hskip 1em plus 0.5em minus 0.4em\relax IEEE, 2024, pp.
  12\,353--12\,360.

\bibitem{deng2022multi}
L.~Deng, W.~Gong, and L.~Li, ``Multi-robot exploration in unknown environments
  via multi-agent deep reinforcement learning,'' in \emph{2022 China Automation
  Congress (CAC)}.\hskip 1em plus 0.5em minus 0.4em\relax IEEE, 2022, pp.
  6898--6902.

\bibitem{mete2023coordinated}
A.~Mete, M.~Mouhoub, and A.~M. Farid, ``Coordinated multi-robot exploration
  using reinforcement learning,'' in \emph{2023 International Conference on
  Unmanned Aircraft Systems (ICUAS)}.\hskip 1em plus 0.5em minus 0.4em\relax
  IEEE, 2023, pp. 265--272.

\bibitem{lin2019end}
J.~Lin, X.~Yang, P.~Zheng, and H.~Cheng, ``End-to-end decentralized multi-robot
  navigation in unknown complex environments via deep reinforcement learning,''
  in \emph{2019 IEEE International Conference on Mechatronics and Automation
  (ICMA)}.\hskip 1em plus 0.5em minus 0.4em\relax IEEE, 2019, pp. 2493--2500.

\bibitem{jun2019goal}
H.~Jun, H.~Kim, and B.~Lee, ``Goal-driven navigation for non-holonomic
  multi-robot system by learning collision,'' in \emph{2019 International
  Conference on Robotics and Automation (ICRA)}.\hskip 1em plus 0.5em minus
  0.4em\relax IEEE, 2019, pp. 1758--1764.

\bibitem{shi2019end}
H.~Shi, L.~Shi, M.~Xu, and K.-S. Hwang, ``End-to-end navigation strategy with
  deep reinforcement learning for mobile robots,'' \emph{IEEE Transactions on
  Industrial Informatics}, vol.~16, no.~4, pp. 2393--2402, 2019.

\bibitem{cao2021multi}
J.~Cao, Y.~Wang, Y.~Liu, and X.~Ni, ``Multi-robot learning dynamic obstacle
  avoidance in formation with information-directed exploration,'' \emph{IEEE
  Transactions on Emerging Topics in Computational Intelligence}, vol.~6,
  no.~6, pp. 1357--1367, 2021.

\bibitem{huang2024collision}
Z.~Huang, Z.~Yang, R.~Krupani, B.~{\c{S}}enba{\c{s}}lar, S.~Batra, and G.~S.
  Sukhatme, ``Collision avoidance and navigation for a quadrotor swarm using
  end-to-end deep reinforcement learning,'' in \emph{2024 IEEE International
  Conference on Robotics and Automation (ICRA)}.\hskip 1em plus 0.5em minus
  0.4em\relax IEEE, 2024, pp. 300--306.

\bibitem{bettini2023heterogeneous}
M.~Bettini, A.~Shankar, and A.~Prorok, ``Heterogeneous multi-robot
  reinforcement learning,'' in \emph{Proceedings of the 2023 International
  Conference on Autonomous Agents and Multiagent Systems}, 2023, pp.
  1485--1494.

\bibitem{seraj2024heterogeneous}
E.~Seraj, R.~Paleja, L.~Pimentel, K.~M. Lee, Z.~Wang, D.~Martin, M.~Sklar,
  J.~Zhang, Z.~Kakish, and M.~Gombolay, ``Heterogeneous policy networks for
  composite robot team communication and coordination,'' \emph{IEEE
  Transactions on Robotics}, 2024.

\bibitem{burgard2000collaborative}
W.~Burgard, M.~Moors, D.~Fox, R.~Simmons, and S.~Thrun, ``Collaborative
  multi-robot exploration,'' in \emph{Proceedings 2000 ICRA. Millennium
  Conference. IEEE International Conference on Robotics and Automation.
  Symposia Proceedings (Cat. No. 00CH37065)}, vol.~1.\hskip 1em plus 0.5em
  minus 0.4em\relax IEEE, 2000, pp. 476--481.

\bibitem{ko2003practical}
J.~Ko, B.~Stewart, D.~Fox, K.~Konolige, and B.~Limketkai, ``A practical,
  decision-theoretic approach to multi-robot mapping and exploration,'' in
  \emph{Proceedings 2003 IEEE/RSJ International Conference on Intelligent
  Robots and Systems (IROS 2003)(Cat. No. 03CH37453)}, vol.~4.\hskip 1em plus
  0.5em minus 0.4em\relax IEEE, 2003, pp. 3232--3238.

\bibitem{hornung2013octomap}
A.~Hornung, K.~M. Wurm, M.~Bennewitz, C.~Stachniss, and W.~Burgard, ``Octomap:
  An efficient probabilistic 3d mapping framework based on octrees,''
  \emph{Autonomous robots}, vol.~34, pp. 189--206, 2013.

\bibitem{duberg2020ufomap}
D.~Duberg and P.~Jensfelt, ``Ufomap: An efficient probabilistic 3d mapping
  framework that embraces the unknown,'' \emph{IEEE Robotics and Automation
  Letters}, vol.~5, no.~4, pp. 6411--6418, 2020.

\bibitem{yang2021graph}
F.~Yang, D.-H. Lee, J.~Keller, and S.~Scherer, ``Graph-based topological
  exploration planning in large-scale 3d environments,'' in \emph{2021 IEEE
  International Conference on Robotics and Automation (ICRA)}.\hskip 1em plus
  0.5em minus 0.4em\relax IEEE, 2021, pp. 12\,730--12\,736.

\bibitem{gao2022meeting}
Y.~Gao, Y.~Wang, X.~Zhong, T.~Yang, M.~Wang, Z.~Xu, Y.~Wang, Y.~Lin, C.~Xu, and
  F.~Gao, ``Meeting-merging-mission: A multi-robot coordinate framework for
  large-scale communication-limited exploration,'' in \emph{2022 IEEE/RSJ
  International Conference on Intelligent Robots and Systems (IROS)}.\hskip 1em
  plus 0.5em minus 0.4em\relax IEEE, 2022, pp. 13\,700--13\,707.

\bibitem{corah2019communication}
M.~Corah, C.~O’Meadhra, K.~Goel, and N.~Michael, ``Communication-efficient
  planning and mapping for multi-robot exploration in large environments,''
  \emph{IEEE Robotics and Automation Letters}, vol.~4, no.~2, pp. 1715--1721,
  2019.

\bibitem{o2018variable}
C.~O’Meadhra, W.~Tabib, and N.~Michael, ``Variable resolution occupancy
  mapping using gaussian mixture models,'' \emph{IEEE Robotics and Automation
  Letters}, vol.~4, no.~2, pp. 2015--2022, 2018.

\bibitem{dijkstra2022note}
E.~W. Dijkstra, ``A note on two problems in connexion with graphs,'' in
  \emph{Edsger Wybe Dijkstra: his life, work, and legacy}, 2022, pp. 287--290.

\bibitem{bayer2021decentralized}
J.~Bayer and J.~Faigl, ``Decentralized topological mapping for multi-robot
  autonomous exploration under low-bandwidth communication,'' in \emph{2021
  European Conference on Mobile Robots (ECMR)}.\hskip 1em plus 0.5em minus
  0.4em\relax IEEE, 2021, pp. 1--7.

\bibitem{wang2021spatial}
Z.~Wang and N.~Papanikolopoulos, ``Spatial action maps augmented with visit
  frequency maps for exploration tasks,'' in \emph{2021 IEEE/RSJ International
  Conference on Intelligent Robots and Systems (IROS)}.\hskip 1em plus 0.5em
  minus 0.4em\relax IEEE, 2021, pp. 3175--3181.

\bibitem{chen2023efficient}
X.~Chen, A.~N. Iyer, Z.~Wang, and A.~H. Qureshi, ``Efficient q-learning over
  visit frequency maps for multi-agent exploration of unknown environments,''
  in \emph{2023 IEEE/RSJ International Conference on Intelligent Robots and
  Systems (IROS)}.\hskip 1em plus 0.5em minus 0.4em\relax IEEE, 2023, pp.
  1893--1900.

\bibitem{de2009role}
J.~De~Hoog, S.~Cameron, and A.~Visser, ``Role-based autonomous multi-robot
  exploration,'' in \emph{2009 Computation World: Future Computing, Service
  Computation, Cognitive, Adaptive, Content, Patterns}.\hskip 1em plus 0.5em
  minus 0.4em\relax IEEE, 2009, pp. 482--487.

\bibitem{cesare2015multi}
K.~Cesare, R.~Skeele, S.-H. Yoo, Y.~Zhang, and G.~Hollinger, ``Multi-uav
  exploration with limited communication and battery,'' in \emph{2015 IEEE
  international conference on robotics and automation (ICRA)}.\hskip 1em plus
  0.5em minus 0.4em\relax IEEE, 2015, pp. 2230--2235.

\bibitem{clark2022propem}
L.~Clark, J.~A. Edlund, M.~S. Net, T.~S. Vaquero, and A.-a. Agha-Mohammadi,
  ``Propem-l: Radio propagation environment modeling and learning for
  communication-aware multi-robot exploration,'' \emph{arXiv preprint
  arXiv:2205.01267}, 2022.

\bibitem{xia2023relink}
L.~Xia, B.~Deng, J.~Pan, X.~Zhang, P.~Duan, B.~Zhou, and H.~Cheng, ``Relink:
  Real-time line-of-sight-based deployment framework of multi-robot for
  maintaining a communication network,'' \emph{IEEE Robotics and Automation
  Letters}, vol.~8, no.~12, pp. 8152--8159, 2023.

\bibitem{flab2021}
\BIBentryALTinterwordspacing
F.~Lab. (2021) An open-source l-bfgs solver. [Online]. Available:
  \url{https://github.com/ZJU-FAST-Lab/LBFGS-Lite}
\BIBentrySTDinterwordspacing

\bibitem{sheng2004multi}
W.~Sheng, Q.~Yang, S.~Ci, and N.~Xi, ``Multi-robot area exploration with
  limited-range communications,'' in \emph{2004 IEEE/RSJ International
  Conference on Intelligent Robots and Systems (IROS)(IEEE Cat. No.
  04CH37566)}, vol.~2.\hskip 1em plus 0.5em minus 0.4em\relax IEEE, 2004, pp.
  1414--1419.

\bibitem{masaba2021gvgexp}
K.~Masaba and A.~Q. Li, ``Gvgexp: Communication-constrained multi-robot
  exploration system based on generalized voronoi graphs,'' in \emph{2021
  International Symposium on Multi-Robot and Multi-Agent Systems (MRS)}.\hskip
  1em plus 0.5em minus 0.4em\relax IEEE, 2021, pp. 146--154.

\bibitem{woosley2021bid}
B.~Woosley, C.~Nieto-Granda, J.~G. Rogers, N.~Fung, and A.~Schang, ``Bid
  prediction for multi-robot exploration with disrupted communications,'' in
  \emph{2021 IEEE International Symposium on Safety, Security, and Rescue
  Robotics (SSRR)}.\hskip 1em plus 0.5em minus 0.4em\relax IEEE, 2021, pp.
  210--216.

\bibitem{schack2024sound}
M.~A. Schack, J.~G. Rogers, and N.~T. Dantam, ``The sound of silence:
  Exploiting information from the lack of communication,'' \emph{IEEE Robotics
  and Automation Letters}, 2024.

\bibitem{DARPA2022}
\BIBentryALTinterwordspacing
{DARPA}. (2022) Darpa subterranean challenge. Accessed: 2022-03-08. [Online].
  Available: \url{https://www.subchallenge.com}
\BIBentrySTDinterwordspacing

\bibitem{orekhov2022darpa}
V.~Orekhov and T.~H. Chung, ``The darpa subterranean challenge: A synopsis of
  the circuits stage.'' \emph{Field Robotics}, vol.~2, no.~1, pp. 735--747,
  2022.

\bibitem{ohradzansky2022multi}
M.~T. Ohradzansky, E.~R. Rush, D.~G. Riley, A.~B. Mills, S.~Ahmad, S.~McGuire,
  H.~Biggie, K.~Harlow, M.~J. Miles, E.~W. Frew, \emph{et~al.}, ``Multi-agent
  autonomy: Advancements and challenges in subterranean exploration,''
  \emph{Field Robotics}, vol.~2, pp. 1068--1104, 2022.

\bibitem{scherer2022resilient}
S.~Scherer, V.~Agrawal, G.~Best, C.~Cao, K.~Cujic, R.~Darnley, R.~DeBortoli,
  E.~Dexheimer, B.~Drozd, R.~Garg, \emph{et~al.}, ``Resilient and modular
  subterranean exploration with a team of roving and flying robots,''
  \emph{Field Robotics}, vol.~2, pp. 678--734, 2022.

\bibitem{roucek2021system}
T.~Roucek, M.~Pecka, P.~C{\i}zek, T.~Petr{\i}cek, J.~Bayer, \emph{et~al.},
  ``System for multi-robotic exploration of underground environments
  ctu-cras-norlab in the darpa subterranean challenge,'' \emph{arXiv preprint
  arXiv:2110.05911}, 2021.

\bibitem{hudson2022heterogeneous}
N.~Hudson, F.~Talbot, M.~Cox, J.~Williams, T.~Hines, A.~Pitt, B.~Wood,
  D.~Frousheger, K.~Lo~Surdo, T.~Molnar, \emph{et~al.}, ``Heterogeneous ground
  and air platforms, homogeneous sensing: Team csiro data61’s approach to the
  darpa subterranean challenge,'' \emph{Field Robotics}, vol.~2, no.~1, pp.
  595--636, 2022.

\bibitem{agha2022nebula}
A.~Agha, K.~Otsu, B.~Morrell, D.~D. Fan, R.~Thakker, A.~Santamaria-Navarro,
  S.-K. Kim, A.~Bouman, X.~Lei, J.~Edlund, \emph{et~al.}, ``Nebula: Team
  costar's robotic autonomy solution that won phase ii of darpa subterranean
  challenge,'' \emph{Field robotics}, vol.~2, pp. 1432--1506, 2022.

\bibitem{tranzatto2022cerberus}
M.~Tranzatto, F.~Mascarich, L.~Bernreiter, C.~Godinho, M.~Camurri, S.~Khattak,
  T.~Dang, V.~Reijgwart, J.~Loeje, D.~Wisth, \emph{et~al.}, ``Cerberus:
  Autonomous legged and aerial robotic exploration in the tunnel and urban
  circuits of the darpa subterranean challenge,'' \emph{arXiv preprint
  arXiv:2201.07067}, vol.~3, 2022.

\bibitem{rizk2019cooperative}
Y.~Rizk, M.~Awad, and E.~W. Tunstel, ``Cooperative heterogeneous multi-robot
  systems: A survey,'' \emph{ACM Computing Surveys (CSUR)}, vol.~52, no.~2, pp.
  1--31, 2019.

\bibitem{santamaria2019towards}
A.~Santamaria-Navarro, R.~Thakker, D.~D. Fan, B.~Morrell, and A.-A.
  Agha-Mohammadi, ``Towards resilient autonomous navigation of drones,'' in
  \emph{The International Symposium of Robotics Research}.\hskip 1em plus 0.5em
  minus 0.4em\relax Springer, 2019, pp. 922--937.

\bibitem{badi2020lamp}
K.~Ebadi, Y.~Chang, M.~Palieri, A.~Stephens, A.~Hatteland, E.~Heiden,
  A.~Thakur, N.~Funabiki, B.~Morrell, S.~Wood, L.~Carlone, and A.-a.
  Agha-mohammadi, ``Lamp: Large-scale autonomous mapping and positioning for
  exploration of perceptually-degraded subterranean environments,'' in
  \emph{2020 IEEE International Conference on Robotics and Automation (ICRA)},
  2020, pp. 80--86.

\bibitem{kim2021plgrim}
S.-K. Kim, A.~Bouman, G.~Salhotra, D.~D. Fan, K.~Otsu, J.~Burdick, and A.-a.
  Agha-mohammadi, ``Plgrim: Hierarchical value learning for large-scale
  exploration in unknown environments,'' in \emph{Proceedings of the
  international conference on automated planning and scheduling}, vol.~31,
  2021, pp. 652--662.

\bibitem{ginting2021chord}
M.~F. Ginting, K.~Otsu, J.~A. Edlund, J.~Gao, and A.-A. Agha-Mohammadi,
  ``Chord: Distributed data-sharing via hybrid ros 1 and 2 for multi-robot
  exploration of large-scale complex environments,'' \emph{IEEE Robotics and
  Automation Letters}, vol.~6, no.~3, pp. 5064--5071, 2021.

\bibitem{khattak2020complementary}
S.~Khattak, H.~Nguyen, F.~Mascarich, T.~Dang, and K.~Alexis, ``Complementary
  multi--modal sensor fusion for resilient robot pose estimation in
  subterranean environments,'' in \emph{2020 International Conference on
  Unmanned Aircraft Systems (ICUAS)}.\hskip 1em plus 0.5em minus 0.4em\relax
  IEEE, 2020, pp. 1024--1029.

\bibitem{schneider2018maplab}
T.~Schneider, M.~Dymczyk, M.~Fehr, K.~Egger, S.~Lynen, I.~Gilitschenski, and
  R.~Siegwart, ``maplab: An open framework for research in visual-inertial
  mapping and localization,'' \emph{IEEE Robotics and Automation Letters},
  vol.~3, no.~3, pp. 1418--1425, 2018.

\bibitem{dang2020graph}
T.~Dang, M.~Tranzatto, S.~Khattak, F.~Mascarich, K.~Alexis, and M.~Hutter,
  ``Graph-based subterranean exploration path planning using aerial and legged
  robots,'' \emph{Journal of Field Robotics}, vol.~37, no.~8, pp. 1363--1388,
  2020.

\bibitem{ramezani2022wildcat}
M.~Ramezani, K.~Khosoussi, G.~Catt, P.~Moghadam, J.~Williams, P.~Borges,
  F.~Pauling, and N.~Kottege, ``Wildcat: Online continuous-time 3d
  lidar-inertial slam,'' \emph{arXiv preprint arXiv:2205.12595}, 2022.

\bibitem{chen1992object}
Y.~Chen and G.~Medioni, ``Object modelling by registration of multiple range
  images,'' \emph{Image and vision computing}, vol.~10, no.~3, pp. 145--155,
  1992.

\bibitem{zhang2014loam}
J.~Zhang, S.~Singh, \emph{et~al.}, ``Loam: Lidar odometry and mapping in
  real-time.'' in \emph{Robotics: Science and systems}, vol.~2, no.~9.\hskip
  1em plus 0.5em minus 0.4em\relax Berkeley, CA, 2014, pp. 1--9.

\bibitem{zhao2021super}
S.~Zhao, H.~Zhang, P.~Wang, L.~Nogueira, and S.~Scherer, ``Super odometry:
  Imu-centric lidar-visual-inertial estimator for challenging environments,''
  in \emph{2021 IEEE/RSJ International Conference on Intelligent Robots and
  Systems (IROS)}.\hskip 1em plus 0.5em minus 0.4em\relax IEEE, 2021, pp.
  8729--8736.

\bibitem{hess2016real}
W.~Hess, D.~Kohler, H.~Rapp, and D.~Andor, ``Real-time loop closure in 2d lidar
  slam,'' in \emph{2016 IEEE international conference on robotics and
  automation (ICRA)}.\hskip 1em plus 0.5em minus 0.4em\relax IEEE, 2016, pp.
  1271--1278.

\bibitem{oleynikova2017voxblox}
H.~Oleynikova, Z.~Taylor, M.~Fehr, R.~Siegwart, and J.~Nieto, ``Voxblox:
  Incremental 3d euclidean signed distance fields for on-board mav planning,''
  in \emph{2017 IEEE/RSJ International Conference on Intelligent Robots and
  Systems (IROS)}.\hskip 1em plus 0.5em minus 0.4em\relax IEEE, 2017, pp.
  1366--1373.

\bibitem{ohradzansky2020reactive}
M.~T. Ohradzansky, A.~B. Mills, E.~R. Rush, D.~G. Riley, E.~W. Frew, and J.~S.
  Humbert, ``Reactive control and metric-topological planning for
  exploration,'' in \emph{2020 IEEE international conference on robotics and
  automation (ICRA)}.\hskip 1em plus 0.5em minus 0.4em\relax IEEE, 2020, pp.
  4073--4079.

\bibitem{riley2021assessment}
D.~G. Riley and E.~W. Frew, ``Assessment of coordinated heterogeneous
  exploration of complex environments,'' in \emph{2021 IEEE Conference on
  Control Technology and Applications (CCTA)}.\hskip 1em plus 0.5em minus
  0.4em\relax IEEE, 2021, pp. 138--143.

\bibitem{azpurua2023survey}
H.~Azp{\'u}rua, M.~Saboia, G.~M. Freitas, L.~Clark, A.-a. Agha-mohammadi,
  G.~Pessin, M.~F. Campos, and D.~G. Macharet, ``A survey on the autonomous
  exploration of confined subterranean spaces: Perspectives from real-word and
  industrial robotic deployments,'' \emph{Robotics and Autonomous Systems},
  vol. 160, p. 104304, 2023.

\bibitem{chung2023into}
T.~H. Chung, V.~Orekhov, and A.~Maio, ``Into the robotic depths: analysis and
  insights from the darpa subterranean challenge,'' \emph{Annual Review of
  Control, Robotics, and Autonomous Systems}, vol.~6, no.~1, pp. 477--502,
  2023.

\bibitem{gielis2022critical}
J.~Gielis, A.~Shankar, and A.~Prorok, ``A critical review of communications in
  multi-robot systems,'' \emph{Current robotics reports}, vol.~3, no.~4, pp.
  213--225, 2022.

\bibitem{ullah2024mobile}
I.~Ullah, D.~Adhikari, H.~Khan, M.~S. Anwar, S.~Ahmad, and X.~Bai, ``Mobile
  robot localization: Current challenges and future prospective,''
  \emph{Computer Science Review}, vol.~53, p. 100651, 2024.

\bibitem{wang2024survey}
Y.~Wang, Y.~Tian, J.~Chen, K.~Xu, and X.~Ding, ``A survey of visual slam in
  dynamic environment: The evolution from geometric to semantic approaches,''
  \emph{IEEE Transactions on Instrumentation and Measurement}, 2024.

\bibitem{placed2023survey}
J.~A. Placed, J.~Strader, H.~Carrillo, N.~Atanasov, V.~Indelman, L.~Carlone,
  and J.~A. Castellanos, ``A survey on active simultaneous localization and
  mapping: State of the art and new frontiers,'' \emph{IEEE Transactions on
  Robotics}, vol.~39, no.~3, pp. 1686--1705, 2023.

\bibitem{yin2023automerge}
P.~Yin, S.~Zhao, H.~Lai, R.~Ge, J.~Zhang, H.~Choset, and S.~Scherer,
  ``Automerge: A framework for map assembling and smoothing in city-scale
  environments,'' \emph{IEEE Transactions on Robotics}, 2023.

\bibitem{baril2021kilometer}
D.~Baril, S.-P. Desch{\^e}nes, O.~Gamache, M.~Vaidis, D.~LaRocque, J.~Laconte,
  V.~Kubelka, P.~Gigu{\`e}re, and F.~Pomerleau, ``Kilometer-scale autonomous
  navigation in subarctic forests: challenges and lessons learned,''
  \emph{arXiv preprint arXiv:2111.13981}, 2021.

\bibitem{poudel2022task}
S.~Poudel and S.~Moh, ``Task assignment algorithms for unmanned aerial vehicle
  networks: A comprehensive survey,'' \emph{Vehicular Communications}, vol.~35,
  p. 100469, 2022.

\bibitem{schulman2017ppo}
\BIBentryALTinterwordspacing
J.~Schulman, F.~Wolski, P.~Dhariwal, A.~Radford, and O.~Klimov, ``Proximal
  policy optimization algorithms,'' 2017. [Online]. Available:
  \url{https://arxiv.org/abs/1707.06347}
\BIBentrySTDinterwordspacing

\bibitem{koenig2004design}
N.~Koenig and A.~Howard, ``Design and use paradigms for gazebo, an open-source
  multi-robot simulator,'' in \emph{2004 IEEE/RSJ international conference on
  intelligent robots and systems (IROS)(IEEE Cat. No. 04CH37566)},
  vol.~3.\hskip 1em plus 0.5em minus 0.4em\relax Ieee, 2004, pp. 2149--2154.

\bibitem{song2021flightmare}
Y.~Song, S.~Naji, E.~Kaufmann, A.~Loquercio, and D.~Scaramuzza, ``Flightmare: A
  flexible quadrotor simulator,'' in \emph{Conference on Robot Learning}.\hskip
  1em plus 0.5em minus 0.4em\relax PMLR, 2021, pp. 1147--1157.

\bibitem{xu2024omnidrones}
B.~Xu, F.~Gao, C.~Yu, R.~Zhang, Y.~Wu, and Y.~Wang, ``Omnidrones: An efficient
  and flexible platform for reinforcement learning in drone control,''
  \emph{IEEE Robotics and Automation Letters}, 2024.

\bibitem{ding2018hierarchical}
W.~Ding, S.~Li, H.~Qian, and Y.~Chen, ``Hierarchical reinforcement learning
  framework towards multi-agent navigation,'' in \emph{2018 IEEE international
  conference on robotics and biomimetics (ROBIO)}.\hskip 1em plus 0.5em minus
  0.4em\relax IEEE, 2018, pp. 237--242.

\bibitem{tang2023autonomous}
H.~Tang, H.~Zhang, Z.~Shi, X.~Chen, W.~Ding, and X.-P. Zhang, ``Autonomous
  swarm robot coordination via mean-field control embedding multi-agent
  reinforcement learning,'' in \emph{2023 IEEE/RSJ International Conference on
  Intelligent Robots and Systems (IROS)}.\hskip 1em plus 0.5em minus
  0.4em\relax IEEE, 2023, pp. 8820--8826.

\bibitem{zhang2022multi}
Z.~Zhang, X.~Wang, Q.~Zhang, and T.~Hu, ``Multi-robot cooperative pursuit via
  potential field-enhanced reinforcement learning,'' in \emph{2022
  International Conference on Robotics and Automation (ICRA)}.\hskip 1em plus
  0.5em minus 0.4em\relax IEEE, 2022, pp. 8808--8814.

\bibitem{lee2024multi}
E.~S. Lee and Y.~M. Kim, ``Multi-agent exploration with similarity score map
  and topological memory,'' \emph{IEEE Robotics and Automation Letters}, 2024.

\bibitem{mozian2020learning}
M.~Mozian, J.~C.~G. Higuera, D.~Meger, and G.~Dudek, ``Learning domain
  randomization distributions for training robust locomotion policies,'' in
  \emph{2020 IEEE/RSJ International Conference on Intelligent Robots and
  Systems (IROS)}.\hskip 1em plus 0.5em minus 0.4em\relax IEEE, 2020, pp.
  6112--6117.

\bibitem{horvath2022object}
D.~Horv{\'a}th, G.~Erd{\H{o}}s, Z.~Istenes, T.~Horv{\'a}th, and S.~F{\"o}ldi,
  ``Object detection using sim2real domain randomization for robotic
  applications,'' \emph{IEEE Transactions on Robotics}, vol.~39, no.~2, pp.
  1225--1243, 2022.

\end{thebibliography}

\vfill\pagebreak

\end{document}